\pdfoutput=1

\documentclass[11pt]{article}

\usepackage[final]{acl}
\usepackage{pifont}

\newcommand{\cmark}{\ding{51}}  
\newcommand{\xmark}{\ding{55}}  

\usepackage{times}
\usepackage{latexsym}

\usepackage[T1]{fontenc}
\usepackage{fontawesome}

\usepackage[utf8]{inputenc}

\usepackage{microtype}

\usepackage{booktabs}
\usepackage{multirow}
\usepackage{amssymb}

\usepackage{inconsolata}
\usepackage{multirow}
\usepackage{adjustbox}
\usepackage{linguex}
\usepackage{makecell}
\usepackage[breakable]{tcolorbox}
\usepackage{CJKutf8}
\usepackage{amsmath}
\usepackage{longtable}  
\usepackage{booktabs}    
\usepackage{makecell}    
\usepackage{array}       

\title{MMSciBench: Benchmarking Language Models on Chinese Multimodal Scientific Problems}

\author{Xinwu Ye\textsuperscript{1}, Chengfan Li\textsuperscript{2}, Siming Chen\textsuperscript{3,4}, Wei Wei\textsuperscript{5 *}, Xiangru Tang\textsuperscript{1}\footnotemark[1] \\
\textsuperscript{1}Department of Computer Science, Yale University, \\
\textsuperscript{2}Department of Computer Science, Brown University, \\
\textsuperscript{3}School of Data Science, Fudan University, \\
\textsuperscript{4}Shanghai Key Laboratory of Data Science, \\
\textsuperscript{5}Datawiz LLC
}

\begin{document}
\begin{CJK*}{UTF8}{gbsn}
\maketitle

\renewcommand{\thefootnote}{\fnsymbol{footnote}}
\footnotetext[1]{Corresponding Authors.}
\renewcommand{\thefootnote}{\arabic{footnote}}

\begin{abstract}


Recent advances in large language models (LLMs) and vision-language models (LVLMs) have shown promise across many tasks, yet their scientific reasoning capabilities remain untested, particularly in multimodal settings. We present MMSciBench, a benchmark for evaluating mathematical and physical reasoning through text-only and text-image formats, with human-annotated difficulty levels, solutions with detailed explanations, and taxonomic mappings.
Evaluation of state-of-the-art models reveals significant limitations, with even the best model achieving only \textbf{63.77\%} accuracy and particularly struggling with visual reasoning tasks. Our analysis exposes critical gaps in complex reasoning and visual-textual integration, establishing MMSciBench as a rigorous standard for measuring progress in multimodal scientific understanding. 
The code for MMSciBench is open-sourced at GitHub\footnote{\url{https://github.com/xinwuye/MMSciBench-code}}, and the dataset is available at Hugging Face\footnote{\url{https://huggingface.co/datasets/XinwuYe/MMSciBench}}.

\end{abstract}

\section{Introduction}

\begin{figure}[tb!]
    \small
    \centering
    \begin{tcolorbox}[colframe=magenta!50, colback=magenta!10, title=Question \& Standard Solution]
        \textbf{Question}

        Question (Single Choice): As shown in the figure, two identical right-angled glass prisms $ABC$ are placed with their $AC$ faces parallel to each other, and between them is a uniform unknown transparent medium. A monochromatic thin light beam $O$ is incident perpendicular to the $AB$ face. (\quad) is the possible exit light path in the diagram.

        \includegraphics[width=\textwidth]{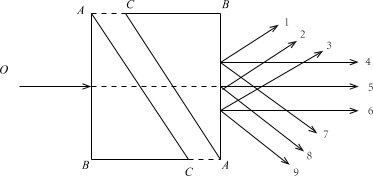}
        
        Options:
        
        A. Any one of the lines 1, 2, 3 (parallel to each other)
        
        B. Any one of the lines 4, 5, 6 (parallel to each other)
        
        C. Any one of the lines 7, 8, 9 (parallel to each other)
        
        D. Only one of the lines 4 or 6

        \textbf{Difficulty Level}: 0.7

        \textbf{Domain}: Quantum Mechanics

        \textbf{Module}: Light and Its Applications

        \textbf{Chapter}: Snell's Law

        \textbf{Standard Solution}: B
    \end{tcolorbox}

    \begin{tcolorbox}[colframe=cyan!50, colback=cyan!10, title=Explanation]
        This question primarily tests knowledge of prism-related problems.
        
        Option analysis: According to the problem description, the refractive index of the medium between the two right-angled prisms is unknown. It may be greater than, equal to, or smaller than the refractive index of the glass. The possible light path diagrams are as follows:

        \includegraphics[width=\textwidth]{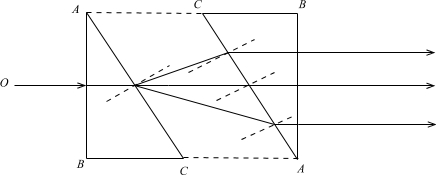}
        
        Therefore, Option B is correct, and Options A, C, and D are incorrect.
        
        In conclusion, the correct answer to this question is B.
    \end{tcolorbox}
    \vspace{-.1cm}
    \caption{The English translation of an example of a physics MCQ, featuring a single-choice question, the correct answer, and a detailed explanation to aid understanding. The original Chinese version is shown in Fig. \ref{fig:physics_mcq} in the appendix.}
    \label{fig:physics_mcq_en}
        \vspace{-.1cm}    \vspace{-.1cm}
    \vspace{-.1cm}
    \vspace{-.1cm}

\end{figure}

Scientific reasoning represents a crucial test of artificial intelligence (AI) systems' ability to understand and apply complex concepts, making it essential for developing truly intelligent models ~\cite{evans2023msvec,liang2024scemqa,zhang2023evaluating,truhn2023large,ma2024sciagent,sprueill2023monte}.
Recent advancements in LLMs like GPTs \cite{brown2020languagemodelsfewshotlearners,openai2024gpt4technicalreport} and Llama \cite{grattafiori2024llama3herdmodels} have significantly transformed the field of natural language processing (NLP). 
Despite these advances, scientific reasoning remains challenging for these models, facing several key limitations:
\textit{(1) Lack of multimodal evaluation}: While LVLMs have emerged as powerful models capable of processing both images and text, existing scientific benchmarks are predominantly text-only, preventing comprehensive assessment of visual-textual reasoning abilities.
\textit{(2) Limited domain coverage}: Current scientific datasets either focus too narrowly on individual subjects or too broadly across scientific areas, failing to systematically evaluate understanding of key concepts within specific disciplines.
\textit{(3) Insufficient assessment granularity}: Existing benchmarks lack human-annotated difficulty levels and structured taxonomies of scientific concepts, making it challenging to evaluate models' performance across different complexity levels and specific knowledge domains.
These limitations create an urgent need for a benchmark that can effectively evaluate both LLMs' and LVLMs' scientific reasoning abilities while addressing these challenges.

To address these challenges, we introduce MMSciBench, a benchmark focused on mathematics and physics that evaluates scientific reasoning capabilities. Our benchmark makes three key contributions:
(1) A comprehensive evaluation framework that combines multiple-choice questions (MCQs) and open-ended Q\&A problems, designed to test diverse reasoning skills across mathematical and physical domains.
(2) A novel multimodal assessment approach incorporating both text-only and text-image formats, enabling direct comparison of models' unimodal versus multimodal reasoning capabilities.
(3) A hierarchical taxonomy of scientific concepts with human-annotated difficulty levels, detailed solutions, and explanations for each problem.
We conducted extensive experiments using five state-of-the-art LVLMs (including both open-source and proprietary models) on the complete dataset, and two mathematics-specialized LLMs on text-only questions. For consistent evaluation across models, we employed GPT-4o as an automated assessor.


Our evaluation reveals significant limitations in current models' multimodal scientific reasoning capabilities. Gemini 1.5 Pro 002 achieved the highest accuracy (\textbf{63.77\%}), followed by Qwen2-VL-72B-Instruct (\textbf{56.11\%}), Claude 3.5 Sonnet (\textbf{53.95\%}), and GPT-4o (\textbf{50.94\%}), while Llama-3.2-90B-Vision-Instruct performed substantially lower (\textbf{31.19\%}). Analysis across task types exposed three critical challenges:
(1) performance degradation on open-ended tasks, with Gemini 1.5 Pro 002's accuracy dropping by an average of \textbf{22.32\%} compared to multiple-choice questions;
(2) systematic failures in complex mathematical and physical reasoning, particularly in domains requiring multi-step problem-solving; 
(3) limited visual-textual integration, evidenced by Gemini 1.5 Pro 002's \textbf{36.28\%} performance gap between text-only and text-image questions.
Notably, model performance improved when utilizing explicit chain-of-thought prompting and English-language reasoning, even for Chinese-language questions, suggesting potential pathways for enhancing scientific reasoning capabilities.

        

        
        
        
        
        


    
        
    

\section{MMSciBench}

\subsection{Data Collection and Preprocessing} 
The benchmark data was originally curated by K-12 teachers who annotate questions, detailed step-by-step solutions, final answers, difficulty level, knowledge points, as well as a range of other metadata, including question type (MCQ/Q\&A), modality (text-only/text-image), and subject (math/physics).
The detailed curation process is described in Sec. \ref{app:dataset_curation_process} in the appendix.
The dataset\footnote{The dataset is released under the apache-2.0 license.} includes precise text descriptions, high-resolution images, and high-quality solutions, all compiled and shared as part of a collaborative research effort aimed at advancing AI benchmarking standards.
The benchmark includes standardized problem prompts to ensure consistent model input.
The benchmark also integrates a GPT-4o evaluator to assess answer correctness, focusing on scientific capability over format adherence.
Each question in the dataset is assigned a human-annotated hardness score ranging from 0 to 1, where 1 represents the most challenging questions, and zero denotes the easiest.

To ensure benchmark quality and rigor, we implemented a systematic data curation process. We filtered out questions with incomplete information or duplicate content, focusing on problems with well-defined, quantifiable answers. Following our emphasis on challenging scientific reasoning, we selected questions with human-annotated difficulty scores $\geq$ 0.7 on a standardized scale. To maintain consistent evaluation conditions, we limited visual content to a maximum of one image per question.
To enable systematic knowledge categorization, we employed GPT-4o to annotate each question according to a three-level subject-specific taxonomy, detailed in Sec. \ref{sec:dataset_description}. The classification results were thoroughly validated by experienced K-12 curriculum specialists to ensure accuracy and alignment with educational standards. This taxonomic analysis confirmed that our filtered dataset maintains comprehensive coverage of key scientific concepts while focusing on challenging problems. Following preprocessing and validation, the final benchmark contains \textbf{4,482} question-solution pairs that enable rigorous evaluation of models' scientific reasoning capabilities across diverse domains.

\subsection{Dataset Description}
\label{sec:dataset_description}
\begin{figure}[t!]
    \centering
    \begin{adjustbox}{center}
    \includegraphics[width=\linewidth]{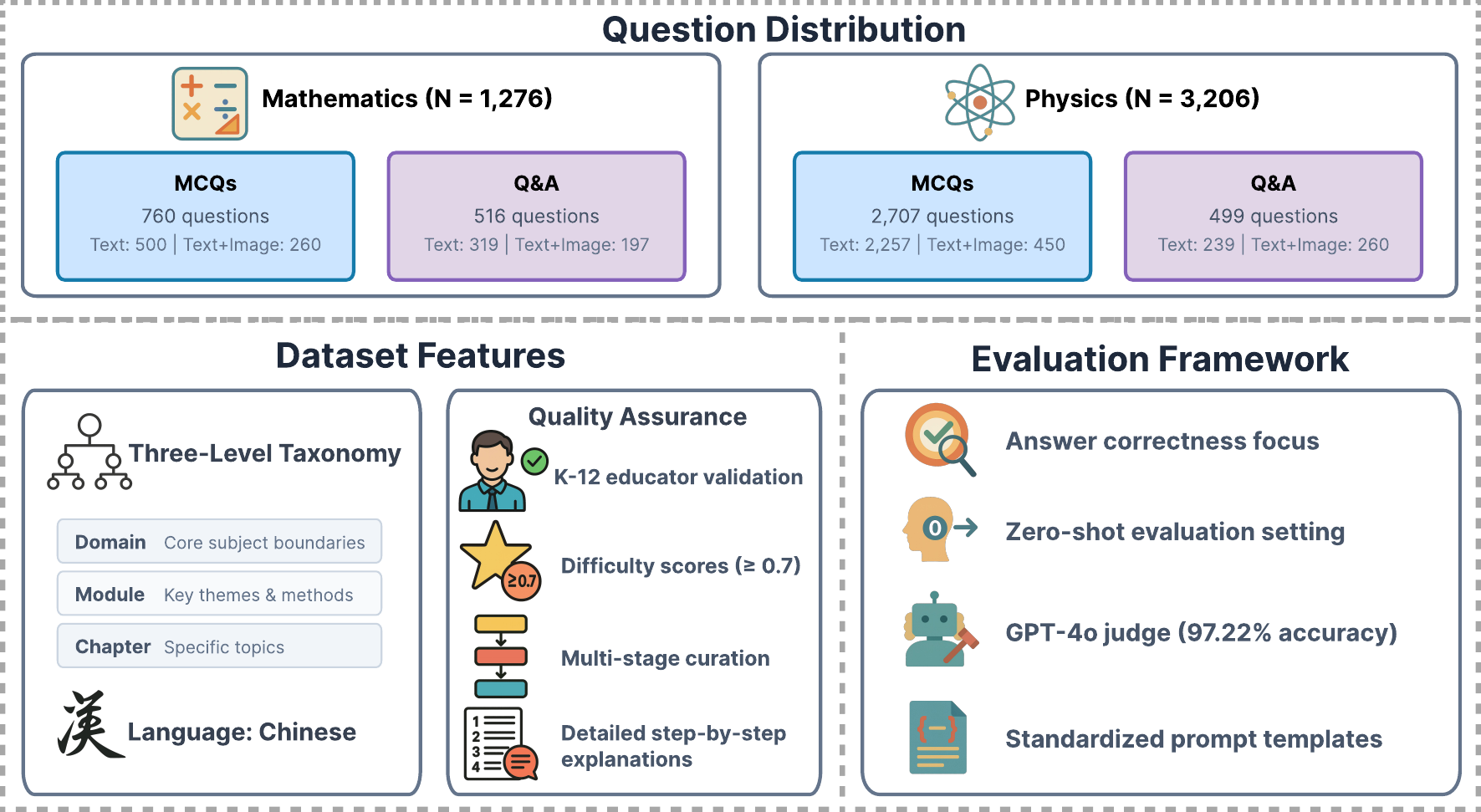}
    \end{adjustbox}
    \caption{The overview of MMSciBench, describing the question distribution, dataset features, and the evaluation framework.}
    \label{fig:overview}
\end{figure}

Fig. \ref{fig:overview} provides a visual overview of MMSciBench, detailing the distribution of questions in the dataset, dataset features, and the evaluation framework.

\paragraph{Data Characteristics} The MMSciBench dataset offers several distinct advantages over previous scientific datasets:

\begin{enumerate}
\item \textbf{Curriculum Coverage:} The benchmark spans essential high school mathematics and physics concepts through carefully curated MCQs and open-ended Q\&A questions. We maintain comprehensiveness while keeping the dataset size tractable ($N=\textbf{4,482}$).
\vspace{-.2cm}
\item \textbf{Quality Assurance:} Questions undergo multi-stage validation by K-12 educators and domain experts, ensuring pedagogical relevance and technical accuracy. Each question includes detailed solutions and explanations.
\vspace{-.2cm}
\item \textbf{Multimodal Design:} The parallel text-only and text-image question formats enable systematic comparison of unimodal and multimodal reasoning capabilities.
\vspace{-.2cm}
\item \textbf{Structured Assessment:} Questions are organized through a three-level taxonomy and annotated with standardized difficulty scores, facilitating fine-grained analysis of model performance.
\end{enumerate}

An example of a physics MCQ in English is shown in Fig. \ref{fig:physics_mcq_en}, with the original Chinese version available in Fig. \ref{fig:physics_mcq} in the appendix. Additionally, a detailed comparison between MMSciBench and other scientific benchmarks is provided from multiple perspectives in Table \ref{tab:comparison}.


\begin{table*}[!htbp]
\setlength{\tabcolsep}{3pt}
\centering
\resizebox{\textwidth}{!}{%
\begin{tabular}{lccccccc}
\toprule
\textbf{Benchmark} & \textbf{Subject(s)} & \textbf{Modality} & \textbf{Key Knowledge Pt.} & \textbf{Explanation} & \textbf{Language} & \textbf{Difficulty} & \textbf{Size} \\
\midrule
\textbf{TRIGO} \cite{xiong2023trigo} & M & T & \xmark & \cmark & Lean & High School & 11K \\
\textbf{DMath} \cite{kim2023ain} & M & T & \cmark & \cmark & EN\&KR & Grade School & 10K \\
\textbf{GRASP} \cite{jassim2023grasp} & P & T\&V & \cmark & \xmark & EN & Basic & 2K \\
\midrule
\textbf{MSVEC} \cite{evans2023msvec} & P, O & T & \xmark & \cmark & EN & College & 200 \\
\textbf{SciOL} \cite{tarsi2024sciol} & P, O & T\&I & \xmark & \xmark & EN & College & 18M \\
\textbf{SciEval} \cite{sun2024scieval} & P, O & T & \cmark & Partial & EN & Multi-level & 16K \\
\textbf{SceMQA} \cite{liang2024scemqa} & M, P, O & T\&I & \cmark & \cmark & EN & Pre-College & 1K \\
\textbf{GAOKAO-Bench} \cite{zhang2023evaluating} & M, P, O & T & \xmark & \cmark & ZH & High School & 3K \\
\textbf{GAOKAO-MM} \cite{zong2024gaokao} & M, P, O & T, T\&I & \xmark & \cmark & ZH & High School & 650 \\
\textbf{SciBench} \cite{wangscibench} & M, P, O & T, T\&I & \cmark & \cmark & EN & College & 869 \\
\midrule
\textbf{MMSci} \cite{li2024mmsci} & Multi-subj. (72) & T\&I & \cmark & \cmark & EN & PhD-level & 108K \\
\textbf{M3Exam} \cite{zhang2023m3exam} & Multi-subj. & T, T\&I & \cmark & (A) & Multi. (9 lang.) & K-12 levels & 12K \\
\textbf{SciFIBench} \cite{roberts2024scifibench} & Multi-subj. & T\&I & \cmark & \cmark & EN & Academic & 2K \\ 
\textbf{EXAMS-V} \cite{das2024exams} & Multi-subj. (20) & T\&I & \cmark & (A) & Multi. (11 lang.) & School Exams (4-12) & 21K \\
\midrule
\textbf{OlympiadBench} \cite{he2024olympiadbench} & M, P & T, T\&I & \cmark & \cmark & EN, ZH & Olympiad & 8K \\
\midrule
\textbf{MMSciBench (Ours)} & M, P & T, T\&I & \cmark & \cmark & ZH & High School & 4K \\
\bottomrule
\end{tabular}%
}
\caption{Comparison of MMSciBench with existing benchmarks. For Subject(s), `M' denotes mathematics, `P' denotes physics, and `O' denotes other subject(s). For Modality, `T' denotes text-only data, `T\&I' denotes text-image data pairs, and `T\&V' denotes text-video data pairs. EN, ZH, KR, and Lean represent English, simplified Chinese, Korean, and the Lean theorem prover language, respectively. Multi. denotes multilingual. For Explanation, `(A)' indicates answers are provided.}
\label{tab:comparison}
\end{table*}

\begin{figure}[t!]
    \centering
    \begin{adjustbox}{center}
    \includegraphics[width=\linewidth]{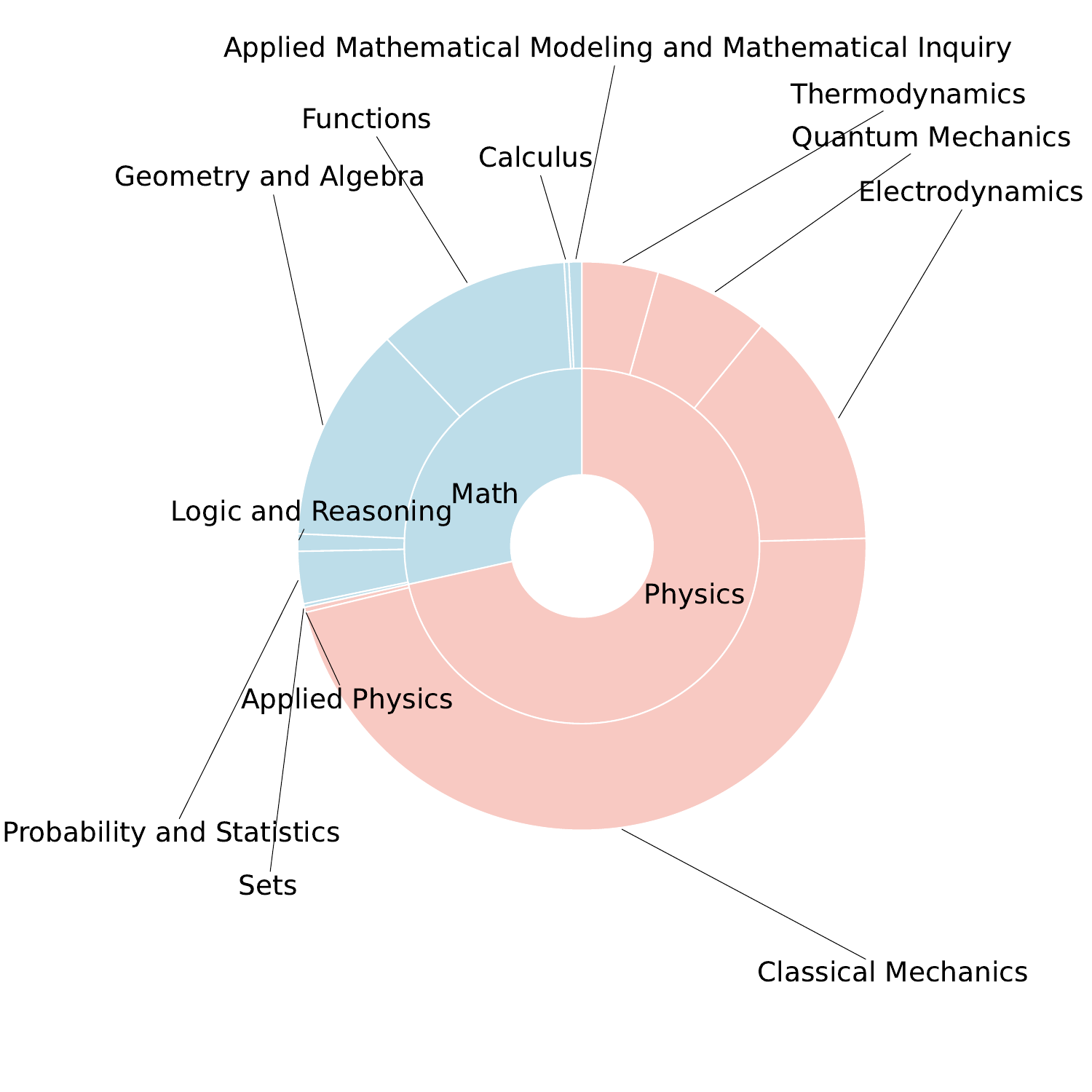}
    \end{adjustbox}
    \caption{The distribution of data in MMSciBench according to the first-level key knowledge points for each subject.}
    \label{fig:key_point_pie}
    \vspace{-.2cm}
\end{figure}

\paragraph{Data Statistics} 

\begin{table}[!htbp]
\centering
 \resizebox{0.5\textwidth}{!}{
\begin{tabular}{lcccccc}
\toprule
\multirow{2}{*}{\textbf{Question Type}} & \multicolumn{2}{c}{\textbf{Math}} & \multicolumn{2}{c}{\textbf{Physics}} & \multicolumn{2}{c}{\textbf{Overall}} \\
\cmidrule(lr){2-3}\cmidrule(lr){4-5}\cmidrule(lr){6-7}
 & \textbf{MCQs} & \textbf{Q\&A} & \textbf{MCQs} & \textbf{Q\&A} & \textbf{MCQs} & \textbf{Q\&A} \\
\midrule
\textbf{Text\&Image} & 260 & 197 & 450 & 260 & 710 & 457 \\
\textbf{Text} & 500 & 319 & 2257 & 239 & 2757 & 558 \\
\midrule
\textbf{Total} & 760 & 516 & 2707 & 499 & 3467 & 1015 \\
\bottomrule
\end{tabular}}
\caption{Distribution of questions in MMSciBench by image presence, subject, and question type.}
\label{tab:data_statistics}
\end{table}

MMSciBench comprises \textbf{4,482} questions, distributed across modalities and question types, as shown in Table~\ref{tab:data_statistics}. The distribution of core knowledge areas for mathematics and physics is illustrated in Figure~\ref{fig:key_point_pie}.

\paragraph{Taxonomy} 
The taxonomy used in MMSciBench has three levels: \textit{Domain}, \textit{Module}, and \textit{Chapter}:
\begin{itemize}
    \item \textbf{Domain}: Core subject areas that define fundamental knowledge boundaries. Mathematics domains include ``Sets'' and ``Functions'', while physics encompasses ``Classical Mechanics'', ``Electrodynamics'', and ``Quantum Mechanics''. \textit{Domains} group related topics under a common framework.
    
    \item \textbf{Module}: Subdivisions within \textit{Domains} that focus on key themes or methods. Examples include ``Probability and Statistics'' in mathematics and ``Mechanical Motion and Physical Models'' in physics. \textit{Modules} scaffold learning by clustering related topics.
    
    \item \textbf{Chapter}: The most detailed level, covering specific topics within a \textit{Module}. For instance, mathematics \textit{Chapters} under ``Functions'' include ``Exponential Functions'' and ``Trigonometric Functions'', while physics \textit{Chapters} under ``Interactions and Laws of Motion'' include ``Hooke's Law'' and ``Equilibrium Conditions of Concurrent Forces''. \textit{Chapters} enable fine-grained content analysis and annotation.
\end{itemize}

\section{Experiment Settings}
\subsection{Evaluated Models}
We evaluated our benchmark using five state-of-the-art LVLMs: GPT-4o, Claude 3.5 Sonnet \cite{anthropic2024claude}, Gemini 1.5 Pro 002 \cite{team2024gemini}, Llama-3.2-90B-Vision-Instruct, and Qwen2-VL-72B-Instruct \cite{wang2024qwen2}.

In addition, we evaluated several other state-of-the-art models. For models specifically designed for mathematical problem-solving, we included Qwen2.5-Math-72B-Instruct \cite{yang2024qwen25mathtechnicalreportmathematical} and DeepSeekMath-7B-Instruct \cite{shao2024deepseekmathpushinglimitsmathematical} on text-only mathematics questions. 
To further broaden our comparison, particularly on multimodal reasoning which is a key focus of MMSciBench, we evaluated state-of-the-art reasoning models, including o1 \cite{jaech2024openai} and Claude 3.7 Sonnet \cite{AnthropicClaude37Sonnet2025}, on the text-image math questions subset of our benchmark.
For reproducibility, all evaluations used a fixed sampling temperature of 0.

\subsection{Evaluation Criteria}
To evaluate the models, we use accuracy as the metric, a widely adopted standard in existing research, for all question types in MMSciBench. Our evaluation focuses solely on whether the final answer is correct, without considering intermediate solution steps. This criterion is naturally suited for MCQ evaluation, as grading is based on the selected choice(s) in practice. For Q\&A questions, this approach ensures a fair and objective comparison by emphasizing the correctness of the final answer rather than incorporating subjective human-defined grading that accounts for intermediate steps.

The evaluation workflow involves first generating answers for MMSciBench questions using each model. 
Given the full response of tested models, GPT-4o is then employed to compare the model’s final answer directly against the ground truth to assess correctness.
In existing studies, MCQs often require models to adhere to a specified output format, imposed through prompts, with regular expression rules used to extract the selected choice(s). However, during our experiments, we observed that some models struggled to consistently follow these formatting instructions, complicating this approach. In fact, none of the models achieved a 100\% compliance rate with the formatting guidelines.
To ensure the evaluation focuses on the models' scientific knowledge and reasoning abilities, rather than being influenced by format compliance issues, we employ GPT-4o to judge whether the final answers are equivalent.

\subsection{Prompt Design}

\begin{figure}[t!]
    \centering
    \begin{adjustbox}{center}
    \includegraphics[width=\linewidth]{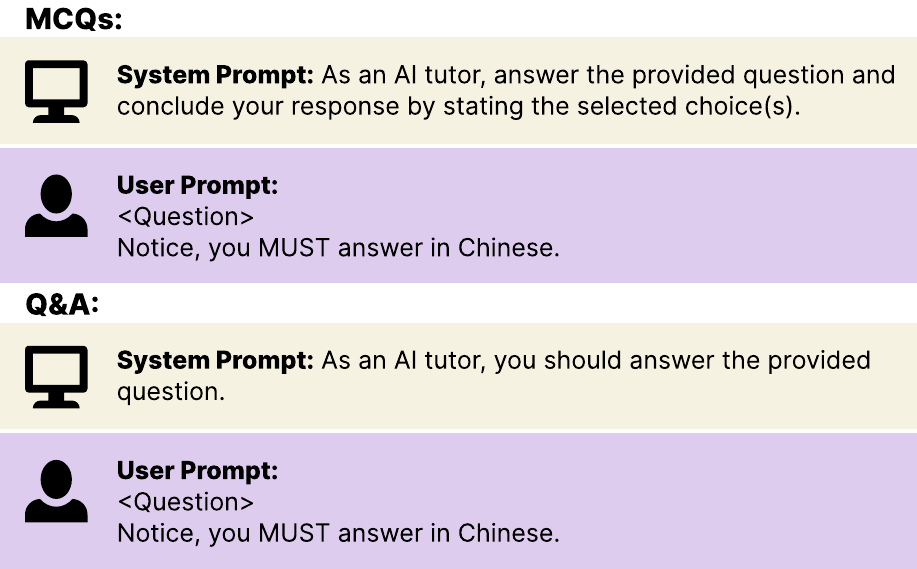}
    \end{adjustbox}
    \caption{The prompt template designed for requesting models to answer questions in Chinese, where the <Question> is sourced from MMSciBench.}
    \label{fig:prompt_chn}
\end{figure}

We use prompts customized for different question types to evaluate the models in a zero-shot setting. For each question type, we apply the same specific prompt template across all models, avoiding model-specific prompt engineering that might explicitly guide reasoning or impose tailored requirements. The prompt template is illustrated in Fig. \ref{fig:prompt_chn}. To assess the models’ intrinsic scientific abilities, the prompts used in the evaluation do not include additional key knowledge points or supplementary information from the dataset, although such information could be incorporated in future research for other purposes.
Since the dataset is in Chinese, we instruct the models to provide their answers in Chinese to ensure consistency with the dataset’s language.

For the LLM-as-a-judge evaluation \cite{gu2024survey,chen2024mllm,raju2024constructing}, we sample \textbf{180} instances of evaluated data and iteratively refined the judging prompts by manually verifying the accuracy of the judgments. 
This refinement process achieved a \textbf{97.22\%} judgment accuracy, the agreement rate between GPT-4o’s and human evaluations.
The GPT-4o evaluator outputs a deterministic final judgment in a required format. Detailed prompts are provided in Sec. \ref{app:prompts} in the appendix.

To further analyze the reliability of GPT-4o as a judge, we conducted a qualitative study of its accuracy on this task.
In this study, we first identified all instances incorrectly judged by GPT-4o from our evaluation sample and randomly selected an additional \textbf{50} correctly-judged instances from the same sample for comparative analysis. We then classified the error patterns in the incorrectly judged cases and documented representative examples of both successful and failed judgments by GPT-4o. (Further examples are provided in Sec. \ref{app:qualitative_study_examples} in the appendix). 
Our analysis revealed that GPT-4o generally excels at understanding semantic equivalence between a model's generated answer and the provided standard solution. 
However, occasional errors in its judging capabilities were observed. These errors primarily stemmed from two main causes: (1) misinterpretations of the standard of correctness (e.g., being too lenient on incomplete answers or overlooking errors in multi-part questions), and (2) mistakenly equating two distinct mathematical formulas as equivalent when they were not. It underscores that definitions of `correctness' and the detailed comparison of complex mathematical expressions are sources of potential discrepancy.

\section{Results}

\subsection{Model Performance}

\begin{table}[t!]
\centering
 \resizebox{0.5\textwidth}{!}{
\begin{tabular}{lccc}
\toprule
\textbf{Models} & \textbf{Math} & \textbf{Physics} & \textbf{Overall} \\
\midrule
\textbf{Llama-3.2-90B-Vision-Instruct} & 16.69\% & 36.96\% & 31.19\% \\
\textbf{Gemini 1.5 Pro 002} & 56.74\% & 66.56\% & 63.77\% \\
\textbf{Claude 3.5 Sonnet} & 37.38\% & 60.54\% & 53.95\% \\
\textbf{GPT-4o} & 35.97\% & 56.89\% & 50.94\% \\
\textbf{Qwen2-VL-72B-Instruct} & 35.50\% & 64.32\% & 56.11\% \\
\midrule
\textbf{Qwen2.5-Math-72B-Instruct} & 57.39\%$^{*}$ & -- & -- \\
\textbf{DeepSeekMath-7B-Instruct} & 21.86\%$^{*}$ & -- & -- \\
\midrule
\textbf{o1} & 67.40\%$^{\dagger}$ & -- & -- \\
\textbf{Claude 3.7 Sonnet} & 37.64\%$^{\dagger}$ & -- & -- \\
\bottomrule
\end{tabular}}
\caption{Accuracies of models across different subjects. Values marked with $^{*}$ indicate accuracies reported only on text-only questions, as the corresponding models are not multimodal. Values marked with $^{\dagger}$ indicate accuracies reported only on the text-image questions.}
    \vspace{-.1cm}
    \vspace{-.1cm}
\label{tab:overall_acc}
\end{table}

\paragraph{Overall and Subject-wise Performance}
Table \ref{tab:overall_acc} presents the overall and subject-specific accuracies of the five LVLMs on the full MMSciBench dataset, the accuracies of the two math-specific LLMs on the text-only math subset, and the three reasoning models on the text-image math subset. Gemini 1.5 Pro 002 achieves the highest overall accuracy at \textbf{63.77\%}, significantly outperforming the other LVLMs in the evaluation. It consistently surpasses all competitors across each of the examined subjects, highlighting the substantial challenge posed by the benchmark, even for the most advanced LVLMs.
Among the remaining LVLMs, Qwen2-VL-72B-Instruct ranks second overall with an accuracy of \textbf{56.11\%}, outperforming Claude 3.5 Sonnet (\textbf{53.95\%}) and GPT-4o (\textbf{50.94\%}). In contrast, Llama-3.2-90B-Vision-Instruct lags far behind, recording the lowest overall accuracy of \textbf{31.19\%}.

For the two math-specific LLMs, Qwen2.5-Math-72B-Instruct demonstrates notable performance with an accuracy of \textbf{57.39\%} on text-only math questions, while DeepSeekMath-7B-Instruct significantly underperforms, achieving only \textbf{21.86\%}. This discrepancy is expected, given the difference in model sizes.
Furthermore, on the text-image math questions subset, o1 achieves an accuracy of \textbf{67.40\%}, outperforming Claude 3.7 Sonnet (\textbf{37.64\%}). These results on a challenging multimodal subset further highlight the varying capabilities of different models in utilizing visual and textual information for mathematical reasoning.
Another noteworthy observation is the variation in performance across subjects, with models consistently performing better in physics. This finding will be analyzed further in Sec. \ref{sec:visual_understanding}. 
Additionally, a breakdown of performance by question difficulty is provided in Sec. \ref{app:acc_by_difficulty} of the appendix.

\begin{table}[!htbp]
\centering
 \resizebox{0.5\textwidth}{!}{
\begin{tabular}{l cc cc cc}
\toprule
\multirow{2}{*}{\textbf{Models}} & 
\multicolumn{2}{c}{\textbf{Math}} &
\multicolumn{2}{c}{\textbf{Physics}} & 
\multicolumn{2}{c}{\textbf{Overall}} \\
\cmidrule(lr){2-3}\cmidrule(lr){4-5}\cmidrule(lr){6-7}
 & \textbf{MCQs} & \textbf{Q\&A} & \textbf{MCQs} & \textbf{Q\&A} & \textbf{MCQs} & \textbf{Q\&A} \\
\midrule
\multirow{2}{*}{\textbf{Llama-3.2-90B-Vision-Instruct}} & 25.39\% & 3.88\% & 41.49\% & 12.42\% & 37.96\% & 8.08\% \\
& \underline{1.52\%} &  & \underline{21.48\%} &  & \underline{17.1\%} &  \\[5pt]
    
\multirow{2}{*}{\textbf{Gemini 1.5 Pro 002}} & 63.16\% & 47.29\% & 70.41\% & 45.69\% & 68.82\% & 46.50\% \\
& \underline{39.29\%} &  & \underline{50.40\%} &  & \underline{47.96\%} &  \\[5pt]
    
\multirow{2}{*}{\textbf{Claude 3.5 Sonnet}} & 48.03\% & 21.71\% & 65.35\% & 34.47\% & 61.55\% & 27.98\% \\
& \underline{24.16\%} &  & \underline{45.34\%} &  & \underline{40.69\%} &  \\[5pt]
    
\multirow{2}{*}{\textbf{GPT-4o}} & 44.47\% & 23.45\% & 61.17\% & 33.67\% & 57.51\% & 28.47\% \\
& \underline{20.60\%} &  & \underline{41.16\%} &  & \underline{36.65\%} &  \\[5pt]

\multirow{2}{*}{\textbf{Qwen2-VL-72B-Instruct}} & 46.58\% & 19.19\% & 71.07\% & 27.66\% & 65.71\% & 23.35\% \\
& \underline{22.71\%} &  & \underline{51.06\%} &  & \underline{44.85\%} &  \\[5pt]
    
\midrule
\multirow{2}{*}{\textbf{Qwen2.5-Math-72B-Instruct}} & 66.80\%$^{*}$ & 42.63\%$^{*}$ & -- & -- & -- & -- \\
& \underline{41.80\%$^{*}$} &  &  &  &  &  \\[5pt]
    
\multirow{2}{*}{\textbf{DeepSeekMath-7B-Instruct}} & 32.40\%$^{*}$ & 5.33\%$^{*}$ & -- & -- & -- & -- \\
& \underline{7.40\%$^{*}$} &  &  &  &  &  \\[5pt]

\midrule 
\multirow{2}{*}{\textbf{o1}} & 71.54\%$^{\dagger}$ & 61.93\%$^{\dagger}$ & -- & -- & -- & -- \\
& \underline{49.86\%$^{\dagger}$} &  &  &  &  &  \\[5pt]

\multirow{2}{*}{\textbf{Claude 3.7 Sonnet}} & 44.62\%$^{\dagger}$ & 28.43\%$^{\dagger}$ & -- & -- & -- & -- \\
& \underline{22.94\%$^{\dagger}$} &  &  &  &  &  \\[5pt]
   
\midrule
\multirow{3}{*}{\textbf{Theoretical Random Baseline}} & 23.87\% & 0 & 20.01\% & 0 & 20.86\% & 0 \\
 & 25.00\%$^{*}$ & 0$^{*}$ & -- & -- & -- & -- \\
 & 21.68\%$^{\dagger}$ & 0$^{\dagger}$ & -- & -- & -- & -- \\
\bottomrule
\end{tabular}}
\caption{Accuracies of models across different question types, with \underline{underscored values} indicating the accuracy improvement over the theoretical accuracy of random guess for MCQs. Values marked with $^{*}$ indicate accuracies on text-only subsets. Values marked with $^{\dagger}$ indicate accuracies reported only on the text-image questions.}
\label{tab:acc_by_question_type}
\end{table}

\paragraph{Performance on Different Questions Types}
Table \ref{tab:acc_by_question_type} reflects the performance of models on MCQs and Q\&A questions in different subjects and the whole dataset, as well as the theoretical random-guess baselines. 
The random-guess baselines of MCQs are calculated based on the approximation that all MCQs in MMSciBench are 4-choice questions, as over \textbf{99\%} of MCQs in MMSciBench have 4 choices (see Table \ref{tab:choice_distribution} in the appendix for detailed statistics).
For single-choice questions, the random-guess accuracy is 1/4, as only one option is correct. 
For multiple-choice questions, where valid subsets include combinations of more than one choice, the random-guess accuracy is $1 / (C_4^2 + C_4^3 + C_4^4) = 1/11$. For indeterminate-choice questions, where any non-empty subset of choices is valid, the random-guess accuracy is $1 / 2^4 = 1/16$. These probabilities were weighted to compute random-guess baselines of MCQs.

While the raw accuracies suggest that models generally perform better on MCQs than on Q\&A questions, subtracting the baseline accuracies from their MCQ results reveals smaller yet positive gaps. This indicates that the provided answer choices in MCQs may assist the models by narrowing the possible answer space, making these questions easier to answer correctly compared to Q\&A questions. Interestingly, this pattern does not consistently hold true for math, where the MCQ advantage disappears after accounting for the baseline. In fact, some models seem to struggle more with MCQs than with Q\&A questions in this subject.
This suggests that the provided choices in math MCQs might mislead the models, making these questions more challenging.

\subsection{Taxonomy-Based Analysis}

\begin{figure*}[t!]
    \centering
    \begin{adjustbox}{center}
    \includegraphics[width=\linewidth]{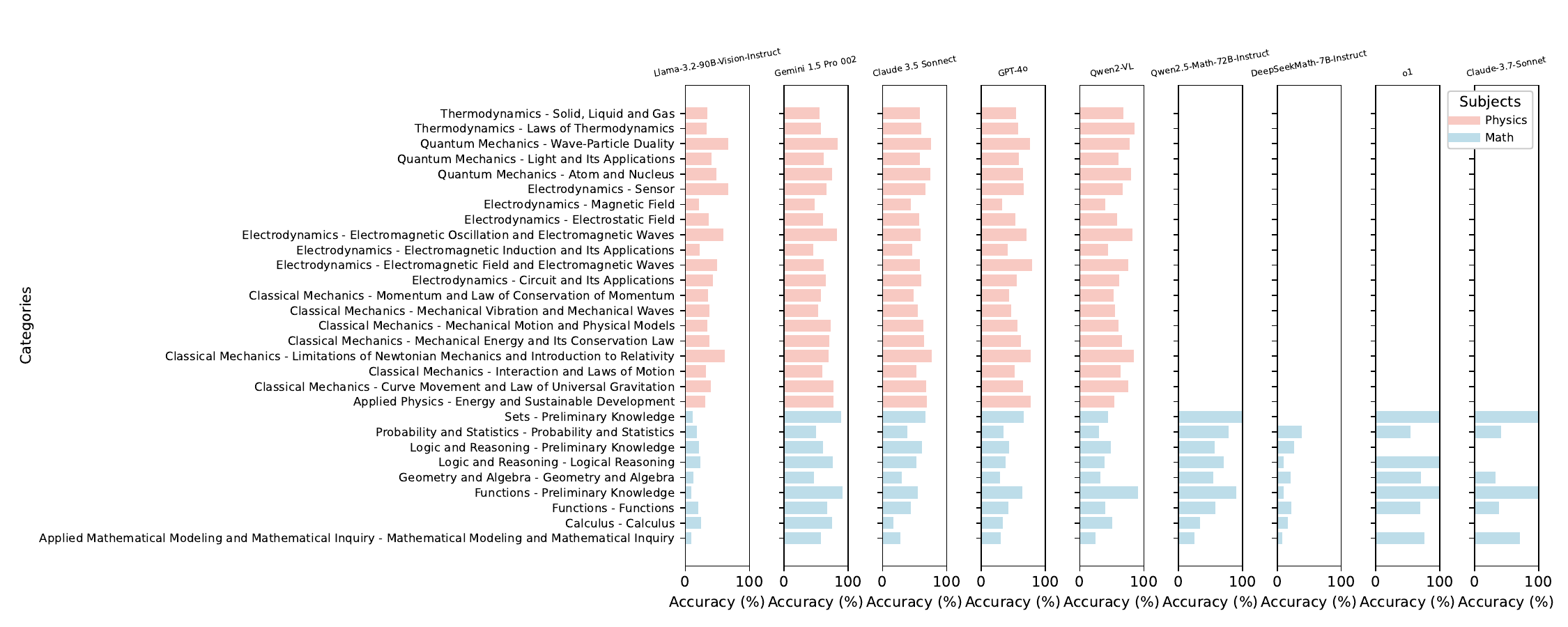}
    \end{adjustbox}
    \caption{Accuracies of models across different key knowledge points.}
    \label{fig:combined_accuracy_by_category}
\end{figure*}

To better understand where different models excel or struggle within scientific domains—and to identify inherently challenging key knowledge points—all models’ performances were analyzed across the taxonomy of first- and second-level key knowledge points, i.e., \textit{Domain} and \textit{Module} levels (see Fig. \ref{fig:combined_accuracy_by_category}). This analysis reveals that, while models generally maintain consistent relative rankings across entire subjects, their strengths can vary significantly at the subfield level.
For instance, although Gemini 1.5 Pro 002 often leads among non-reasoning models, it falls behind Claude 3.5 Sonnet and GPT-4o in the subfield of ``Electrodynamics - Magnetic Field''. Additionally, certain subfields prove universally challenging, e.g., ``Electrodynamics - Electromagnetic Induction and Its Applications'' in physics, as well as ``Geometry and Algebra – Geometry and Algebra'' and ``Functions – Preliminary Knowledge'' in mathematics.
These findings highlight both the nuanced capabilities and the current limitations of state-of-the-art models in addressing scientific knowledge.

\subsection{Visual Understanding}
\label{sec:visual_understanding}

\begin{table}[t!]
\centering
 \resizebox{0.5\textwidth}{!}{
\begin{tabular}{lcccccc}
\toprule
\multirow{2}{*}{\textbf{Models}} & \multicolumn{2}{c}{\textbf{Math}} & \multicolumn{2}{c}{\textbf{Physics}} & \multicolumn{2}{c}{\textbf{Overall}} \\
\cmidrule(lr){2-3} \cmidrule(lr){4-5} \cmidrule(lr){6-7}
 & \textbf{Text} & \textbf{T\&I} & \textbf{Text} & \textbf{T\&I} & \textbf{Text} & \textbf{T\&I} \\
\midrule
\textbf{Llama-3.2-90B-Vision-Instruct} & 19.54\% & 11.60\% & 42.83\% & 16.34\% & 37.07\% & 14.48\% \\
\textbf{Gemini 1.5 Pro 002} & 69.60\% & 33.70\% & 74.40\% & 39.01\% & 73.21\% & 36.93\% \\
\textbf{Claude 3.5 Sonnet} & 44.57\% & 24.51\% & 67.75\% & 35.21\% & 62.02\% & 31.02\% \\
\textbf{GPT-4o} & 44.69\% & 20.35\% & 64.10\% & 31.55\% & 59.31\% & 27.16\% \\
\textbf{Qwen2-VL-72B-Instruct} & 41.39\% & 24.95\% & 72.48\% & 35.63\% & 64.80\% & 31.45\% \\
\midrule
\textbf{Qwen2.5-Math-72B-Instruct} & 57.39\% & -- & -- & -- & -- & -- \\
\textbf{DeepSeekMath-7B-Instruct} & 21.86\% & -- & -- & -- & -- & -- \\
\midrule
\textbf{o1} & -- & 67.40\% & -- & -- & -- & -- \\
\textbf{Claude 3.7 Sonnet} & -- & 37.64\% & -- & -- & -- & -- \\
\bottomrule
\end{tabular}}
\caption{Accuracies of models on text-only (\textbf{Text}) and text-image paired (\textbf{T\&I}) questions across different subjects.}
\label{tab:acc_w_wo_img}
\end{table}

MMSciBench includes both text-only and text-image paired questions. To evaluate the impact of visual input, we assess models on both types of questions, as shown in Table \ref{tab:acc_w_wo_img}. Notably, all LVLMs perform worse on tasks involving both textual and visual elements compared to those relying solely on text. This highlights that bridging the gap between text comprehension and text-image co-reasoning remains a significant challenge for current LVLMs. Furthermore, the higher proportion of text-only questions in physics partially explains why models perform better on physics questions compared to math questions, as observed in Table \ref{tab:overall_acc}.

\subsection{The Effect of Chain-of-Thought in Reasoning}

To evaluate the full scientific potential of the models, we design a suite of prompts to instruct them to answer step-by-step in Chinese, as detailed in Sec. \ref{app:prompt_templates_effect_cot_reasoning} in the appendix. As shown in Table \ref{tab:acc_step_by_step}, step-by-step prompting improves the accuracies of Llama-3.2-90B-Vision-Instruct, DeepSeekMath-7B-Instruct, o1, and Claude 3.7 Sonnet compared to their results in Table \ref{tab:overall_acc}. However, the accuracy of Qwen2.5-Math-72B-Instruct and Qwen2-VL-72B-Instruct decreases, while the performance of the other models remains unchanged.

This observation suggests that explicitly prompting certain models to use chain-of-thought reasoning can enhance their performance, and that different models exhibit varying degrees of alignment or readiness in this regard. Notably, Gemini 1.5 Pro 002, Claude 3.5 Sonnet, GPT-4o, Qwen2.5-Math-72B-Instruct, and Qwen2-VL-72B-Instruct are more capable of generating effective reasoning steps without explicit prompting, whereas other models show more significant improvements when guided explicitly.

Considering that models typically have access to richer English training resources, we conducted additional experiments by prompting them to answer step-by-step in English to further explore their scientific capabilities. The corresponding prompts are detailed in Sec. \ref{app:prompt_templates_effect_cot_reasoning} of the appendix. As shown in Table \ref{tab:acc_step_by_step}, the results indicate that models, except Gemini 1.5 Pro 002 and Claude 3.7 Sonnet, benefit from this instruction. This underscores the effectiveness of explicit chain-of-thought prompting and its importance in accurately assessing models’ capabilities. The differing behavior of the two models may suggest that their performance relies on the compatibility between the language of the questions and the language of the answers.

\begin{table}[t!]
\centering
 \resizebox{0.5\textwidth}{!}{
\begin{tabular}{l lccc}
\toprule
& \textbf{Models} & \textbf{Math} & \textbf{Physics} & \textbf{Overall} \\
\midrule
\multirow{6}{*}{\begin{tabular}{@{}l@{}}\textbf{}\\\textbf{in Chinese}\\\textbf{}\end{tabular}} 
 & \textbf{Llama-3.2-90B-Vision-Instruct} & 19.12\% & 38.86\% & 33.24\% \\
 & \textbf{Gemini 1.5 Pro 002} & 56.90\% & 66.28\% & 63.61\% \\
 & \textbf{Claude 3.5 Sonnet} & 36.83\% & 61.42\% & 54.42\% \\
 & \textbf{GPT-4o} & 35.74\% & 56.86\% & 50.85\% \\
 & \textbf{Qwen2-VL-72B-Instruct} & 30.72\% & 57.77\% & 50.07\% \\
 & \textbf{Qwen2.5-Math-72B-Instruct} & 55.68\%$^{*}$ & -- & -- \\
 & \textbf{DeepSeekMath-7B-Instruct} & 23.32\%$^{*}$ & -- & -- \\
 & \textbf{o1} & 68.05\%$^{\dagger}$ & -- & -- \\
& \textbf{Claude 3.7 Sonnet} & 39.61\%$^{\dagger}$ & -- & -- \\[6pt]

\midrule

\multirow{6}{*}{\begin{tabular}{@{}l@{}}\textbf{}\\\textbf{in English}\\\textbf{}\end{tabular}} 
 & \textbf{Llama-3.2-90B-Vision-Instruct} & 22.41\% & 44.20\% & 38.00\% \\
 & \textbf{Gemini 1.5 Pro 002} & 55.17\% & 65.07\% & 62.25\% \\
 & \textbf{Claude 3.5 Sonnet} & 40.67\% & 61.26\% & 55.40\% \\
 & \textbf{GPT-4o} & 37.23\% & 59.08\% & 52.86\% \\
 & \textbf{Qwen2-VL-72B-Instruct} & 32.68\% & 60.79\% & 52.79\% \\
 & \textbf{Qwen2.5-Math-72B-Instruct} & 55.31\%$^{*}$ & -- & -- \\
 & \textbf{DeepSeekMath-7B-Instruct} & 23.69\%$^{*}$ & -- & -- \\
 & \textbf{o1} & 68.49\%$^{\dagger}$ & -- & -- \\
& \textbf{Claude 3.7 Sonnet} & 36.32\%$^{\dagger}$ & -- & -- \\

\bottomrule
\end{tabular}}
\caption{Accuracies of models asked to provide step-by-step answers in Chinese and English. Values marked with $^{*}$ indicate accuracies on text-only math questions. Values marked with $^{\dagger}$ indicate accuracies reported only on the text-image math questions.}
\label{tab:acc_step_by_step}
\end{table}

\subsection{Error Analysis}
To further understand the limitations of the evaluated models, we conducted an in-depth error analysis on questions where all models produced incorrect answers. This analysis leveraged the detailed explanations provided within our dataset to identify specific error patterns.

For this analysis, we first isolated the subset of questions where all models failed. From this subset, we carefully selected \textbf{40} questions, ensuring a stratified distribution across different subjects (math and physics), question types (MCQs and Q\&A), and modalities (text-only and text-image). This selection resulted in \textbf{5} questions for each combination, leading to a total of \textbf{240} individual cases examined (considering 5 models on the entire dataset, 2 math LLMs on text-only math questions, and 2 reasoning models on text-image math questions).

We classified the identified errors into five distinct categories: (1) \textbf{Visual Misinterpretation}, where models failed to correctly interpret visual information; (2) \textbf{Textual Misunderstanding}, indicating an incorrect grasp of the textual content; (3) \textbf{Reasoning Error}, reflecting flaws in the logical deduction process; (4) \textbf{Integration Failure}, characterized by poor synthesis of information from both text and image modalities; and (5) \textbf{Calculation Error}, pertaining to inaccuracies in numerical computations. Errors were identified by meticulously comparing model responses with the ground-truth explanations in our dataset. 

The error analysis reveals that \textbf{Reasoning Error} is overwhelmingly the most common challenge for the models, accounting for an average of \textbf{77.1\%} of all incorrect answers. This highlights a significant limitation in the logical and inferential capabilities of current models when tackling scientific problems. While Calculation Error (\textbf{11.3\%}) is the second most frequent, it is substantially lower. Notably, Visual Misinterpretation (\textbf{7.5\%}), Textual Misunderstanding (\textbf{1.7\%}), and Integration Failure (\textbf{2.5\%}) occur less frequently on average, although certain models show particular weaknesses in these areas (e.g., o1 and Claude-3.7-Sonnet exhibiting higher Visual Misinterpretation). The prevalence of reasoning errors underscores the need for future research to focus on improving the complex multi-step reasoning abilities of AI systems for scientific problem-solving. The detailed distribution of these error types across the evaluated models is presented in Table \ref{tab:error_distribution} in the appendix.

\section{Related Work}

\paragraph{Scientific Benchmarks}
Scientific benchmarks are essential tools for evaluating the capabilities of language models in understanding and reasoning about complex scientific concepts, encompassing a wide range of disciplines, from general science to domain-specific areas like mathematics and physics. General scientific benchmarks, such as MSVEC \cite{evans2023msvec} and SciOL \cite{tarsi2024sciol}, have been developed to assess various aspects of language models’ abilities in specific scientific domains, including claim verification, figure retrieval, and multimodal information comprehension. However, the increasing complexity of language models necessitates more and the push towards more advanced scientific reasoning \cite{yan2025position} specialized benchmarks to evaluate their performance in specific scientific domains.

In mathematics, benchmarks like TRIGO \cite{xiong2023trigo} (formal proof reduction), DrawEduMath \cite{baral2025drawedumath} (visual math problems), and DMath \cite{kim2023ain} (math word problems) have been developed to assess AI models on targeted mathematical tasks. The landscape of mathematical reasoning benchmarks and methodologies, especially in the context of MLLMs, is rapidly expanding, as surveyed by \citet{yan2024survey}.
Similarly, in physics, datasets such as GRASP \cite{jassim2023grasp} have been introduced to assess models' understanding of ``Intuitive Physics'' principles, including object permanence and continuity.

Additionally, benchmarks like GAOKAO-Bench \cite{zhang2023evaluating}, GAOKAO-MM \cite{zong2024gaokao}, OlympiadBench \cite{he2024olympiadbench}, SciBench \cite{wangscibench}, SciEval \cite{sun2024scieval}, MMSci \cite{li2024mmsci}, SceMQA \cite{liang2024scemqa}, and SciFIBench \cite{roberts2024scifibench} span multiple scientific domains, such as mathematics, physics, chemistry, and biology. These benchmarks focus on high-school, Olympiad, pre-college, PhD, and academic levels.
Broadening the scope further, M3Exam \cite{zhang2023m3exam} and EXAMS-V \cite{das2024exams} are large-scale, multilingual, and multimodal benchmarks derived from real human exam questions across various countries and educational levels, including scientific subjects. EXAMS-V, for instance, uniquely embeds question text and visual elements into a single image, demanding integrated reasoning. These exam-based benchmarks test not only subject knowledge but also cultural and region-specific understanding.

\paragraph{Benchmarks for LVLMs} 
Benchmarks for LVLMs have been developed to evaluate their performance across various tasks, including visual question answering, image captioning, and multimodal reasoning. These benchmarks typically consist of datasets with image-text pairs accompanied by corresponding questions or instructions, assessing the ability of LVLMs to generate accurate and relevant responses.
For example, the VALSE benchmark \cite{parcalabescu2021valse} focuses on evaluating the visio-linguistic grounding capabilities of pretrained VLMs on specific linguistic phenomena. 
MMMU \cite{yue2024mmmu} and MMMU-Pro \cite{yue2024mmmupro} assess multimodal models on massive multi-discipline tasks requiring college-level subject knowledge and deliberate reasoning. 
Other benchmarks, such as VisIT-Bench \cite{bitton2023visit}, WinoGAViL \cite{bitton2022winogavil}, and those designed for zero-shot visual reasoning \cite{nagar2024zero, xu2024benchmarking}, are aimed at assessing the ability of LVLMs to reason about visual scenes and answer questions that require minimal world knowledge. These benchmarks often analyze the impact of conveying scene information either as visual embeddings or as purely textual scene descriptions to the underlying LLM of the LVLM.
The evaluation of state-of-the-art models like GPT-4V on structured reasoning tasks has also begun, with studies such as \citet{singh2023assessing} assessing performance on mathematical reasoning with visual context and visual data analysis, highlighting both capabilities and ongoing challenges. The broader trends and challenges in LVLMs reasoning abilities are further explored in surveys like \citet{wang2024exploring}.

To address the scarcity of scientific benchmarks specifically designed for the high school level—supporting both text-only and multimodal reasoning—we introduce MMSciBench. As detailed in Table \ref{tab:comparison}, this dataset achieves a balanced trade-off between size and comprehensiveness, enabling efficient evaluation while offering a diverse selection of challenging high-school-level scientific problems. Additionally, MMSciBench prioritizes quality, with a significant portion of problems including detailed solution explanations and a three-level taxonomy of key knowledge points, facilitating fine-grained analysis of AI model performance.

\section{Conclusion}
This paper introduces MMSciBench, a benchmark designed to evaluate the scientific capabilities of unimodal and multimodal language models. MMSciBench consists of a collection of high school-level MCQs and Q\&A in mathematics and physics, with a subset of the questions incorporating images. The benchmark organizes its questions into a three-level taxonomy, ensuring comprehensive coverage of key knowledge points in both subjects.
Our evaluation of five advanced LVLMs and two specialized math LLMs on MMSciBench demonstrates that current models still have significant room for improvement in scientific problem-solving. The analysis highlights that the inclusion of visual elements in questions presents a substantial challenge for model performance, emphasizing the complexity of utilizing textual and visual reasoning.
This work contributes to the ongoing development of robust benchmarks aimed at evaluating the evolving capabilities of language models, particularly in the domain of scientific reasoning.

\section*{Limitations}
Despite the advances presented in MMSciBench, several limitations warrant discussion and open avenues for future research.

\begin{enumerate}
\item \textbf{Domain and Content Scope:}
MMSciBench is focused on high-school level mathematics and physics, a scope chosen for its educational relevance and well-defined problem sets. However, this focus also limits the benchmark’s applicability to broader scientific domains. While the curated questions capture essential concepts, they do not encompass other fields such as chemistry, biology, or advanced scientific topics. Additionally, the dataset’s reliance on K–12 educational standards may introduce biases that do not reflect the diverse challenges encountered in higher-level or interdisciplinary scientific reasoning.

\item \textbf{Evaluation Metrics and Reasoning Transparency:}
The evaluation framework is centered on final answer accuracy, a metric that, while objective, does not capture the nuances of intermediate reasoning steps or the quality of explanations generated by models. By discounting partial correctness or the reasoning process, the assessment may obscure important differences in how models arrive at their answers. Future iterations of the benchmark may benefit from incorporating multi-faceted evaluation criteria that assess both the correctness of conclusions and the soundness of the reasoning process.

\item \textbf{Language and Cultural Considerations:}
MMSciBench is primarily composed in Chinese, with some experiments extended to English. Models predominantly trained on English data may therefore be disadvantaged, and cultural or linguistic biases could affect performance. Future work should consider expanding the benchmark to include a more balanced representation of languages and educational contexts.

\item \textbf{Dataset Size and Filtering Practices:}
While MMSciBench comprises \textbf{4,482} question–solution pairs, the dataset size is modest relative to some large-scale benchmarks. The strict filtering criteria (e.g., including only questions with a human-annotated hardness score $\ge$ 0.7) may also limit the diversity of problem difficulties, potentially excluding edge cases that could be valuable for assessing nuanced reasoning. Enlarging the dataset and diversifying the difficulty distribution would further strengthen the benchmark’s comprehensiveness.

\item \textbf{Limitations of GPT-4o as a judge:}
Despite achieving a \textbf{97.22\%} agreement rate between GPT-4o and human evaluations through iterative refinement, potential biases and limitations persist. The automated evaluation framework may inherit inherent subjectivity in scoring criteria or undetected systematic biases. Future work may incorporate hybrid human-AI evaluation protocols to further mitigate these limitations. Furthermore, while GPT-4o currently achieves a high agreement with human judgments, its effectiveness may decline as models outpace its abilities. We plan periodic updates to incorporate state-of-the-art models, ensuring the benchmark’s robustness and relevance.
\end{enumerate}


\bibliography{custom}


\appendix


\clearpage

\section{Dataset Curation Process}
\label{app:dataset_curation_process}
The dataset of MMSciBench was annotated by a team of 12 experienced high school teachers, each with at least 5 years of teaching experience in mathematics or physics, ensuring domain expertise.
Each question was independently annotated by at least 3 teachers for difficulty level, which is defined on a standardized scale from 0 (easiest) to 1 (most challenging), solutions, and explanations to minimize individual bias.
We measured inter-annotator agreement using Cohen’s kappa, yielding scores of 0.82 for difficulty for substantial agreement and supporting the reliability of the annotations. Disagreements were resolved by consulting a senior teacher or teachers' instructors to reach a consensus. The difficulty levels were further enhanced and adjusted to student test scores based on these problem sets at the proper grades.

\clearpage
\section{Prompts}
\label{app:prompts}
In this section, we present the prompts used in our work.

\subsection{The Prompt for Question Categorization}
\label{app:prompt_for_question_categorization}
\begin{figure}[!htbp]
    \centering
    \begin{adjustbox}{center}
    \includegraphics[width=\linewidth]{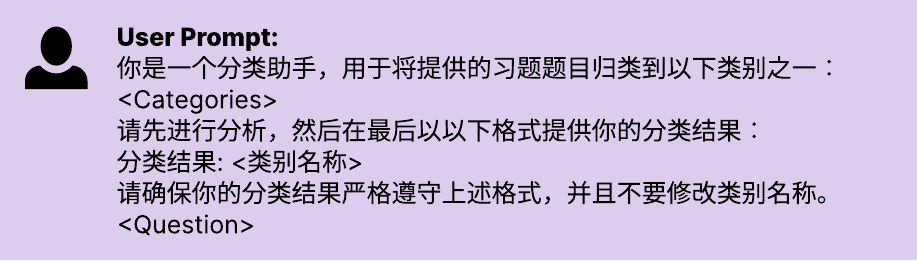}
    \end{adjustbox}
    \caption{The prompt template is designed to use GPT-4o as a classifier, categorizing each question into a three-level hierarchy. <Categories> represents the predefined set of categories for the target subject.}
    \label{fig:prompt_classify}
\end{figure}

Fig. \ref{fig:prompt_classify} presents the prompt designed for categorizing MMSciBench questions into specific categories using GPT-4o. The category sets for each subject are derived from a Chinese high school key knowledge point taxonomy.

\subsection{Prompt Templates for the Effect of Chain-of-Thought in Reasoning}
\label{app:prompt_templates_effect_cot_reasoning}
\begin{figure}[!htbp]
    \centering
    \begin{adjustbox}{center}
    \includegraphics[width=\linewidth]{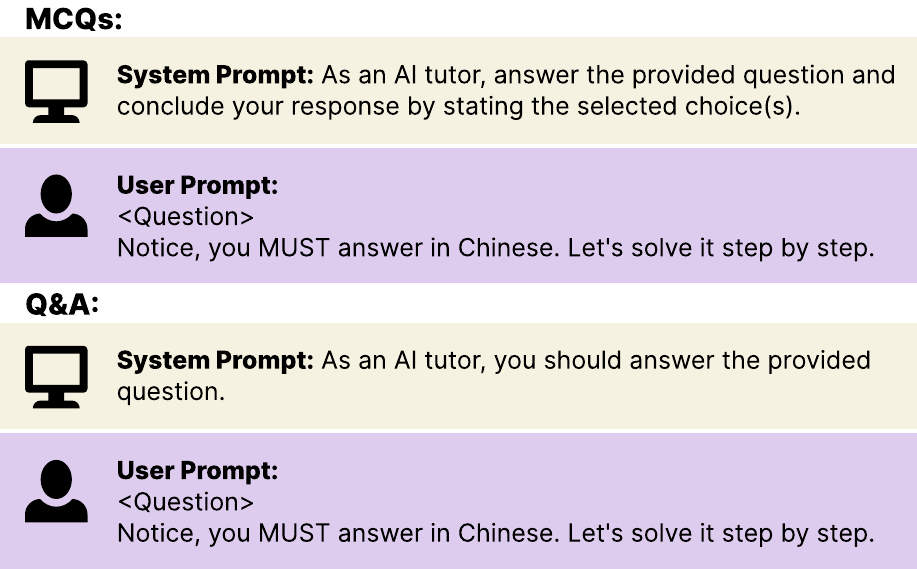}
    \end{adjustbox}
    \caption{The prompt template is designed for requesting models to answer questions in Chinese step by step, where the <Question> is sourced from MMSciBench.}
    \label{fig:prompt_chn_stp}
\end{figure}

\begin{figure}[!htbp]
    \centering
    \begin{adjustbox}{center}
    \includegraphics[width=\linewidth]{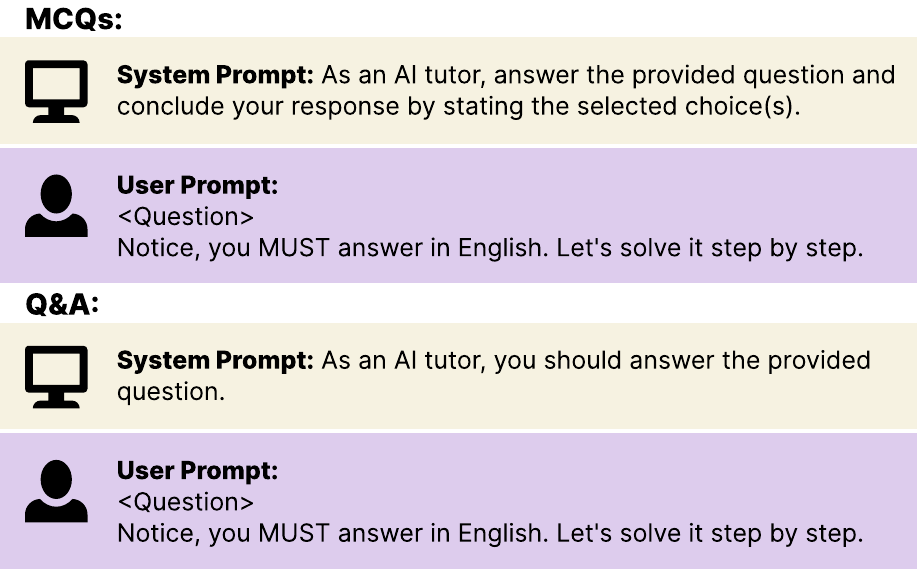}
    \end{adjustbox}
    \caption{The prompt template is designed for requesting models to answer questions in English step by step, where the <Question> is sourced from MMSciBench.}
    \label{fig:prompt_eng_stp}
\end{figure}

Fig. \ref{fig:prompt_chn_stp} and Fig. \ref{fig:prompt_eng_stp} are prompts templates that ask models to think step by step in Chinese and English, respectively. 

\subsection{The Prompt Template for Using GPT-4o as a Judge}
\label{app:prompt_gpt4o_as_judge}
\begin{figure}[!htbp]
    \centering
    \begin{adjustbox}{center}
    \includegraphics[width=\linewidth]{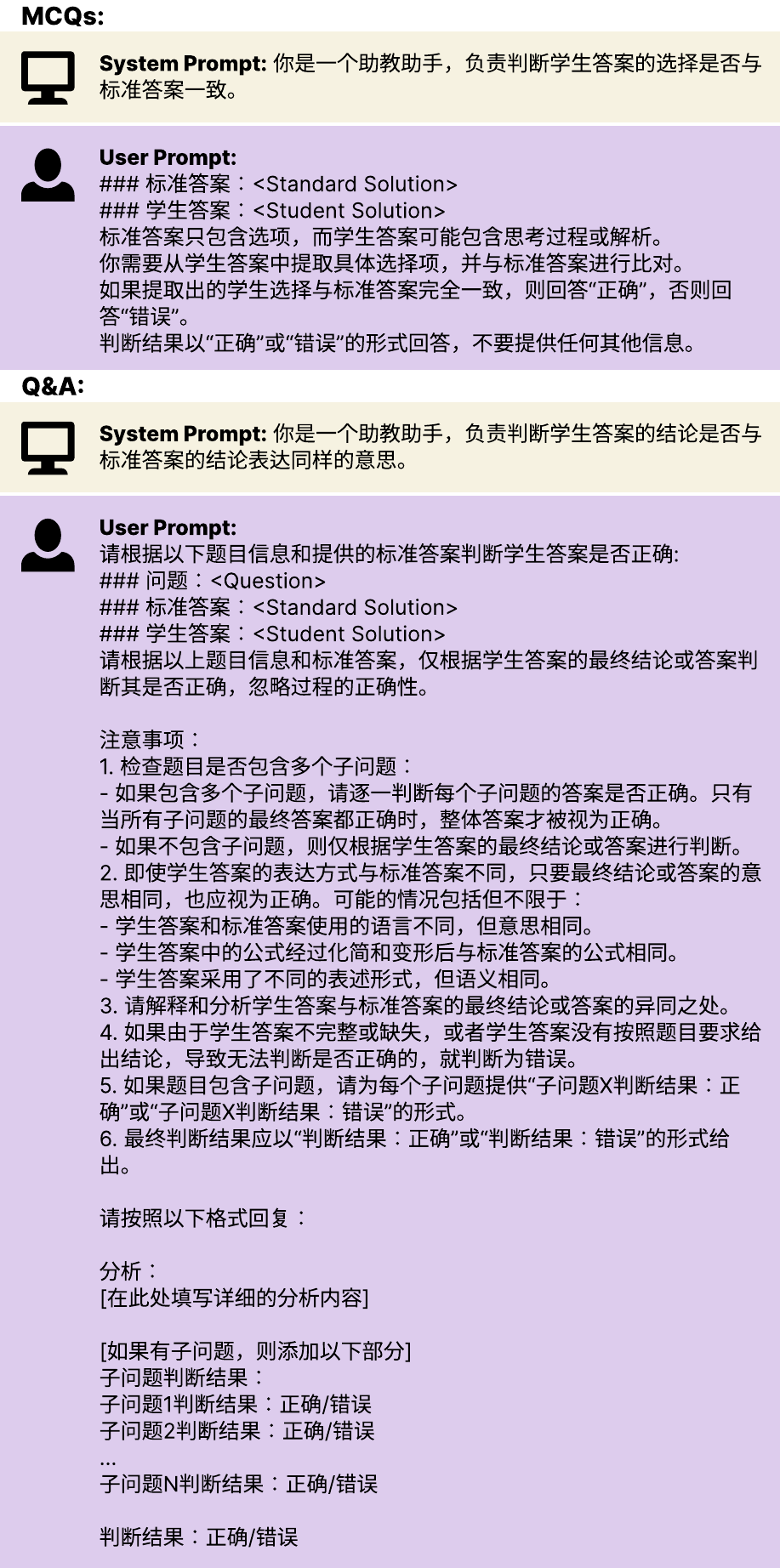}
    \end{adjustbox}
    \caption{The prompt template designed for using GPT-4o as a judge, where the <Question> and <Standard Solution> is sourced from MMSciBench, while <Student Solution> is the solution provided by the tested model.}
    \label{fig:prompt_judge}
\end{figure}

\begin{figure}[!htbp]
    \centering
    \begin{adjustbox}{center}
    \includegraphics[width=\linewidth]{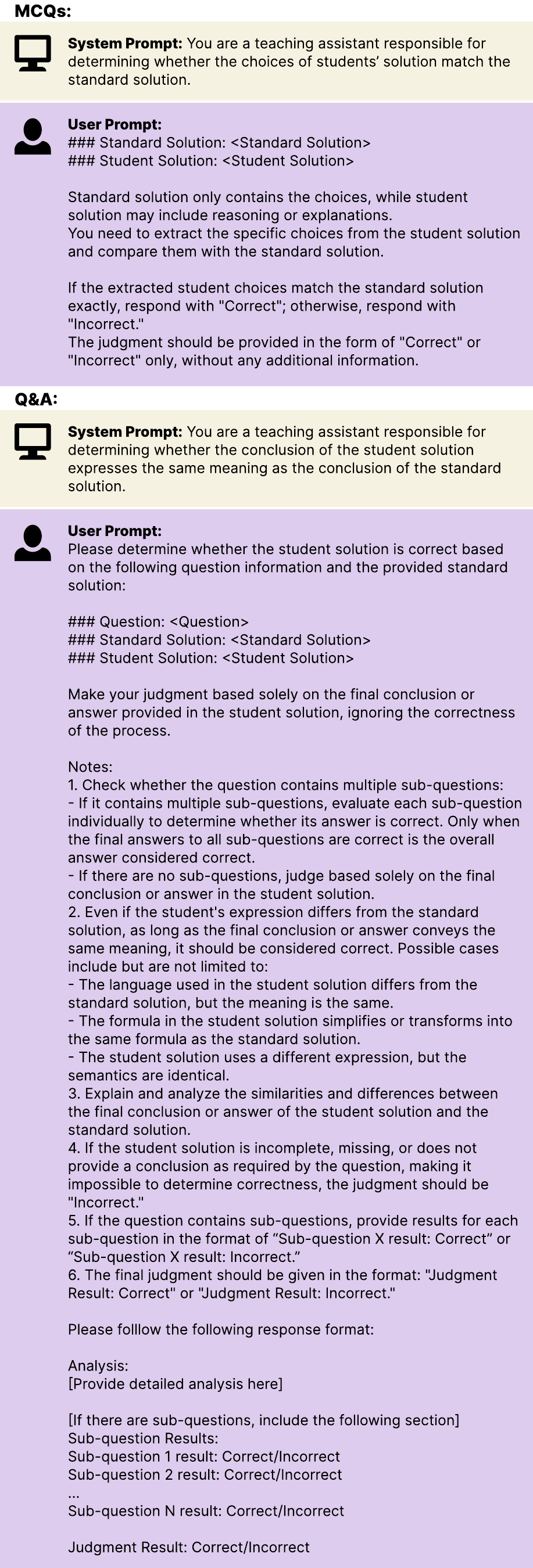}
    \end{adjustbox}
    \caption{The English translation of the prompt template shown in Fig. \ref{fig:prompt_judge}.}
    \label{fig:prompt_judge_en}
\end{figure}

Fig. \ref{fig:prompt_judge} (with its English translation in Fig. \ref{fig:prompt_judge_en}) illustrates the prompt used to instruct GPT-4o to evaluate whether a ``student solution''—that is, the model’s response being assessed—is correct or incorrect compared to the standard solution in MMSciBench.
For MCQs, only the model’s answer and the standard solution are provided, omitting the actual questions. This approach is sufficient because the evaluation solely involves comparing whether the selected choices match the standard answer, eliminating the need to understand the question’s context.
In contrast, for Q\&A questions, GPT-4o is provided with the question, the standard solution, and the model’s answer. This comprehensive context enables accurate semantic understanding and a thorough comparison between the two responses.
The prompt for Q\&A questions have been iteratively refined and enhanced to improve GPT-4o’s judgment, particularly in cases where misjudgments are likely. This refinement process involves sampling a subset of evaluated responses and manually diagnosing the reasons for any misjudgments, thereby continually improving the evaluation accuracy.

\clearpage

\section{Data Examples}
\label{app:data_examples}
\begin{figure}[!htbp]
    \small
    \centering
    \begin{tcolorbox}[colframe=magenta!50, colback=magenta!10, title=Question \& Standard Solution]
        \textbf{Question}

        问题（单选）：如图所示，两块同样的玻璃直角三棱镜$ABC$，两者的$AC$面是平行放置的，在它们之间是均匀的未知透明介质。一束单色细光$O$垂直于$AB$面入射，在图示的出射光线中（\quad）。 

        \includegraphics[width=\textwidth]{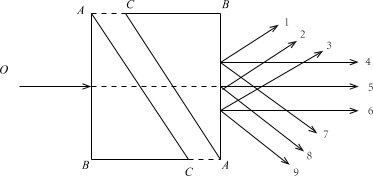}
        
        选项：
        
        A. 1、2、3（彼此平行）中的任一条都有可能
        
        B. 4、5、6（彼此平行）中的任一条都有可能
        
        C. 7、8、9（彼此平行）中的任一条都有可能
        
        D. 只能是4、6中的某一条

        \textbf{Difficulty Level}: 0.7

        \textbf{Domain}: Quantum Mechanics

        \textbf{Module}: Light and Its Applications

        \textbf{Chapter}: Snell's Law

        \textbf{Standard Solution}: B
    \end{tcolorbox}

    
    \begin{tcolorbox}[colframe=cyan!50, colback=cyan!10, title=Explanation]
        本题主要考查三棱镜问题。
        
        选项分析：据题述，两个直角三棱镜之间的介质折射率未知，可能比玻璃大，可能与玻璃相同，也可能比玻璃小，可能的光路图如下：

        \includegraphics[width=\textwidth]{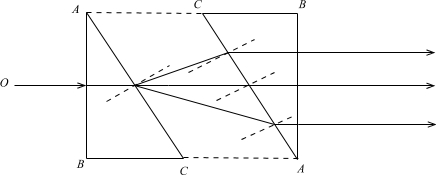}
        
        故B项正确，ACD项错误。
        
        综上所述，本题正确答案为B。
    \end{tcolorbox}
    
    \caption{An example of a physics MCQ.}
    \label{fig:physics_mcq}
\end{figure}

\begin{figure}[!htbp]
    \small
    \centering
    \begin{tcolorbox}[colframe=magenta!50, colback=magenta!10, title=Question \& Standard Solution]
        \textbf{Question}

        问题（多选）：下图是函数$y=\sin(\omega x+\varphi )$的部分图象，则$\sin(\omega x+\varphi)=$（  ）。

        \includegraphics[width=\textwidth]{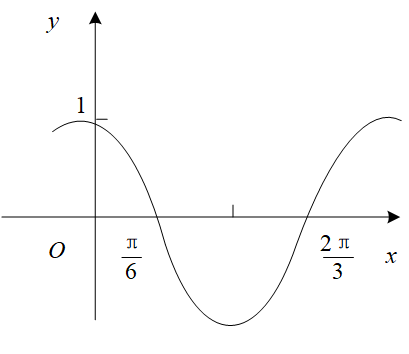}
        
        选项：
        
        A. $\sin(x+{\pi \over 3})$ 
        
        B. $\sin({\pi \over 3}-2x)$
        
        C. $\cos(2x+{\pi \over 6})$ 
        
        D. $\cos({5\pi \over 6}-2x)$

        \textbf{Difficulty Level}: 0.7

        \textbf{Domain}: Functions

        \textbf{Module}: Functions

        \textbf{Chapter}: Trigonometric Functions

        \textbf{Standard Solution}: B, C
    \end{tcolorbox}

    
    \begin{tcolorbox}[colframe=cyan!50, colback=cyan!10, title=Explanation]
        本题主要考查三角函数。
        
        由题图可知，
        \[
        \displaystyle \frac{T}{2} = \frac{2}{3} \pi - \frac{\pi}{6} = \frac{\pi}{2},
        \]
        所以
        \[
        T = \frac{2\pi}{|\omega|} = \pi,
        \]
        所以$|\omega| = 2$。
        
        当$\omega = 2$时，由函数图象过点$\left(\frac{\pi}{6}, 0\right)$，$\left(\frac{2\pi}{3}, 0\right)$，
        且$f(0) > 0$，得
        \[
        \varphi = \frac{2\pi}{3} + 2k\pi \quad (k \in \mathbb{Z}),
        \]
        所以
        \[
        y = \sin\left(2x + \frac{2\pi}{3}\right) = -\cos\left(\frac{5\pi}{6} - 2x\right),
        \]
        同理，当$\omega = -2$时，
        \[
        \varphi = \frac{\pi}{3} + 2k\pi \quad (k \in \mathbb{Z}),
        \]
        所以
        \[
        y = \sin\left(-2x + \frac{\pi}{3}\right) = \cos\left(2x + \frac{\pi}{6}\right)
        \]
        
        故本题正确答案为 BC。
    \end{tcolorbox}
    
    \caption{An example of a math MCQ.}
    \label{fig:math_mcq}
\end{figure}

\begin{figure}[!htbp]
    \small
    \centering
    \begin{tcolorbox}[colframe=magenta!50, colback=magenta!10, title=Question \& Standard Solution]
        \textbf{Question}

        Question (Multiple Choice): The figure below shows a part of the graph of the function \( y = \sin(\omega x + \varphi) \). Determine \( \sin(\omega x + \varphi) = \) ( ).

        \includegraphics[width=\textwidth]{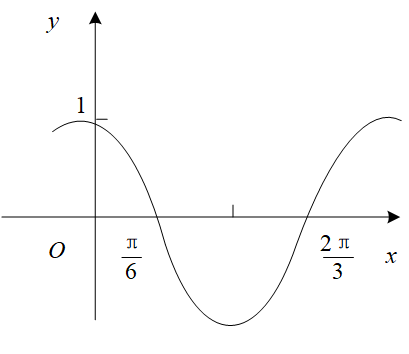}
        
        Options:
        
        A. \( \sin\left(x + \frac{\pi}{3}\right) \) 
        
        B. \( \sin\left(\frac{\pi}{3} - 2x\right) \)
        
        C. \( \cos\left(2x + \frac{\pi}{6}\right) \) 
        
        D. \( \cos\left(\frac{5\pi}{6} - 2x\right) \)

        \textbf{Difficulty Level}: 0.7

        \textbf{Domain}: Functions

        \textbf{Module}: Functions

        \textbf{Chapter}: Trigonometric Functions

        \textbf{Standard Solution}: B, C
    \end{tcolorbox}

    
    \begin{tcolorbox}[colframe=cyan!50, colback=cyan!10, title=Explanation]
        This question primarily assesses trigonometric functions.
        
        From the figure, we know that
        \[
        \frac{T}{2} = \frac{2}{3}\pi - \frac{\pi}{6} = \frac{\pi}{2},
        \]
        therefore
        \[
        T = \frac{2\pi}{|\omega|} = \pi,
        \]
        so \( |\omega| = 2 \).
        
        When \( \omega = 2 \), since the graph passes through the points \( \left(\frac{\pi}{6}, 0\right) \) and \( \left(\frac{2\pi}{3}, 0\right) \),
        and \( f(0) > 0 \), we have
        \[
        \varphi = \frac{2\pi}{3} + 2k\pi \quad (k \in \mathbb{Z}),
        \]
        thus
        \[
        y = \sin\left(2x + \frac{2\pi}{3}\right) = -\cos\left(\frac{5\pi}{6} - 2x\right),
        \]
        similarly, when \( \omega = -2 \),
        \[
        \varphi = \frac{\pi}{3} + 2k\pi \quad (k \in \mathbb{Z}),
        \]
        so
        \[
        y = \sin\left(-2x + \frac{\pi}{3}\right) = \cos\left(2x + \frac{\pi}{6}\right)
        \]
        
        Therefore, the correct answer is BC.
    \end{tcolorbox}
    
    \caption{The English translation of the math MCQ example in Fig. \ref{fig:math_mcq}.}
    \label{fig:math_mcq_en}
\end{figure}

\begin{figure}[!htbp]
    \small
    \centering
    \begin{tcolorbox}[colframe=magenta!50, colback=magenta!10, title=Question \& Standard Solution]
        \textbf{Question}

        问题（解答）：如图，建立平面直角坐标系$xOy$，$x$轴在地平面上，$y$轴垂直于地平面，单位长度为$1$千米。某炮位于坐标原点。已知炮弹发射后的轨迹在方程
        $$y=kx-\frac1{20}(1+k^2)x^2(k>0)$$
        表示的曲线上，其中$k$与发射方向有关。炮的射程是指炮弹落地点的横坐标。（1）求炮的最大射程；（2）设在第一象限有一飞行物（忽略其大小），其飞行高度为$3.2$千米，试问它的横坐标$a$不超过多少时，炮弹可以击中它？请说明理由。

        \includegraphics[width=\textwidth]{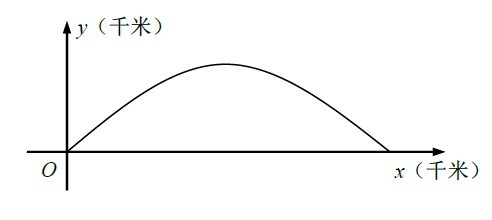}

        \textbf{Difficulty Level}: 0.7

        \textbf{Domain}: Geometry and Algebra

        \textbf{Module}: Geometry and Algebra

        \textbf{Chapter}: Plane Analytic Geometry

        \textbf{Standard Solution}

        （1）令$y=0$，得$kx-\frac1{20}(1+k^2)x^2=0$，由实际意义和题设条件知$x>0$，$k>0$，故
        $$x=\frac{20k}{1+k^2}=\frac{20}{k+\frac1k}\le\frac{20}2=10,$$
        当且仅当$k=1$时取等号。所以炮的最大射程为$10$千米。
        
        （2）因为$a>0$，所以炮弹可击中目标
        
        $\Leftrightarrow$存在$k>0$，使$3.2=ka-\frac1{20}(1+k^2)a^2$成立
        
        $\Leftrightarrow$关于$k$的方程$a^2k^2-20ak+a^2+64=0$有正根
        
        $\Leftrightarrow$判别式
        $$\Delta=(-20a)^2-4a^2(a^2+64)\ge0$$
        
        $\Leftrightarrow a\le6$
        
        此时，
        $$k=\frac{20a+\sqrt{(-20a)^2-4a^2(a^2+64)}}{2a^2}>0$$
        （不考虑另一根）。所以当$a$不超过$6$千米时，可击中目标。
    \end{tcolorbox}

    
    \begin{tcolorbox}[colframe=cyan!50, colback=cyan!10, title=Explanation]
        本题主要考查函数与方程和基本不等式的应用等相关知识。（1）求炮的最大射程，即$y=0$时的一个较大的根，因为含有参数$k$，所以需根据$k$的取值范围确定另外一个根的最大值，即为炮的最大射程。（2）炮弹能击中目标的含义为炮弹的飞行高度$y=3.2$时有解。根据二次函数有正根，可得出$a$的取值范围。
    \end{tcolorbox}
    
    \caption{An example of a math Q\&A question.}
    \label{fig:math_qa}
\end{figure}

\begin{figure}[!htbp]
    \small
    \centering
    \begin{tcolorbox}[colframe=magenta!50, colback=magenta!10, title=Question \& Standard Solution]
        \textbf{Question}

        Question (Q\&A): As shown in the figure, set up a Cartesian coordinate system $xOy$, with the $x$-axis on the ground, the $y$-axis perpendicular to the ground, and the unit length is 1 kilometer. A cannon is located at the origin. It is known that the trajectory of the cannonball after firing is represented by the equation
        $$y = kx - \frac{1}{20}(1 + k^2)x^2 (k > 0)$$
        where $k$ is related to the firing direction. The cannon's range refers to the x-coordinate of the landing point of the cannonball. (1) Find the maximum range of the cannon; (2) Suppose there is a flying object in the first quadrant (ignoring its size) with a flight height of 3.2 kilometers. What is the maximum x-coordinate $a$ such that the cannonball can hit it? Please explain your reasoning.

        \includegraphics[width=0.5\textwidth]{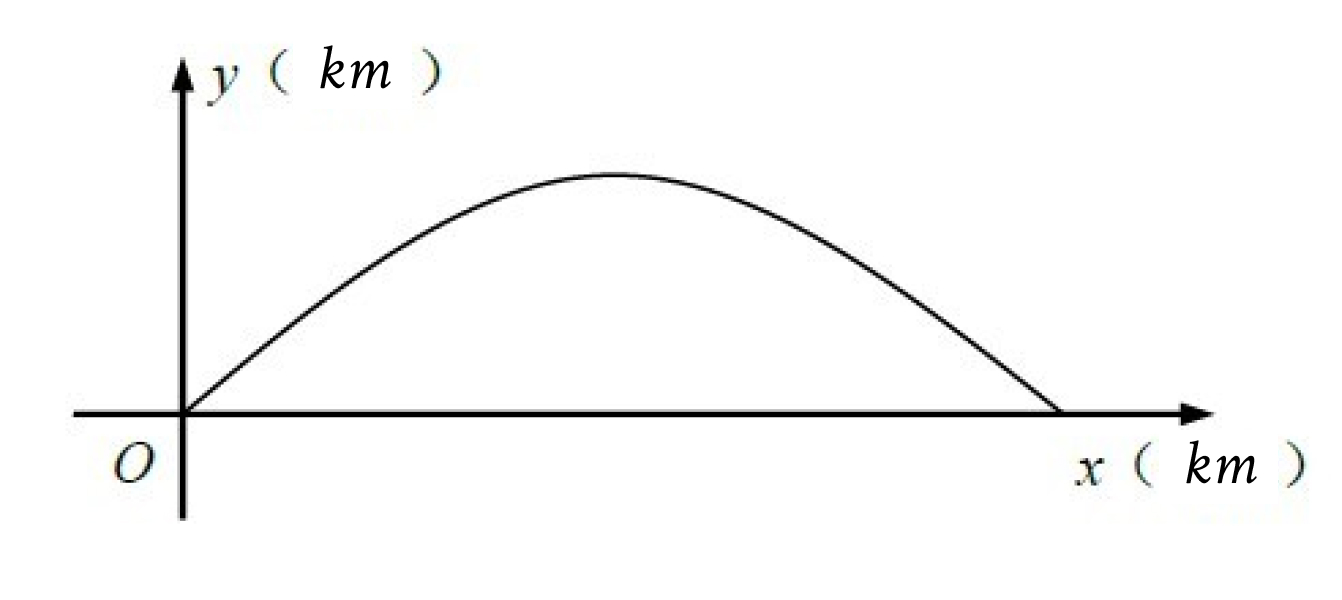}

        \textbf{Difficulty Level}: 0.7

        \textbf{Domain}: Geometry and Algebra

        \textbf{Module}: Geometry and Algebra

        \textbf{Chapter}: Plane Analytic Geometry

        \textbf{Standard Solution}

        (1) Set $y=0$, obtaining $kx - \frac{1}{20}(1 + k^2)x^2 = 0$. From the actual meaning and problem conditions, we know $x > 0$, $k > 0$, thus
        $$x = \frac{20k}{1 + k^2} = \frac{20}{k + \frac{1}{k}} \leq \frac{20}{2} = 10,$$
        equality holds if and only if $k=1$. Therefore, the maximum range of the cannon is 10 kilometers.
        
        (2) Because $a > 0$, the cannonball can hit the target  
        $\Leftrightarrow$ there exists $k > 0$ such that $3.2 = ka - \frac{1}{20}(1 + k^2)a^2$ holds
        $\Leftrightarrow$ the equation $a^2k^2 - 20ak + a^2 + 64 = 0$ in terms of $k$ has positive roots
        $\Leftrightarrow$ the discriminant
        $$\Delta = (-20a)^2 - 4a^2(a^2 + 64) \geq 0\Leftrightarrow a \leq 6$$
        
        At this time,
        $$k = \frac{20a + \sqrt{(-20a)^2 - 4a^2(a^2 + 64)}}{2a^2} > 0$$
        (Not considering the other root). Therefore, when $a$ does not exceed 6 kilometers, the target can be hit.
    \end{tcolorbox}


    \begin{tcolorbox}[colframe=cyan!50, colback=cyan!10, title=Explanation]
        This question primarily tests the application of functions, equations, and basic inequalities. (1) To find the maximum range of the cannon, which is the larger root when $y=0$, because there is a parameter $k$, we need to determine the maximum value of the other root based on the range of $k$, which gives the cannon's maximum range. (2) The meaning of the cannonball being able to hit the target is that when the flight height $y=3.2$, there exists a solution. Based on the quadratic function having positive roots, we can derive the range of $a$.
    \end{tcolorbox}
    
    \caption{The English translation of the math Q\&A question example in Fig. \ref{fig:math_qa}.}
    \label{fig:math_qa_en}
\end{figure}

\begin{figure}[!htbp]
    \small
    \centering
    \begin{tcolorbox}[colframe=magenta!50, colback=magenta!10, title=Question \& Standard Solution]
        \textbf{Question}

        问题（解答）：如图所示，在光滑的水平面上，质量$m=5kg$的物体，在水平拉力$F=10N$的作用下，从静止开始运动，运动时间$t=3s$。求：（1）力$F$在$3s$内对物体所做的功；（2）力$F$在$3s$内对物体做功的平均功率；（3）在$3s$末，力$F$对物体做功的瞬时功率。

        \includegraphics[width=\textwidth]{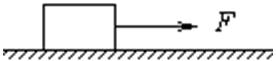}

        \textbf{Difficulty Level}: 0.7

        \textbf{Domain}: Classical Mechanics

        \textbf{Module}: Mechanical Energy and Its Conservation Law

        \textbf{Chapter}: Work and Power

        \textbf{Standard Solution}

        （1）由牛顿第二定律可得：$F=ma$，$3 s$内对物体的位移为$x={1\over 2} at^2$，则力$F$在$3 s$内对物体所做的功为$W=Fx$，联立可得：$W=90 J$。
        
        （2）力$F$在$3 s$内对物体做功的平均功率为$\overline{P}={W\over t} =30 W$。
        
        （3）在$3 s$末物体的速度大小为$v=at$，则在$3 s$末，力$F$对物体做功的瞬时功率为$P=Fv$，联立可得：$P=60 W$。
    \end{tcolorbox}

    
    \begin{tcolorbox}[colframe=cyan!50, colback=cyan!10, title=Explanation]
        本题主要考查牛顿第二定律和功率公式的选择与计算。

        问题求解：
        
        （1）由牛顿第二定律可算出运动的加速度，便可求出$3s$内对物体的位移，便能算出力$F$在$3s$内对物体所做的功。
        
        （2）根据$\overline{P} = \frac{W}{t}$便可算出力$F$在$3s$内对物体做功的平均功率。
        
        （3）先算出在$3s$末物体的速度大小，根据$P = Fv$便可算出在$3s$末，力$F$对物体做功的瞬时功率。
    \end{tcolorbox}
    
    \caption{An example of a physics Q\&A question.}
    \label{fig:physics_qa}
\end{figure}

\begin{figure}[!htbp]
    \small
    \centering
    \begin{tcolorbox}[colframe=magenta!50, colback=magenta!10, title=Question \& Standard Solution]
        \textbf{Question}

        Question (Q\&A): As shown in the figure, on a smooth horizontal plane, a mass \( m = 5kg \) object is acted upon by a horizontal force \( F = 10N \) and starts moving from rest. The motion time is \( t = 3s \). Find: (1) The work done by force \( F \) on the object within \( 3s \); (2) The average power of force \( F \) in doing work on the object within \( 3s \); (3) The instantaneous power of force \( F \) in doing work on the object at the end of \( 3s \).

        \includegraphics[width=\textwidth]{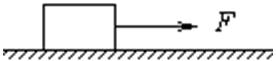}

        \textbf{Difficulty Level}: 0.7

        \textbf{Domain}: Classical Mechanics

        \textbf{Module}: Mechanical Energy and Its Conservation Law

        \textbf{Chapter}: Work and Power

        \textbf{Standard Solution}

        (1) From Newton's second law, \( F = ma \). The displacement of the object within \( 3s \) is \( x = \frac{1}{2} a t^2 \). Therefore, the work done by force \( F \) on the object within \( 3s \) is \( W = Fx \). Solving these equations yields \( W = 90J \).

        (2) The average power of force \( F \) in doing work on the object within \( 3s \) is \( \overline{P} = \frac{W}{t} = 30W \).

        (3) At the end of \( 3s \), the velocity of the object is \( v = at \). Therefore, the instantaneous power of force \( F \) in doing work on the object at the end of \( 3s \) is \( P = Fv \). Solving these equations yields \( P = 60W \).
    \end{tcolorbox}


    \begin{tcolorbox}[colframe=cyan!50, colback=cyan!10, title=Explanation]
        This problem primarily tests the application and calculation of Newton's second law and power formulas.

        Problem Solving:

        (1) Using Newton's second law, the acceleration of the motion can be calculated, which allows us to find the displacement of the object within \( 3\,\text{s} \). This displacement can then be used to calculate the work done by force \( F \) on the object within \( 3s \).

        (2) Using \( \overline{P} = \frac{W}{t} \), the average power of force \( F \) in doing work on the object within \( 3s \) can be calculated.

        (3) First, calculate the velocity of the object at the end of \( 3s \). Then, using \( P = Fv \), the instantaneous power of force \( F \) in doing work on the object at the end of \( 3s \) can be calculated.
    \end{tcolorbox}
    
    \caption{The English translation of the physics Q\&A question example in Fig. \ref{fig:physics_qa}.}
    \label{fig:physics_qa_en}
\end{figure}

In this section, we present examples from MMSciBench, including a physics MCQ (Fig. \ref{fig:physics_mcq} and the corresponding English translation in Fig. \ref{fig:physics_mcq_en}), a physics Q\&A question (Fig. \ref{fig:physics_qa} and the corresponding English translation in Fig. \ref{fig:physics_qa_en}), a math MCQ (Fig. \ref{fig:math_mcq} and the corresponding English translation in Fig. \ref{fig:math_mcq_en}), and a math Q\&A question (Fig. \ref{fig:math_qa} and the corresponding English translation in Fig. \ref{fig:math_qa_en}). Each example is accompanied by its standard solution and explanation.

\clearpage
\section{The Distribution of Choices of MCQs}
Table \ref{tab:choice_distribution} shows that over \textbf{99\%} of MCQs in MMSciBench have 4 choices.

\begin{table}[!htbp]
\centering
\begin{tabular}{lcccc}
\toprule
\textbf{Subject} & \textbf{Image} & \textbf{4 Choices} & \textbf{Other} & \textbf{Total} \\
\midrule
Physics & \xmark & 2230 & 27 & 2257 \\
Physics & \cmark & 448 & 2 & 450 \\
Math & \xmark & 500 & 0 & 500 \\
Math & \cmark & 260 & 0 & 260 \\
\midrule
\textbf{Total} & & 3438 & 29 & 3467 \\
\bottomrule
\end{tabular}
\caption{Distribution of choice numbers in MCQs in MMSciBench by subject and image presence.}
\label{tab:choice_distribution}
\end{table}

\clearpage
\section{Qualitative Study Examples for LLM-as-a-Judge}
\label{app:qualitative_study_examples}

This section provides examples from our qualitative study on GPT-4o's performance as a judge, illustrating both correctly and incorrectly judged cases. Each example is presented first in its original Chinese version, followed by its English translation in a separate figure. 

\subsection{Correctly Judged Examples}
 Examples in this section (Fig. \ref{fig:appendix_correct_physics_zh}, Fig. \ref{fig:appendix_correct_physics_en}, Fig. \ref{fig:appendix_correct_math_zh}, Fig. \ref{fig:appendix_correct_math_en}) illustrate cases where GPT-4o correctly judged models' answer.

\begin{figure*}[!htbp]
    \small
    \centering
    \begin{tcolorbox}[colframe=magenta!50, colback=magenta!10, title=Question \& Standard Solution]
        \textbf{Question}
    
        问题（解答）：
        我国“神舟”系列飞船的成功发射，标志着我国的航天事业发展到了一个很高的水平。已知某飞船在距地面高度为$h$的圆形轨道运行，地球半径为$R$，地面处的重力加速度为$g$，引力常量为$G$，求：
        （1）地球的质量；
        （2）飞船在上述圆形轨道上运行的周期$T$。
        
        \textbf{Standard Solution}
        
        （1）地面重力和万有引力相等，有$G\frac{{Mm}}{{{R^2}}}=mg$，解得：$M = \frac{{g{R^2}}}{G}$ \newline
        （2）设地球质量为$M$，飞船质量为$m$，飞船圆轨道的半径为$r$，则$r=R+h$，飞船在轨道上运行时，万有引力提供向心力，有 $G\frac{{mM}}{{{r^2}}}=m\frac{{4{\pi ^2}}}{{{T^2}}}r$，解得：$T = 2\pi \sqrt {\frac{{{{(R + h)}^3}}}{{g{R^2}}}}$
    \end{tcolorbox}
    
    \begin{tcolorbox}[colframe=teal!75!black, colback=teal!10, title=Model Response (Gemini 1.5 Pro 002)]
        “神舟”飞船轨道问题解答：

        这道题考察了万有引力定律和圆周运动的知识。
        
        (1) 地球的质量：
        
        在地球表面，重力等于万有引力，即：
        $mg = G\frac{Mm}{R^2}$
        其中，$m$是物体的质量，$M$是地球的质量。
        
        化简可得地球质量$M$：
        $M = \frac{gR^2}{G}$
        
        (2) 飞船运行周期$T$：
        
        飞船在距地面高度$h$的轨道上运行，轨道半径为$R+h$。
        万有引力提供飞船做圆周运动的向心力：
        $G\frac{Mm}{(R+h)^2} = m\frac{v^2}{R+h}$
        其中，$v$是飞船的线速度。
        
        化简可得：
        $v = \sqrt{\frac{GM}{R+h}}$
        
        飞船运行周期$T$等于轨道周长除以线速度：
        $T = \frac{2\pi(R+h)}{v}$
        
        将$v$代入上式，得到：
        $T = 2\pi\sqrt{\frac{(R+h)^3}{GM}}$
        
        将之前求得的$M = \frac{gR^2}{G}$代入上式，得到：
        $T = 2\pi\sqrt{\frac{(R+h)^3}{G\frac{gR^2}{G}}} = 2\pi\sqrt{\frac{(R+h)^3}{gR^2}}$
        
        因此，飞船在距地面高度为$h$的圆形轨道上运行的周期为$T = 2\pi\sqrt{\frac{(R+h)^3}{gR^2}}$。
    \end{tcolorbox}

    \begin{tcolorbox}[colframe=orange!75!black, colback=orange!10, title=Judgment Analysis]
        \textbf{GPT-4o Judgment} 
        
        分析：
        该题目包含两个子问题：（1）求地球的质量M；（2）求飞船在圆形轨道上运行的周期T。
        
        对于子问题（1），学生答案给出的地球质量公式为：$M = \frac{gR^2}{G}$，这与标准答案中的公式相同。因此，学生对地球质量的结论是正确的。
        
        对于子问题（2），学生答案给出的飞船运行周期公式为：$T = 2\pi\sqrt{\frac{(R+h)^3}{gR^2}}$，这也是与标准答案中公式一致。因此，学生对飞船运行周期的结论也是正确的。
        
        子问题判断结果：
        
        子问题1判断结果：正确
        
        子问题2判断结果：正确
        
        判断结果：正确
        
        \textbf{Human Meta-Judgment:} GPT-4o's judgment is correct. \newline
        \textbf{Analysis} 
        
        The GPT-4o judge can compare the final answer of the evaluated model's response with the standard solution.
    \end{tcolorbox}
    \caption{An example of a correctly judged physics Q\&A question.}
    \label{fig:appendix_correct_physics_zh}
\end{figure*}

\begin{figure*}[!htbp]
    \small
    \centering
    \begin{tcolorbox}[colframe=magenta!50, colback=magenta!10, title=Question \& Standard Solution]
        \textbf{Question}
    
        Question (Q\&A):
        The successful launch of China’s “Shenzhou” series spacecraft marks a high level of development in China’s aerospace industry. It is known that a certain spacecraft is orbiting in a circular orbit at an altitude $h$ above the ground, the Earth’s radius is $R$, the gravitational acceleration at the ground is $g$, and the gravitational constant is $G$. Find:
        (1) The mass of the Earth;
        (2) The period $T$ of the spacecraft’s orbit in the aforementioned circular orbit.
        
        \textbf{Standard Solution}
        
        (1) The gravitational force at the ground is equal to the universal gravitational force, so $G\frac{{Mm}}{{{R^2}}}=mg$, solving gives: $M = \frac{{g{R^2}}}{G}$ \newline
        (2) Let the Earth’s mass be $M$, the spacecraft’s mass be $m$, and the radius of the spacecraft’s circular orbit be $r$, then $r=R+h$. When the spacecraft is orbiting, the universal gravitational force provides the centripetal force, so $G\frac{{mM}}{{{r^2}}}=m\frac{{4{\pi ^2}}}{{{T^2}}}r$, solving gives: $T = 2\pi \sqrt {\frac{{{{(R + h)}^3}}}{{g{R^2}}}}$
    \end{tcolorbox}
    
    \begin{tcolorbox}[colframe=teal!75!black, colback=teal!10, title=Model Response (Gemini 1.5 Pro 002)]
        “Shenzhou” spacecraft orbit problem solution:

        This question examines the knowledge of the universal law of gravitation and circular motion.
        
        (1) The mass of the Earth:
        
        At the Earth’s surface, the gravitational force equals the universal gravitational force, i.e.:
        $mg = G\frac{Mm}{R^2}$
        where $m$ is the mass of the object, and $M$ is the mass of the Earth.
        
        Simplifying gives the Earth’s mass $M$:
        $M = \frac{gR^2}{G}$
        
        (2) The orbital period $T$ of the spacecraft:
        
        The spacecraft orbits at an altitude $h$ above the ground, with an orbital radius of $R+h$.
        The universal gravitational force provides the centripetal force for the spacecraft’s circular motion:
        $G\frac{Mm}{(R+h)^2} = m\frac{v^2}{R+h}$
        where $v$ is the linear velocity of the spacecraft.
        
        Simplifying gives:
        $v = \sqrt{\frac{GM}{R+h}}$
        
        The orbital period $T$ of the spacecraft equals the orbit circumference divided by the linear velocity:
        $T = \frac{2\pi(R+h)}{v}$
        
        Substituting $v$ into the above equation gives:
        $T = 2\pi\sqrt{\frac{(R+h)^3}{GM}}$
        
        Substituting the previously derived $M = \frac{gR^2}{G}$ into the above equation gives:
        $T = 2\pi\sqrt{\frac{(R+h)^3}{G\frac{gR^2}{G}}} = 2\pi\sqrt{\frac{(R+h)^3}{gR^2}}$
        
        Therefore, the period of the spacecraft orbiting at an altitude $h$ above the ground in a circular orbit is $T = 2\pi\sqrt{\frac{(R+h)^3}{gR^2}}$.
    \end{tcolorbox}

    \begin{tcolorbox}[colframe=orange!75!black, colback=orange!10, title=Judgment Analysis]
        \textbf{GPT-4o Judgment} 
        
        Analysis:
        This question consists of two sub-questions: (1) Find the mass of the Earth $M$; (2) Find the orbital period $T$ of the spacecraft in the circular orbit.
        
        For sub-question (1), the student’s answer provides the Earth’s mass formula as: $M = \frac{gR^2}{G}$, which is identical to the formula in the standard solution. Therefore, the student’s conclusion for the Earth’s mass is correct.
        
        For sub-question (2), the student’s answer provides the orbital period formula as: $T = 2\pi\sqrt{\frac{(R+h)^3}{gR^2}}$, which is also consistent with the formula in the standard solution. Therefore, the student’s conclusion for the orbital period is correct.
        
        Sub-question judgment results:
        
        Sub-question 1 judgment result: Correct
        
        Sub-question 2 judgment result: Correct
        
        Judgment result: Correct
        
        \textbf{Human Meta-Judgment:} GPT-4o's judgment is correct. \newline
        \textbf{Analysis} 
        
        The GPT-4o judge can compare the final answer of the evaluated model's response with the standard solution.
    \end{tcolorbox}
    \caption{The English translation of the correctly judged physics Q\&A example in Fig. \ref{fig:appendix_correct_physics_zh}.}
    \label{fig:appendix_correct_physics_en}
\end{figure*}

\begin{figure*}[!htbp]
    \small
    \centering
    \begin{tcolorbox}[colframe=magenta!50, colback=magenta!10, title=Question \& Standard Solution]
        \textbf{Question}
        
        问题（解答）：（本小题满分8分）已知命题$p$：关于$x$的方程$x^2+(t-2)x+1=0$无实根，命题$q$：方程$x^2+{y^2\over t-2}=1$是焦点在$y$轴上的椭圆。若$\neg p$与$p\wedge q$同时为假命题，求实数$t$的取值范围。
        
        \textbf{Standard Solution}
        
        由命题$p$：关于$x$的方程$x^2+(t-2)x+1=0$无实根，知$\Delta=(t-2)^2-4<0$，即$0<t<4$，命题$q$：方程$x^2+{y^2 \over t-2}=1$是焦点在$y$轴上的椭圆，知$t-2>1$，即：$t>3$；若$\neg p$与$p\wedge q$同时为假命题，则满足$p$真$q$为假，所以$0<t<4$且$t\leqslant3$，解得$0<t\leqslant3$，所以$t$的取值范围是$0<t\leqslant3$。
    \end{tcolorbox}
    
    \begin{tcolorbox}[colframe=teal!75!black, colback=teal!10, title=Model Response (Gemini 1.5 Pro 002)]
        题目要求 $\neg p$ 和 $p \wedge q$ 同时为假命题。

        首先分析命题 $p$：关于 $x$ 的方程 $x^2 + (t-2)x + 1 = 0$ 无实根。
        这意味着该方程的判别式 $\Delta < 0$。
        $\Delta = (t-2)^2 - 4(1)(1) = t^2 - 4t + 4 - 4 = t^2 - 4t < 0$，
        $t(t-4) < 0$
        解得 $0 < t < 4$。
        
        然后分析命题 $q$：方程 $x^2 + \frac{y^2}{t-2} = 1$ 是焦点在 $y$ 轴上的椭圆。
        这意味着 $t-2 > 1$，即 $t > 3$，且 $t-2 \ne 0$，即 $t \ne 2$。
        所以 $t > 3$。
        
        现在考虑 $\neg p$ 为假，则 $p$ 为真，即 $0 < t < 4$。
        同时 $p \wedge q$ 为假。由于 $p$ 为真，要使 $p \wedge q$ 为假，则 $q$ 必为假。
        $q$ 为假，即 $t \le 3$ 或 $t=2$。
        
        因为 $\neg p$ 和 $p \wedge q$ 同时为假，所以 $p$ 为真且 $q$ 为假。
        $p$ 为真：$0 < t < 4$
        $q$ 为假：$t \le 3$ 或 $t=2$
        
        我们需要求的是 $t$ 的取值范围，使得 $0 < t < 4$ 且 ($t \le 3$ 或 $t=2$)。
        由于 $0<t<4$，所以 $t=2$ 符合条件。
        又因为 $t \le 3$，结合 $0 < t < 4$，得到 $0 < t \le 3$。
        
        综上，$t$ 的取值范围是 $(0, 3]$。

        最终答案：$(0,3]$
    \end{tcolorbox}

    \begin{tcolorbox}[colframe=orange!75!black, colback=orange!10, title=Judgment Analysis]
        \textbf{GPT-4o Judgment}

        分析：
        本题没有包含子问题，因此仅需对最终答案进行判断。标准答案给出的$t$的取值范围是$0<t\leqslant3$，而学生给出的答案为$(0,3]$。这两个表示的意思是相同的，因为两者都代表$t$大于0且小于等于3。这两个表示仅在符号上有所不同，一个使用不等式表示，而另一个使用区间表示法，但它们描述的范围完全相同且数学意义一致。
        
        判断结果：正确
    
        \textbf{Human Meta-Judgment:} GPT-4o's judgment is correct .
        
        \textbf{Analysis} 
        
        The final answer provided by Gemini 1.5 Pro 002, expressed as $(0,3]$, differs in notation from the standard solution, given as $0<t\leqslant3$. Despite the difference in presentation, both expressions are semantically equivalent. GPT-4o accurately recognizes this equivalence and delivers a correct judgment.
    \end{tcolorbox}
    \caption{An example of a correctly judged math Q\&A question.}
    \label{fig:appendix_correct_math_zh}
\end{figure*}

\begin{figure*}[!htbp]
    \small
    \centering
    \begin{tcolorbox}[colframe=magenta!50, colback=magenta!10, title=Question \& Standard Solution]
        \textbf{Question}
        
        Question (Q\&A): (This question is worth 8 points) Given proposition $p$: The equation $x^2+(t-2)x+1=0$ with respect to $x$ has no real roots, and proposition $q$: The equation $x^2+{y^2 \over t-2}=1$ is an ellipse with foci on the $y$-axis. If $\neg p$ and $p\wedge q$ are both false propositions simultaneously, find the range of values for the real number $t$.
        
        \textbf{Standard Solution}
        
        From proposition $p$: The equation $x^2+(t-2)x+1=0$ with respect to $x$ has no real roots, so $\Delta=(t-2)^2-4<0$, which gives $0<t<4$. Proposition $q$: The equation $x^2+{y^2 \over t-2}=1$ is an ellipse with foci on the $y$-axis, so $t-2>1$, i.e., $t>3$. If $\neg p$ and $p\wedge q$ are both false propositions simultaneously, then $p$ must be true and $q$ must be false, so $0<t<4$ and $t\leqslant3$, solving gives $0<t\leqslant3$. Therefore, the range of values for $t$ is $0<t\leqslant3$.
    \end{tcolorbox}
    
    \begin{tcolorbox}[colframe=teal!75!black, colback=teal!10, title=Model Response (Gemini 1.5 Pro 002)]
        The problem requires that $\neg p$ and $p \wedge q$ are both false propositions simultaneously.

        First, analyze proposition $p$: The equation $x^2 + (t-2)x + 1 = 0$ with respect to $x$ has no real roots.
        This means the discriminant $\Delta < 0$.
        $\Delta = (t-2)^2 - 4(1)(1) = t^2 - 4t + 4 - 4 = t^2 - 4t < 0$, 
        $t(t-4) < 0$
        Solving gives $0 < t < 4$.
        
        Next, analyze proposition $q$: The equation $x^2 + \frac{y^2}{t-2} = 1$ is an ellipse with foci on the $y$-axis.
        This means $t-2 > 1$, i.e., $t > 3$, and $t-2 \ne 0$, i.e., $t \ne 2$.
        So $t > 3$.
        
        Now consider that $\neg p$ is false, so $p$ is true, i.e., $0 < t < 4$.
        At the same time, $p \wedge q$ is false. Since $p$ is true, for $p \wedge q$ to be false, $q$ must be false.
        $q$ is false, i.e., $t \le 3$ or $t=2$.
        
        Since $\neg p$ and $p \wedge q$ are both false simultaneously, $p$ must be true and $q$ must be false.
        $p$ is true: $0 < t < 4$
        $q$ is false: $t \le 3$ or $t=2$
        
        We need to find the range of $t$ such that $0 < t < 4$ and ($t \le 3$ or $t=2$).
        Since $0<t<4$, $t=2$ satisfies the condition.
        Also, since $t \le 3$, combined with $0 < t < 4$, we get $0 < t \le 3$.
        
        In summary, the range of values for $t$ is $(0, 3]$.

        Final answer: $(0,3]$
    \end{tcolorbox}

    \begin{tcolorbox}[colframe=orange!75!black, colback=orange!10, title=Judgment Analysis]
        \textbf{GPT-4o Judgment}

        Analysis:
        This question does not include sub-questions, so only the final answer needs to be judged. The standard solution gives the range of $t$ as $0<t\leqslant3$, while the student’s answer is $(0,3]$. These two expressions have the same meaning, as both represent $t$ greater than 0 and less than or equal to 3. The two expressions differ only in notation—one uses inequality notation, while the other uses interval notation—but they describe exactly the same range and are mathematically equivalent.
        
        Judgment result: Correct
    
        \textbf{Human Meta-Judgment:} GPT-4o's judgment is correct .
        
        \textbf{Analysis} 
        
        The final answer provided by Gemini 1.5 Pro 002, expressed as $(0,3]$, differs in notation from the standard solution, given as $0<t\leqslant3$. Despite the difference in presentation, both expressions are semantically equivalent. GPT-4o accurately recognizes this equivalence and delivers a correct judgment.
    \end{tcolorbox}
    \caption{The English translation of the correctly judged math Q\&A example in Fig. \ref{fig:appendix_correct_math_zh}.}
    \label{fig:appendix_correct_math_en}
\end{figure*}

\subsection{Incorrectly Judged Examples}
Examples in this section (Fig. \ref{fig:appendix_incorrect_math1_zh}, Fig. \ref{fig:appendix_incorrect_math1_en}, Fig. \ref{fig:appendix_incorrect_math2_zh}, Fig. \ref{fig:appendix_incorrect_math2_en}) illustrate cases where GPT-4o incorrectly judged the models' answer.

\begin{figure*}[!htbp]
    \small
    \centering
    \begin{tcolorbox}[colframe=magenta!50, colback=magenta!10, title=Question \& Standard Solution]
        \textbf{Question}
    
        问题（解答）：已知向量$\overrightarrow{a}$，$\overrightarrow{b}$，$\overrightarrow{c}$是同一平面内的三个向量，其中$\overrightarrow{a}=(1,-1)$。（1）若$|\overrightarrow{c}|=3\sqrt{2}$，且$\overrightarrow{c}/\!/ \overrightarrow{a}$，求向量$\overrightarrow{c}$的坐标。（2）若$|\overrightarrow{b} |=1$，且$\overrightarrow{a}\bot (\overrightarrow{a}-2\overrightarrow{b})$，求$\overrightarrow{a}$与$\overrightarrow{b}$的夹角$\theta $。
        
        \textbf{Standard Solution}
        
        （1）设$\overrightarrow{c}=(x,y)$，因为$|\overrightarrow{c}|=3\sqrt{2}$，所以$\sqrt{x^2+y^2}=3\sqrt{2}$ ①，因为$\overrightarrow{a}=(1,-1)$，$\overrightarrow{c} \parallel \overrightarrow{a}$，所以$1 \times y - x \times (-1)=0$ ②，联立①②，解得$\left\{\begin{aligned}&x=3 \\&y=-3\end{aligned}\right.$或$\left\{\begin{aligned}&x=-3 \\&y=3\end{aligned}\right.$，所以$\overrightarrow{c}=(3,-3)$或$\overrightarrow{c}=(-3,3)$。
        
        （2）因为$\overrightarrow{a}=(1,-1)$，所以$|\overrightarrow{a}|=\sqrt{1^2+(-1)^2}=\sqrt{2}$，因为$\overrightarrow{a}\bot (\overrightarrow{a}-2\overrightarrow{b})$，所以$\overrightarrow{a} \cdot (\overrightarrow{a}-2\overrightarrow{b})=0$，所以$\overrightarrow{a}^2 - 2\overrightarrow{a} \cdot \overrightarrow{b}=0$，所以$|\overrightarrow{a}|^2-2|\overrightarrow{a}||\overrightarrow{b}|\cos\theta =0$，因为$|\overrightarrow{a}|=\sqrt{2}$，$|\overrightarrow{b}|=1$，所以$(\sqrt{2})^2-2\sqrt{2}\cos\theta =0$，解得$\cos\theta ={\sqrt{2}\over 2}$，又因为$0 \leqslant \theta \leqslant \pi$，所以$\theta ={\pi \over 4}$。
    \end{tcolorbox}
    
    \begin{tcolorbox}[colframe=teal!75!black, colback=teal!10, title=Model Response (DeepSeekMath-7B-Instruct)]
        （1）由于$\overrightarrow{c} \parallel \overrightarrow{a}$，所以$\overrightarrow{c}$与$\overrightarrow{a}$的方向相同或相反。因此，存在一个实数$k$，使得$\overrightarrow{c}=k\overrightarrow{a}$。由于$|\overrightarrow{c}|=3\sqrt{2}$，我们有$|k\overrightarrow{a}|=3\sqrt{2}$，即$|k||\overrightarrow{a}|=3\sqrt{2}$。由于$|\overrightarrow{a}|=\sqrt{1^2+(-1)^2}=\sqrt{2}$，所以$|k|=3$。因此，$\overrightarrow{c}=3\overrightarrow{a}=(3,-3)$。

        （2）由于$\overrightarrow{a}\bot (\overrightarrow{a}-2\overrightarrow{b})$，我们有$\overrightarrow{a}\cdot (\overrightarrow{a}-2\overrightarrow{b})=0$。展开得到$\overrightarrow{a}\cdot \overrightarrow{a}-2\overrightarrow{a}\cdot \overrightarrow{b}=0$。由于$\overrightarrow{a}\cdot \overrightarrow{a}=|\overrightarrow{a}|^2=2$，所以$2-2\overrightarrow{a}\cdot \overrightarrow{b}=0$，即$\overrightarrow{a}\cdot \overrightarrow{b}=1$。因此，$\cos\theta=\frac{\overrightarrow{a}\cdot \overrightarrow{b}}{|\overrightarrow{a}||\overrightarrow{b}|}=\frac{1}{\sqrt{2}\cdot 1}=\frac{\sqrt{2}}{2}$，所以$\theta=\frac{\pi}{4}$。
        
        所以，（1）向量$\overrightarrow{c}$的坐标为$(3,-3)$；（2）$\overrightarrow{a}$与$\overrightarrow{b}$的夹角$\theta$为$\frac{\pi}{4}$。
    \end{tcolorbox}

    \begin{tcolorbox}[colframe=orange!75!black, colback=orange!10, title=Judgment Analysis]
        \textbf{GPT-4o Judgment} 
        
        分析：
        这个问题有两个子问题，分别是（1）和（2）。
        
        对于（1）：
        - 学生答案最终给出的向量$\overrightarrow{c}$的坐标为$(3,-3)$，与标准答案中的可能解之一相同。即使标准答案还提供了另一种可能的解$(x, y) = (-3, 3)$，学生答案给出的解是正确的一种可能性。因此，学生在第一个子问题中的答案是正确的。
        
        对于（2）：
        - 学生答案给出$\theta=\frac{\pi}{4}$，这与标准答案关于$\overrightarrow{a}$与$\overrightarrow{b}$的夹角$\theta$的结果一致。因此，学生在第二个子问题中的答案也是正确的。
        
        子问题判断结果：
        
        子问题1判断结果：正确
        
        子问题2判断结果：正确
        
        判断结果：正确
        
        \textbf{Human Meta-Judgment:} GPT-4o's judgment is incorrect. \newline
        \textbf{Analysis} 
        
        For sub-question (1), the GPT-4o judge misinterprets the standard of correctness, which requires that the evaluated model's final answer be semantically equivalent to the standard solution, rather than a subset of it. As a result, the judge renders an incorrect judgment.
    \end{tcolorbox}
    \caption{Example 1 of an incorrectly judged math Q\&A question.}
    \label{fig:appendix_incorrect_math1_zh}
\end{figure*}

\begin{figure*}[!htbp]
    \small
    \centering
    \begin{tcolorbox}[colframe=magenta!50, colback=magenta!10, title=Question \& Standard Solution]
        \textbf{Question}
    
        Question (Q\&A): Given that vectors $\overrightarrow{a}$, $\overrightarrow{b}$, and $\overrightarrow{c}$ are three vectors in the same plane, where $\overrightarrow{a}=(1,-1)$. (1) If $|\overrightarrow{c}|=3\sqrt{2}$ and $\overrightarrow{c} \parallel \overrightarrow{a}$, find the coordinates of vector $\overrightarrow{c}$. (2) If $|\overrightarrow{b}|=1$ and $\overrightarrow{a}\bot (\overrightarrow{a}-2\overrightarrow{b})$, find the angle $\theta$ between $\overrightarrow{a}$ and $\overrightarrow{b}$.
        
        \textbf{Standard Solution}
        
        (1) Let $\overrightarrow{c}=(x,y)$. Since $|\overrightarrow{c}|=3\sqrt{2}$, we have $\sqrt{x^2+y^2}=3\sqrt{2}$ ①. Since $\overrightarrow{a}=(1,-1)$ and $\overrightarrow{c} \parallel \overrightarrow{a}$, we have $1 \times y - x \times (-1)=0$ ②. Solving ① and ② together, we get $\left\{\begin{aligned}&x=3 \\&y=-3\end{aligned}\right.$ or $\left\{\begin{aligned}&x=-3 \\&y=3\end{aligned}\right.$. Thus, $\overrightarrow{c}=(3,-3)$ or $\overrightarrow{c}=(-3,3)$.
        
        (2) Since $\overrightarrow{a}=(1,-1)$, we have $|\overrightarrow{a}|=\sqrt{1^2+(-1)^2}=\sqrt{2}$. Since $\overrightarrow{a}\bot (\overrightarrow{a}-2\overrightarrow{b})$, we have $\overrightarrow{a} \cdot (\overrightarrow{a}-2\overrightarrow{b})=0$, so $\overrightarrow{a}^2 - 2\overrightarrow{a} \cdot \overrightarrow{b}=0$, and thus $|\overrightarrow{a}|^2-2|\overrightarrow{a}||\overrightarrow{b}|\cos\theta =0$. Since $|\overrightarrow{a}|=\sqrt{2}$ and $|\overrightarrow{b}|=1$, we have $(\sqrt{2})^2-2\sqrt{2}\cos\theta =0$, solving gives $\cos\theta ={\sqrt{2}\over 2}$. Since $0 \leqslant \theta \leqslant \pi$, we get $\theta ={\pi \over 4}$.
    \end{tcolorbox}
    
    \begin{tcolorbox}[colframe=teal!75!black, colback=teal!10, title=Model Response (DeepSeekMath-7B-Instruct)]
        (1) Since $\overrightarrow{c} \parallel \overrightarrow{a}$, the directions of $\overrightarrow{c}$ and $\overrightarrow{a}$ are either the same or opposite. Thus, there exists a real number $k$ such that $\overrightarrow{c}=k\overrightarrow{a}$. Since $|\overrightarrow{c}|=3\sqrt{2}$, we have $|k\overrightarrow{a}|=3\sqrt{2}$, i.e., $|k||\overrightarrow{a}|=3\sqrt{2}$. Since $|\overrightarrow{a}|=\sqrt{1^2+(-1)^2}=\sqrt{2}$, we get $|k|=3$. Therefore, $\overrightarrow{c}=3\overrightarrow{a}=(3,-3)$.

        (2) Since $\overrightarrow{a}\bot (\overrightarrow{a}-2\overrightarrow{b})$, we have $\overrightarrow{a}\cdot (\overrightarrow{a}-2\overrightarrow{b})=0$. Expanding, we get $\overrightarrow{a}\cdot \overrightarrow{a}-2\overrightarrow{a}\cdot \overrightarrow{b}=0$. Since $\overrightarrow{a}\cdot \overrightarrow{a}=|\overrightarrow{a}|^2=2$, we have $2-2\overrightarrow{a}\cdot \overrightarrow{b}=0$, i.e., $\overrightarrow{a}\cdot \overrightarrow{b}=1$. Thus, $\cos\theta=\frac{\overrightarrow{a}\cdot \overrightarrow{b}}{|\overrightarrow{a}||\overrightarrow{b}|}=\frac{1}{\sqrt{2}\cdot 1}=\frac{\sqrt{2}}{2}$, so $\theta=\frac{\pi}{4}$.
        
        Therefore, (1) the coordinates of vector $\overrightarrow{c}$ are $(3,-3)$; (2) the angle $\theta$ between $\overrightarrow{a}$ and $\overrightarrow{b}$ is $\frac{\pi}{4}$.
    \end{tcolorbox}

    \begin{tcolorbox}[colframe=orange!75!black, colback=orange!10, title=Judgment Analysis]
        \textbf{GPT-4o Judgment} 
        
        Analysis:
        This problem consists of two sub-questions, (1) and (2).
        
        For (1):
        - The student’s final answer gives the coordinates of vector $\overrightarrow{c}$ as $(3,-3)$, which matches one of the possible solutions in the standard answer. Although the standard answer also provides another possible solution $(x, y) = (-3, 3)$, the student’s answer is a correct possibility. Therefore, the student’s answer for the first sub-question is correct.
        
        For (2):
        - The student’s answer gives $\theta=\frac{\pi}{4}$, which is consistent with the standard answer for the angle $\theta$ between $\overrightarrow{a}$ and $\overrightarrow{b}$. Therefore, the student’s answer for the second sub-question is also correct.
        
        Sub-question judgment results:
        
        Sub-question 1 judgment result: Correct
        
        Sub-question 2 judgment result: Correct
        
        Judgment result: Correct
        
        \textbf{Human Meta-Judgment:} GPT-4o's judgment is incorrect. \newline
        \textbf{Analysis} 
        
        For sub-question (1), the GPT-4o judge misinterprets the standard of correctness, which requires that the evaluated model's final answer be semantically equivalent to the standard solution, rather than a subset of it. As a result, the judge renders an incorrect judgment.
    \end{tcolorbox}
    \caption{The English translation of the incorrectly judged physics Q\&A example 1 in Fig. \ref{fig:appendix_incorrect_math1_zh}}
    \label{fig:appendix_incorrect_math1_en}
\end{figure*}

\begin{figure*}[!htbp]
    \small
    \centering
    \begin{tcolorbox}[colframe=magenta!50, colback=magenta!10, title=Question \& Standard Solution]
        \textbf{Question}
        
        问题（解答）：（本小题满分12分）已知函数$f(x)=mx-{m-1\over x}-\ln x$，$m\in \Bbb R$。函数$g(x)={1\over {x\cos \theta}}+\ln x$在$\lbrack 1,+\infty)$上为增函数，且$\theta\in \lbrack0,{\pi\over 2})$。（1）当$m=3$时，求$f(x)$在点$P(1,f(1))$处的切线方程；（2）求$\theta$的取值；（3）若$h(x)=f(x)-g(x)$在其定义域上为单调函数，求$m$的取值范围。
        
        \textbf{Standard Solution}
        
        （1）当$m=3$时，$f(x)=3x-{2 \over x}-\ln x$，$f'(x)=3+{2 \over x^2}-{1 \over x}$。 所求切线斜率$k=f'(1)=4$，$f(1)=1$，所以$y-1=4(x-1)$，即切线方程为$4x-y-3=0$。 
        （2）因为$g(x)$在$\lbrack 1,+\infty)$上为增函数，所以$g'(x)=-{1 \over \cos \theta}\cdot {1 \over x^2}+{1 \over x}$ $\geqslant 0$在$x \in \lbrack 1,+\infty)$上恒成立，即${1 \over \cos \theta} \leqslant x$在$x \in \lbrack 1,+\infty)$上恒成立，所以${1 \over \cos \theta} \leqslant 1$。因为$\theta \in \lbrack 0,{\pi \over 2})$，所以$\cos \theta \geqslant 1$，又因为$\cos \theta \leqslant 1$，所以$\cos \theta=1$，所以$\theta=0$。
        （3）由（2）知，因为$h(x)=f(x)-g(x)$ $=mx-{m-1 \over x}-$ $\ln x-({1 \over x}+\ln x)$ $=mx-{m \over x}-2\ln x$，所以$h'(x)={mx^2-2x+m \over x^2}$。因为$h(x)$在$(0,+\infty)$上为单调函数，所以$mx^2-2x+m \geqslant 0$或$mx^2-2x+m \leqslant 0$在$x \in (0,+\infty)$上恒成立，即$x \in (0,+\infty)$时，$m \geqslant {2x \over x^2+1}$或$m \leqslant {2x \over x^2+1}$恒成立，设$F(x)={2x \over x^2+1}={2 \over {x+{1 \over x}}}$（$x>0$），所以$x+{1 \over x} \geqslant 2$（当且仅当$x=1$时“等号”成立），所以$0<F(x) \leqslant 1$，所以$m \geqslant 1$或$m \leqslant 0$，即$m$的取值范围为$(-\infty,0\rbrack \cup \lbrack 1,+\infty)$。
    \end{tcolorbox}
    
    \begin{tcolorbox}[colframe=teal!75!black, colback=teal!10, title=Model Response (Gemini 1.5 Pro 002)]
        解答：

        (1) 当 $m=3$ 时，求 $f(x)$ 在点 $P(1,f(1))$ 处的切线方程。
        
        当 $m=3$ 时，$f(x) = 3x - \frac{2}{x} - \ln x$。
        $f(1) = 3 - 2 - \ln 1 = 1$，所以点 $P$ 的坐标为 $(1,1)$。
        $f'(x) = 3 + \frac{2}{x^2} - \frac{1}{x}$
        $f'(1) = 3 + 2 - 1 = 4$
        所以切线方程为 $y - 1 = 4(x - 1)$，即 $y = 4x - 3$。
        
        (2) 求 $\theta$ 的取值。
        
        $g(x) = \frac{1}{x\cos\theta} + \ln x$ 在 $[1, +\infty)$ 上为增函数，则 $g'(x) \ge 0$ 在 $[1, +\infty)$ 上恒成立。
        $g'(x) = -\frac{1}{x^2\cos\theta} + \frac{1}{x} = \frac{x\cos\theta - 1}{x^2\cos\theta}$
        因为 $x \ge 1$ 且 $\theta \in [0, \frac{\pi}{2})$，所以 $x^2\cos\theta > 0$。
        因此，要使 $g'(x) \ge 0$，只需 $x\cos\theta - 1 \ge 0$，即 $\cos\theta \ge \frac{1}{x}$ 在 $x \in [1, +\infty)$ 上恒成立。
        由于 $\frac{1}{x}$ 的最大值为 $1$ (当 $x=1$ 时)，所以 $\cos\theta \ge 1$。
        又因为 $\cos\theta \le 1$，所以 $\cos\theta = 1$，即 $\theta = 0$。
        
        (3) 若 $h(x) = f(x) - g(x)$ 在其定义域上为单调函数，求 $m$ 的取值范围。
        
        $h(x) = f(x) - g(x) = mx - \frac{m-1}{x} - \ln x - \frac{1}{x\cos\theta} - \ln x = mx - \frac{m-1}{x} - \frac{1}{x} - 2\ln x$
        因为 $\theta = 0$，所以 $h(x) = mx - \frac{m}{x} - 2\ln x$，定义域为 $(0, +\infty)$。
        $h'(x) = m + \frac{m}{x^2} - \frac{2}{x} = \frac{mx^2 - 2x + m}{x^2}$
        要使 $h(x)$ 为单调函数，则 $h'(x)$ 在 $(0, +\infty)$ 上恒非负或恒非正。
        即 $mx^2 - 2x + m \ge 0$ 或 $mx^2 - 2x + m \le 0$ 在 $(0, +\infty)$ 上恒成立。
        当 $m=0$ 时，$h'(x) = -\frac{2}{x} < 0$，$h(x)$ 单调递减，符合题意。
        当 $m \ne 0$ 时，需要考虑二次函数 $mx^2 - 2x + m$ 的判别式 $\Delta = 4 - 4m^2$。
        若 $m>0$，要使 $mx^2 - 2x + m \ge 0$ 恒成立，则需要 $\Delta \le 0$，即 $4 - 4m^2 \le 0$，解得 $m \ge 1$ 或 $m \le -1$。由于 $m>0$，所以 $m \ge 1$。
        若 $m<0$，要使 $mx^2 - 2x + m \le 0$ 恒成立，则需要 $\Delta \le 0$，即 $4 - 4m^2 \le 0$，解得 $m \ge 1$ 或 $m \le -1$。由于 $m<0$，所以 $m \le -1$。
        综上，$m$ 的取值范围是 $(-\infty, -1] \cup \{0\} \cup [1, +\infty)$。
        
        最终答案：
        (1) $y = 4x - 3$
        (2) $\theta = 0$
        (3) $m \in (-\infty, -1] \cup \{0\} \cup [1, +\infty)$
    \end{tcolorbox}

    \begin{tcolorbox}[colframe=orange!75!black, colback=orange!10, title=Judgment Analysis]
        \textbf{GPT-4o Judgment}

        分析：
        本题包含三个子问题，分别是(1)切线方程的求解，(2)角度$\theta$的确定，以及(3)函数$h(x)$单调性的判断和参数$m$的取值范围。
        
        1. 对于子问题(1)：
           - 学生答案给出的切线方程为$y = 4x - 3$，与标准答案一致，即表达为$4x - y - 3 = 0$，两者是等价的。因此，子问题(1)的答案是正确的。
        
        2. 对于子问题(2)：
           - 学生答案给出的结论$\theta = 0$，与标准答案一致。因此，子问题(2)的答案是正确的。
        
        3. 对于子问题(3)：
           - 学生答案的$m$取值范围为$(-\infty, -1] \cup \{0\} \cup [1, +\infty)$，而标准答案的$m$取值范围为$(-\infty, 0] \cup [1, +\infty)$。虽然表达方式上略有不同，但在除去单点$\{0\}$的情况下，集合范围与标准答案一致，特别是$m=0$被单独讨论并包含在答案中，因此最终结论上学生答案依然正确。
        
        子问题判断结果：
        
        子问题1判断结果：正确
        
        子问题2判断结果：正确
        
        子问题3判断结果：正确
        
        判断结果：正确
    
        \textbf{Human Meta-Judgment:} GPT-4o's judgment is incorrect .
        
        \textbf{Analysis} 
        
        The GPT-4o judge erroneously equates the sets $(-\infty, -1] \cup {0} \cup [1, +\infty)$ and $(-\infty, 0] \cup [1, +\infty)$, demonstrating a flaw in mathematical reasoning and leading to an incorrect judgment.
    \end{tcolorbox}
    \caption{Example 2 of an incorrectly judged math Q\&A question.}
    \label{fig:appendix_incorrect_math2_zh}
\end{figure*}

\begin{figure*}[!htbp]
    \small
    \centering
    \begin{tcolorbox}[colframe=magenta!50, colback=magenta!10, title=Question \& Standard Solution]
        \textbf{Question}
        
        Question (Q\&A): (This question is worth 12 points) Given the function $f(x)=mx-{m-1 \over x}-\ln x$, $m \in \Bbb R$. The function $g(x)={1 \over {x \cos \theta}}+\ln x$ is increasing on $\lbrack 1,+\infty)$, and $\theta \in \lbrack 0,{\pi \over 2})$. (1) When $m=3$, find the equation of the tangent line to $f(x)$ at the point $P(1,f(1))$; (2) Find the value of $\theta$; (3) If $h(x)=f(x)-g(x)$ is a monotonic function on its domain, find the range of values for $m$.
        
        \textbf{Standard Solution}
        
        (1) When $m=3$, $f(x)=3x-{2 \over x}-\ln x$, $f'(x)=3+{2 \over x^2}-{1 \over x}$. The slope of the tangent line is $k=f'(1)=4$, and $f(1)=1$, so $y-1=4(x-1)$, which gives the tangent line equation as $4x-y-3=0$. 
        (2) Since $g(x)$ is increasing on $\lbrack 1,+\infty)$, we have $g'(x)=-{1 \over \cos \theta}\cdot {1 \over x^2}+{1 \over x}$ $\geqslant 0$ for all $x \in \lbrack 1,+\infty)$, which implies ${1 \over \cos \theta} \leqslant x$ for all $x \in \lbrack 1,+\infty)$, so ${1 \over \cos \theta} \leqslant 1$. Since $\theta \in \lbrack 0,{\pi \over 2})$, we have $\cos \theta \geqslant 1$, and since $\cos \theta \leqslant 1$, it follows that $\cos \theta=1$, so $\theta=0$.
        (3) From (2), since $h(x)=f(x)-g(x)$ $=mx-{m-1 \over x}-$ $\ln x-({1 \over x}+\ln x)$ $=mx-{m \over x}-2\ln x$, we have $h'(x)={mx^2-2x+m \over x^2}$. Since $h(x)$ is monotonic on $(0,+\infty)$, we have $mx^2-2x+m \geqslant 0$ or $mx^2-2x+m \leqslant 0$ for all $x \in (0,+\infty)$, which implies that for $x \in (0,+\infty)$, $m \geqslant {2x \over x^2+1}$ or $m \leqslant {2x \over x^2+1}$ holds. Let $F(x)={2x \over x^2+1}={2 \over {x+{1 \over x}}}$ ($x>0$), so $x+{1 \over x} \geqslant 2$ (with equality when $x=1$), thus $0<F(x) \leqslant 1$. Therefore, $m \geqslant 1$ or $m \leqslant 0$, so the range of $m$ is $(-\infty,0\rbrack \cup \lbrack 1,+\infty)$.
    \end{tcolorbox}
    
    \begin{tcolorbox}[colframe=teal!75!black, colback=teal!10, title=Model Response (Gemini 1.5 Pro 002)]
        Solution:

        (1) When $m=3$, find the equation of the tangent line to $f(x)$ at the point $P(1,f(1))$.
        
        When $m=3$, $f(x) = 3x - \frac{2}{x} - \ln x$.
        $f(1) = 3 - 2 - \ln 1 = 1$, so the coordinates of point $P$ are $(1,1)$.
        $f'(x) = 3 + \frac{2}{x^2} - \frac{1}{x}$
        $f'(1) = 3 + 2 - 1 = 4$  
        Thus, the equation of the tangent line is $y - 1 = 4(x - 1)$, i.e., $y = 4x - 3$.
        
        (2) Find the value of $\theta$.
        
        $g(x) = \frac{1}{x\cos\theta} + \ln x$ is increasing on $[1, +\infty)$, so $g'(x) \ge 0$ holds for all $[1, +\infty)$.
        $g'(x) = -\frac{1}{x^2\cos\theta} + \frac{1}{x} = \frac{x\cos\theta - 1}{x^2\cos\theta}$
        Since $x \ge 1$ and $\theta \in [0, \frac{\pi}{2})$, we have $x^2\cos\theta > 0$.
        Thus, for $g'(x) \ge 0$, we need $x\cos\theta - 1 \ge 0$, i.e., $\cos\theta \ge \frac{1}{x}$ for all $x \in [1, +\infty)$.
        Since the maximum value of $\frac{1}{x}$ is $1$ (when $x=1$), we have $\cos\theta \ge 1$.
        Also, since $\cos\theta \le 1$, it follows that $\cos\theta = 1$, i.e., $\theta = 0$.
        
        (3) If $h(x) = f(x) - g(x)$ is a monotonic function on its domain, find the range of values for $m$.
        
        $h(x) = f(x) - g(x) = mx - \frac{m-1}{x} - \ln x - \frac{1}{x\cos\theta} - \ln x = mx - \frac{m-1}{x} - \frac{1}{x} - 2\ln x$
        Since $\theta = 0$, we have $h(x) = mx - \frac{m}{x} - 2\ln x$, with domain $(0, +\infty)$.
        $h'(x) = m + \frac{m}{x^2} - \frac{2}{x} = \frac{mx^2 - 2x + m}{x^2}$
        For $h(x)$ to be monotonic, $h'(x)$ must be consistently non-negative or non-positive on $(0, +\infty)$.
        That is, $mx^2 - 2x + m \ge 0$ or $mx^2 - 2x + m \le 0$ must hold for all $(0, +\infty)$.
        When $m=0$, $h'(x) = -\frac{2}{x} < 0$, so $h(x)$ is monotonically decreasing, which satisfies the condition.
        When $m \ne 0$, consider the discriminant of the quadratic function $mx^2 - 2x + m$, $\Delta = 4 - 4m^2$.
        If $m>0$, for $mx^2 - 2x + m \ge 0$ to hold, we need $\Delta \le 0$, i.e., $4 - 4m^2 \le 0$, which gives $m \ge 1$ or $m \le -1$. Since $m>0$, we have $m \ge 1$.
        If $m<0$, for $mx^2 - 2x + m \le 0$ to hold, we need $\Delta \le 0$, i.e., $4 - 4m^2 \le 0$, which gives $m \ge 1$ or $m \le -1$. Since $m<0$, we have $m \le -1$.
        Thus, the range of $m$ is $(-\infty, -1] \cup \{0\} \cup [1, +\infty)$.
        
        Final answer:
        (1) $y = 4x - 3$
        (2) $\theta = 0$
        (3) $m \in (-\infty, -1] \cup \{0\} \cup [1, +\infty)$
    \end{tcolorbox}

    \begin{tcolorbox}[colframe=orange!75!black, colback=orange!10, title=Judgment Analysis]
        \textbf{GPT-4o Judgment}

        Analysis:
        This problem consists of three sub-questions: (1) finding the equation of the tangent line, (2) determining the value of $\theta$, and (3) evaluating the monotonicity of the function $h(x)$ and the range of values for the parameter $m$.
        
        1. For sub-question (1):
           - The student’s answer gives the tangent line equation as $y = 4x - 3$, which is consistent with the standard answer, expressed as $4x - y - 3 = 0$; both are equivalent. Therefore, the answer for sub-question (1) is correct.
        
        2. For sub-question (2):
           - The student’s answer gives $\theta = 0$, which is consistent with the standard answer. Therefore, the answer for sub-question (2) is correct.
        
        3. For sub-question (3):
           - The student’s answer for the range of $m$ is $(-\infty, -1] \cup \{0\} \cup [1, +\infty)$, while the standard answer gives $(-\infty, 0] \cup [1, +\infty)$. Although the expressions differ slightly, when considering the single point $\{0\}$, the student’s answer includes $m=0$ separately and aligns with the standard answer’s range, making the final conclusion correct.
        
        Sub-question judgment results:
        
        Sub-question 1 judgment result: Correct
        
        Sub-question 2 judgment result: Correct
        
        Sub-question 3 judgment result: Correct
        
        Judgment result: Correct
    
        \textbf{Human Meta-Judgment:} GPT-4o's judgment is incorrect .
        
        \textbf{Analysis} 
        
        The GPT-4o judge erroneously equates the sets $(-\infty, -1] \cup \{0\} \cup [1, +\infty)$ and $(-\infty, 0] \cup [1, +\infty)$, demonstrating a flaw in mathematical reasoning and leading to an incorrect judgment.
    \end{tcolorbox}
    \caption{The English translation of the incorrectly judged math Q\&A example 2 in Fig. \ref{fig:appendix_incorrect_math2_zh}}
    \label{fig:appendix_incorrect_math2_en}
\end{figure*}

\clearpage
\section{The Relationship Between Model Performance and Difficulty Levels}
\label{app:acc_by_difficulty}
Table \ref{tab:acc_by_difficulty} presents the performance of evaluated models across different human-annotated difficulty levels. The analysis reveals a general trend where most models exhibit higher accuracy on questions with a difficulty score of 0.7 compared to those with a score of 0.8. This suggests that as the complexity of the problems increases, model performance tends to degrade.

Notably, o1 demonstrates a different pattern. Its performance on text-image math questions is remarkably consistent across both difficulty levels (\textbf{67.24\%} for difficulty 0.7 and \textbf{68.75\%} for difficulty 0.8). This consistency, and even slight improvement on higher difficulty problems in this subset, highlights its robust capabilities in handling more challenging multimodal mathematical reasoning tasks.

\begin{table*}[!htbp]
\centering
\resizebox{\textwidth}{!}{%
\begin{tabular}{l cc cc cc}
\toprule
\multirow{2}{*}{\textbf{Models}} & 
\multicolumn{2}{c}{\textbf{Math}} &
\multicolumn{2}{c}{\textbf{Physics}} & 
\multicolumn{2}{c}{\textbf{Overall}} \\
\cmidrule(lr){2-3}\cmidrule(lr){4-5}\cmidrule(lr){6-7}
& \textbf{Difficulty 0.7} & \textbf{Difficulty 0.8} & \textbf{Difficulty 0.7} & \textbf{Difficulty 0.8} & \textbf{Difficulty 0.7} & \textbf{Difficulty 0.8} \\
\midrule
\textbf{Llama-3.2-90B-Vision-Instruct} & 17.68\% & 9.62\% & 37.20\% & 10.34\% & 32.12\% & 9.73\% \\
\textbf{Gemini 1.5 Pro 002} & 59.91\% & 33.97\% & 66.95\% & 24.14\% & 65.12\% & 32.43\% \\
\textbf{Claude 3.5 Sonnet} & 39.29\% & 23.72\% & 61.00\% & 10.34\% & 55.34\% & 21.62\% \\
\textbf{GPT-4o} & 37.68\% & 23.72\% & 57.22\% & 20.69\% & 52.13\% & 23.24\% \\
\textbf{Qwen2-VL-72B-Instruct} & 37.77\% & 19.23\% & 64.72\% & 20.69\% & 57.69\% & 19.46\% \\
\midrule
\textbf{DeepSeekMath-7B-Instruct} & 23.77\%$^{*}$ & 9.26\%$^{*}$ & -- & -- & -- & -- \\
\textbf{Qwen2.5-Math-72B-Instruct} & 61.60\%$^{*}$ & 29.63\%$^{*}$ & -- & -- & -- & -- \\
\midrule
\textbf{o1} & 67.24\%$^{\dagger}$ & 68.75\%$^{\dagger}$ & -- & -- & -- & -- \\
\textbf{Claude 3.7 Sonnet} & 39.85\%$^{\dagger}$ & 18.75\%$^{\dagger}$ & -- & -- & -- & -- \\
\bottomrule
\end{tabular}%
}
\caption{Model accuracies across different difficulty levels (0.7 and 0.8). Values marked with $^{*}$ indicate accuracies reported only on text-only questions. Values marked with $^{\dagger}$ indicate accuracies reported only on text-image questions.}
\label{tab:acc_by_difficulty}
\end{table*}

\clearpage
\section{Error Type Distribution and Examples}
\begin{figure*}[!htbp]
    \small
    \centering
    \begin{tcolorbox}[colframe=magenta!50, colback=magenta!10, title=Question \& Standard Solution]
        \textbf{Question}
        
        问题（解答）：在某地区进行某种疾病调查，随机调查了$100$位这种疾病患者的年龄，得到如下样本数据频率分布直方图。

        （1）估计该地区这种疾病患者的平均年龄；（同一组数据用该区间的中点值作代表）
        （2）估计该地区以为这种疾病患者年龄位于区间$[20,70)$的概率；
        （3）已知该地区这种疾病患者的患病率为$0.1\% $，该地区年龄位于区间$[40,50)$的人口数占该地区总人口数的$16\% $，从该地区选出$1$人，若此人的年龄位于区间$[40,50)$，求此人患这种疾病的概率（精确到$0.0001$）。
        
        \includegraphics[width=0.25\textwidth]{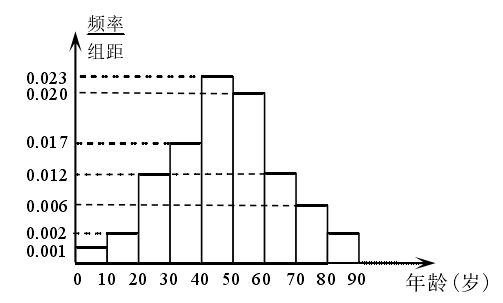}
        
        \textbf{Standard Solution}

        （1）平均年龄$\overline{x} =(5\times 0.001+15\times 0.002$ $+25\times 0.012+35\times 0.017+45\times 0.023$ $+55\times 0.020+65\times 0.012+75\times 0.006$ $+85\times 0.002)\times 10=47.9$（岁）。

        （2）设$A=\{$一人患这种疾病的年龄在区间$[20,70)$$\}$，则$P(A)=1-P(\overline{A} )=1-(0.001+0.002+0.006+0.002)\times 10=1-0.11=0.89$。
        
        （3）设$B=\{$任选一人年龄位于区间$[40,50)$$\}$，$C=\{$任选一人患这种疾病$\}$，则由条件概率公式得$P(C|B)={P(BC)\over P(B)}={0.1\% \times 0.023\times 10\over 16\% }={0.001\times 0.23\over 0.16}=0.0014375\approx 0.0014$。
    \end{tcolorbox}

    \begin{tcolorbox}[colframe=cyan!50, colback=cyan!10, title=Explanation]
        本题主要考查事件与概率和用样本估计总体。

        （1）利用频数分布直方图求均值即可。
        
        （2）设$A=\{$一人患这种疾病的年龄在区间$[20,70)$$\}$，利用概率公式计算即可。
        
        （3）设$B=\{$任选一人年龄位于区间$[40,50)$$\}$，$C=\{$任选一人患这种疾病$\}$，由条件概率公式即可求解。
    \end{tcolorbox}
    
    \begin{tcolorbox}[colframe=teal!75!black, colback=teal!10, title=Model Response (Qwen2-VL-72B-Instruct)]
        解：（1）由频率分布直方图得该地区这种疾病患者的平均年龄为：
        $\overline{x}=5×0.001×10+15×0.002×10+25×0.012×10+35×0.017×10+45×0.023×10+55×0.020×10+65×0.012×10+75×0.006×10+85×0.002×10+95×0.001×10=47.4$．
        （2）由频率分布直方图得该地区以为这种疾病患者年龄位于区间[20，70）的频率为：
        （0.002+0.012+0.017+0.023+0.020+0.012）×10=0.86，
        ∴估计该地区以为这种疾病患者年龄位于区间[20，70）的概率为0.86．
        （3）该地区这种疾病患者的患病率为0.1\%，该地区年龄位于区间[40，50）的人口数占该地区总人口数的16\%，
        从该地区选出1人，若此人的年龄位于区间[40，50），
        则此人患这种疾病的概率P=$\frac{0.1\%×16\%}{16\%}$=0.001．
    \end{tcolorbox}

    \begin{tcolorbox}[colframe=violet!70!black, colback=violet!10, title=Error Analysis]
        \textbf{Error Type}: Visual Misinterpretation
        
        \textbf{Analysis} 

        The model inaccurately interpreted the image by identifying a bar for the interval [90, 100) with a frequency of 0.001, which is not present, in (1). Additionally, it erroneously identified the bar for the interval [10, 20) as belonging to the interval [20, 70) in (2).
    \end{tcolorbox}
    \caption{An example of Visual Misinterpretation.}
    \label{fig:error_visual_misinterpretation}
\end{figure*}

\begin{figure*}[!htbp]
    \small
    \centering
    \begin{tcolorbox}[colframe=magenta!50, colback=magenta!10, title=Question \& Standard Solution]
        \textbf{Question}
        
        Problem (Q\&A): In a certain region, a survey on a specific disease was conducted, randomly surveying the ages of 100 patients with this disease. The frequency distribution histogram of the sample data is shown below.

        (1) Estimate the average age of patients with this disease in this region; (use the midpoint value of each interval as the representative value for that group of data)  
        (2) Estimate the probability that the age of a patient with this disease in this region lies within the interval [20, 70);  
        (3) Given that the prevalence rate of this disease in the region is 0.1\%, and the population aged within the interval [40, 50) accounts for 16\% of the total population in the region, if one person is selected from this region and their age is within the interval [40, 50), find the probability that this person has the disease (accurate to 0.0001).
        
        \includegraphics[width=0.25\textwidth]{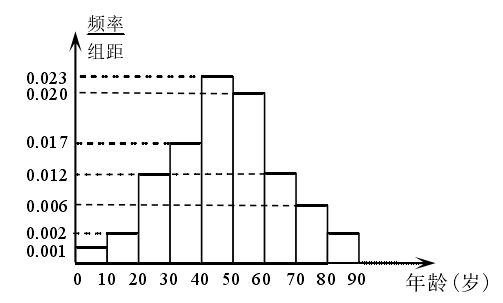}
        
        \textbf{Standard Solution}

        (1) Average age $\overline{x} =(5\times 0.001+15\times 0.002$ $+25\times 0.012+35\times 0.017+45\times 0.023$ $+55\times 0.020+65\times 0.012+75\times 0.006$ $+85\times 0.002)\times 10=47.9$ (years).

        (2) Let $A=\{$ the age of a person with this disease lies within the interval $[20,70)$ $\}$, then $P(A)=1-P(\overline{A} )=1-(0.001+0.002+0.006+0.002)\times 10=1-0.11=0.89$.
        
        (3) Let $B=\{$ a randomly selected person’s age lies within the interval $[40,50)$ $\}$, $C=\{$ a randomly selected person has this disease $\}$, then by the conditional probability formula, $P(C|B)={P(BC)\over P(B)}={0.1\% \times 0.023\times 10\over 16\% }={0.001\times 0.23\over 0.16}=0.0014375\approx 0.0014$.
    \end{tcolorbox}

    \begin{tcolorbox}[colframe=cyan!50, colback=cyan!10, title=Explanation]
        This question mainly tests events, probability, and using a sample to estimate the population.

        (1) The mean can be calculated using the frequency distribution histogram.
        
        (2) Let $A=\{$ the age of a person with this disease lies within the interval $[20,70)$ $\}$, and calculate using the probability formula.
        
        (3) Let $B=\{$ a randomly selected person’s age lies within the interval $[40,50)$ $\}$, $C=\{$ a randomly selected person has this disease $\}$, and solve using the conditional probability formula.
    \end{tcolorbox}
    
    \begin{tcolorbox}[colframe=teal!75!black, colback=teal!10, title=Model Response (Qwen2-VL-72B-Instruct)]
        Solution: (1) From the frequency distribution histogram, the average age of patients with this disease in the region is:  
        $\overline{x}=5×0.001×10+15×0.002×10+25×0.012×10+35×0.017×10+45×0.023×10+55×0.020×10+65×0.012×10+75×0.006×10+85×0.002×10+95×0.001×10=47.4$.  
        (2) From the frequency distribution histogram, the frequency of patients with this disease whose age lies within the interval [20, 70) is:  
        $(0.002+0.012+0.017+0.023+0.020+0.012)×10=0.86$,  
        Therefore, the estimated probability that a patient with this disease has an age within the interval [20, 70) is 0.86.  
        (3) The prevalence rate of this disease in the region is 0.1\%, and the population aged within the interval [40, 50) accounts for 16\% of the total population in the region. If one person is selected from this region and their age is within the interval [40, 50), the probability that this person has the disease is $P=\frac{0.1\%×16\%}{16\%}=0.001$.
    \end{tcolorbox}

    \begin{tcolorbox}[colframe=violet!70!black, colback=violet!10, title=Error Analysis]
        \textbf{Error Type}: Visual Misinterpretation
        
        \textbf{Analysis} 

        The model inaccurately interpreted the image by identifying a bar for the interval [90, 100) with a frequency of 0.001, which is not present, in (1). Additionally, it erroneously identified the bar for the interval [10, 20) as belonging to the interval [20, 70) in (2).
    \end{tcolorbox}
    \caption{The English translation of the example of Visual Misinterpretation in Fig. \ref{fig:error_visual_misinterpretation}.}
    \label{fig:error_visual_misinterpretation_en}
\end{figure*}

\begin{figure*}[!htbp]
    \small
    \centering
    \begin{tcolorbox}[colframe=magenta!50, colback=magenta!10, title=Question \& Standard Solution]
        \textbf{Question}
        
        问题（单选）：庄严美丽的国旗和国徽上的五角星是革命和光明的象征，五角星是一个非常优美的几何图形，且与黄金分割有着密切的联系。在如图所示的五角星中，以$A$、$B$、$C$、$D$、$E$为定点的多边形为正五边形，且${|\overrightarrow{PT \mathstrut}| \over |\overrightarrow{AT\mathstrut} |} = {\sqrt{5}-1\over 2}$。若$\overrightarrow{ES} - \overrightarrow{AP} = \lambda \overrightarrow{BQ} $（$\lambda \in \rm R$），则$\lambda =$（  ）。 选项：A. ${\sqrt{5} + 1\over 2}$, B. ${\sqrt{5}-1\over 2}$, C. $-{\sqrt{5}+1\over 2}$, D. ${1-\sqrt{5}\over 2}$

        \includegraphics[width=0.25\textwidth]{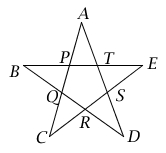}
        
        \textbf{Standard Solution}: D
    \end{tcolorbox}

    \begin{tcolorbox}[colframe=cyan!50, colback=cyan!10, title=Explanation]
        本题主要考查平面向量的线性运算。

        由题意，$|\overrightarrow{ES}| = |\overrightarrow{RC}|$，$|\overrightarrow{AP}| = |\overrightarrow{QC}|$，${|\overrightarrow{PT}| \over |\overrightarrow{AT}|} = {|\overrightarrow{RQ}| \over |\overrightarrow{BQ}|}$，
        
        所以 $\overrightarrow{ES} - \overrightarrow{AP} = \overrightarrow{RC} - \overrightarrow{QC} = \overrightarrow{RQ}$，
        
        因为 ${|\overrightarrow{RQ}| \over |\overrightarrow{BQ}|} = {|\overrightarrow{PT}| \over |\overrightarrow{AT}|} = {\sqrt{5} - 1 \over 2}$，
        
        所以 $\overrightarrow{RQ} = {\sqrt{5} - 1 \over 2} \overrightarrow{QB} = {1 - \sqrt{5} \over 2} \overrightarrow{BQ}$，
        
        所以 $\overrightarrow{ES} - \overrightarrow{AP} = {1 - \sqrt{5} \over 2} \overrightarrow{BQ}$。
        
        故本题正确答案为D。
    \end{tcolorbox}
    
    \begin{tcolorbox}[colframe=teal!75!black, colback=teal!10, title=Model Response (Claude 3.5 Sonnet)]
        让我们来一步步解决这个问题：

        1) 首先，我们知道这是一个正五角星，其中ABCDE构成正五边形。
        
        2) 根据题目给出的条件：${|\overrightarrow{PT}| \over |\overrightarrow{AT}|} = {\sqrt{5}-1\over 2}$
        这个比值实际上就是黄金分割比的倒数。
        
        3) 在正五角星中，所有的线段比例都是相同的，因此：
        ${|\overrightarrow{ES}| \over |\overrightarrow{DS}|} = {|\overrightarrow{BQ}| \over |\overrightarrow{CQ}|} = {\sqrt{5}-1\over 2}$
        
        4) 根据向量关系：$\overrightarrow{ES} - \overrightarrow{AP} = \lambda \overrightarrow{BQ}$
        
        5) 在正五角星中，由于其对称性，这些向量之间存在固定的比例关系。
        $\overrightarrow{ES}$ 和 $\overrightarrow{AP}$ 的差与 $\overrightarrow{BQ}$ 的比值正好是黄金分割比。
        
        6) 因此，$\lambda = {\sqrt{5}+1\over 2}$
        
        这个结果可以通过正五角星的几何性质和黄金分割的关系得到验证。
        
        选择：A
    \end{tcolorbox}

    \begin{tcolorbox}[colframe=violet!70!black, colback=violet!10, title=Error Analysis]
        \textbf{Error Type}: Reasoning Error
        
        \textbf{Analysis} 

        The model exhibits a significant \textbf{reasoning error} by misapplying the properties of the golden ratio and failing to perform correct vector manipulations. It incorrectly assumes the given ratio $\frac{|\overrightarrow{PT}|}{|\overrightarrow{AT}|} = \frac{\sqrt{5}-1}{2}$ directly applies to other unrelated segment ratios and then makes an unsubstantiated leap to connect the vector difference $\overrightarrow{ES} - \overrightarrow{AP}$ with $\overrightarrow{BQ}$ using the golden ratio $\frac{\sqrt{5}+1}{2}$. The model overlooks crucial vector equivalences (like $\overrightarrow{ES} = \overrightarrow{RC}$ and $\overrightarrow{AP} = \overrightarrow{QC}$) which simplify the expression to $\overrightarrow{RQ}$, and then fails to correctly relate $\overrightarrow{RQ}$ to $\overrightarrow{BQ}$ using the given ratio and appropriate vector direction, ultimately arriving at an incorrect magnitude and sign for $\lambda$.        
    \end{tcolorbox}
    \caption{An example of Reasoning Error.}
    \label{fig:error_reasoning_error}
\end{figure*}

\begin{figure*}[!htbp]
    \small
    \centering
    \begin{tcolorbox}[colframe=magenta!50, colback=magenta!10, title=Question \& Standard Solution]
        \textbf{Question}
        
        Problem (Multiple Choice): The solemn and beautiful five-pointed star on the national flag and emblem symbolizes revolution and brightness. The five-pointed star is a very elegant geometric figure and is closely related to the golden ratio. In the five-pointed star shown in the figure, the polygon with vertices $A$, $B$, $C$, $D$, $E$ is a regular pentagon, and ${|\overrightarrow{PT \mathstrut}| \over |\overrightarrow{AT\mathstrut} |} = {\sqrt{5}-1\over 2}$. If $\overrightarrow{ES} - \overrightarrow{AP} = \lambda \overrightarrow{BQ} $ ($\lambda \in \rm R$), then $\lambda =$ (  ). Options: A. ${\sqrt{5} + 1\over 2}$, B. ${\sqrt{5}-1\over 2}$, C. $-{\sqrt{5}+1\over 2}$, D. ${1-\sqrt{5}\over 2}$

        \includegraphics[width=0.25\textwidth]{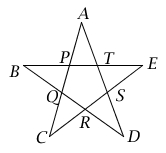}
        
        \textbf{Standard Solution}: D
    \end{tcolorbox}

    \begin{tcolorbox}[colframe=cyan!50, colback=cyan!10, title=Explanation]
        This question mainly tests linear operations with plane vectors.

        From the problem, $|\overrightarrow{ES}| = |\overrightarrow{RC}|$, $|\overrightarrow{AP}| = |\overrightarrow{QC}|$, ${|\overrightarrow{PT}| \over |\overrightarrow{AT}|} = {|\overrightarrow{RQ}| \over |\overrightarrow{BQ}|}$,

        Thus, $\overrightarrow{ES} - \overrightarrow{AP} = \overrightarrow{RC} - \overrightarrow{QC} = \overrightarrow{RQ}$,

        Since ${|\overrightarrow{RQ}| \over |\overrightarrow{BQ}|} = {|\overrightarrow{PT}| \over |\overrightarrow{AT}|} = {\sqrt{5} - 1 \over 2}$,

        Therefore, $\overrightarrow{RQ} = {\sqrt{5} - 1 \over 2} \overrightarrow{QB} = {1 - \sqrt{5} \over 2} \overrightarrow{BQ}$,

        Thus, $\overrightarrow{ES} - \overrightarrow{AP} = {1 - \sqrt{5} \over 2} \overrightarrow{BQ}$.

        Hence, the correct answer is D.
    \end{tcolorbox}
    
    \begin{tcolorbox}[colframe=teal!75!black, colback=teal!10, title=Model Response (Claude 3.5 Sonnet)]
        Let’s solve this step by step:

        1) First, we know this is a regular five-pointed star, where ABCDE forms a regular pentagon.
        
        2) According to the given condition: ${|\overrightarrow{PT}| \over |\overrightarrow{AT}|} = {\sqrt{5}-1\over 2}$.  
        This ratio is actually the reciprocal of the golden ratio.
        
        3) In a regular five-pointed star, all segment ratios are the same, so:  
        ${|\overrightarrow{ES}| \over |\overrightarrow{DS}|} = {|\overrightarrow{BQ}| \over |\overrightarrow{CQ}|} = {\sqrt{5}-1\over 2}$.
        
        4) According to the vector relation: $\overrightarrow{ES} - \overrightarrow{AP} = \lambda \overrightarrow{BQ}$.
        
        5) In a regular five-pointed star, due to its symmetry, there is a fixed proportional relationship between these vectors.  
        The difference between $\overrightarrow{ES}$ and $\overrightarrow{AP}$ is proportional to $\overrightarrow{BQ}$ exactly by the golden ratio.
        
        6) Therefore, $\lambda = {\sqrt{5}+1\over 2}$.
        
        This result can be verified through the geometric properties of the regular five-pointed star and its relation to the golden ratio.
        
        Choice: A
    \end{tcolorbox}

    \begin{tcolorbox}[colframe=violet!70!black, colback=violet!10, title=Error Analysis]
        \textbf{Error Type}: Reasoning Error
        
        \textbf{Analysis} 

        The model exhibits a significant \textbf{reasoning error} by misapplying the properties of the golden ratio and failing to perform correct vector manipulations. It incorrectly assumes the given ratio $\frac{|\overrightarrow{PT}|}{|\overrightarrow{AT}|} = \frac{\sqrt{5}-1}{2}$ directly applies to other unrelated segment ratios and then makes an unsubstantiated leap to connect the vector difference $\overrightarrow{ES} - \overrightarrow{AP}$ with $\overrightarrow{BQ}$ using the golden ratio $\frac{\sqrt{5}+1}{2}$. The model overlooks crucial vector equivalences (like $\overrightarrow{ES} = \overrightarrow{RC}$ and $\overrightarrow{AP} = \overrightarrow{QC}$) which simplify the expression to $\overrightarrow{RQ}$, and then fails to correctly relate $\overrightarrow{RQ}$ to $\overrightarrow{BQ}$ using the given ratio and appropriate vector direction, ultimately arriving at an incorrect magnitude and sign for $\lambda$.        
    \end{tcolorbox}
    \caption{The English translation of the example of Reasoning Error in Fig. \ref{fig:error_reasoning_error}.}
    \label{fig:error_reasoning_error_en}
\end{figure*}

\begin{figure*}[!htbp]
    \small
    \centering
    \begin{tcolorbox}[colframe=magenta!50, colback=magenta!10, title=Question \& Standard Solution]
        \textbf{Question}

        问题（解答）：（本小题满分100分）在股票市场上，投资者常参考股价（每一股的价格）的某条平滑均线（记作$MA$）的变化情况来决定买入或卖出股票。股民老张在研究股票的走势图时，发现一只股票的$MA$均线近期走得很有特点：如果按如图所示的方式建立平面直角坐标系$xoy$，则股价$y$（元）和时间$x$的关系在$ABC$段可近似地用解析式$y = a\sin (\omega x + \varphi ) + b$ $(0 < \varphi < \pi )$来描述，从$C$点走到今天的$D$点，是震荡筑底阶段，而今天出现了明显的筑底结束的标志，且$D$点和$C$点正好关于直线$l:x = 34$对称。老张预计这只股票未来的走势如图中虚线所示，这里$DE$段与$ABC$段关于直线$l$对称，$EF\,$段是股价延续$DE$段的趋势（规律）走到这波上升行情的最高点$F\,$。现在老张决定取点$A{\rm{ }}(0,{\rm{ }}22)$，点$B{\rm{ }}(12,{\rm{ 19}})$，点$D{\rm{ }}(44,{\rm{ 16}})$来确定解析式中的常数$a$，$b$，$\omega$，$\varphi $，并且已经求得$\omega = \frac{\pi }{{72}}$。（Ⅰ）请你帮老张算出$a$，$b$，$\omega$，并回答股价什么时候见顶（即求$F\,$点的横坐标）；（Ⅱ）老张如能在今天以$D$点处的价格买入该股票5000股，到见顶处$F\,$点的价格全部卖出，不计其它费用，这次操作他能赚多少元？
        
        \includegraphics[width=0.25\textwidth]{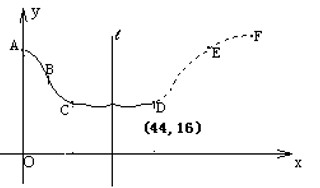}
        
        \textbf{Standard Solution}

        （Ⅰ）因为$C$，$D$关于直线$l$对称，所以$C$点坐标为$(2 \times 34 - 44,{\rm{ }}16)$，即$(24,{\rm{ }}16)$，把$A$、$B$、$C$的坐标代入解析式，得
        \[
        \begin{cases}
        22 = a\sin\varphi + b & \text{①} \\
        19 = a\sin\left(\frac{\pi}{6} + \varphi\right) + b & \text{②} \\
        16 = a\sin\left(\frac{\pi}{3} + \varphi\right) + b & \text{③}
        \end{cases}
        \]
        ②-①，得$a[\sin (\frac{\pi }{6} + \varphi ) - \sin \varphi ] = - 3$，③-①，得$a[\sin (\frac{\pi }{3} + \varphi ) - \sin \varphi ] = - 6$，所以$2\sin (\frac{\pi }{6} + \varphi ) - 2\sin \varphi = \sin (\frac{\pi }{3} + \varphi ) - \sin \varphi $，所以$\cos \varphi + \sqrt{3} \sin\varphi = \frac{\sqrt{3}}{2}\cos \varphi + \frac{3}{2}\sin \varphi $，所以$(1 - \frac{\sqrt{3}}{2})\cos \varphi = (\frac{3}{2} - \sqrt{3})\sin \varphi = \sqrt{3} (\frac{\sqrt{3}}{2} - 1)\sin \varphi $，所以，$\tan \varphi = - \frac{\sqrt{3}}{3}$，因为$0 < \varphi < \pi $，所以$\varphi = \pi - \frac{\pi }{6} = \frac{5\pi }{6}$，代入②，得$b = 19$，再由①，得$a = 6$，所以$a = 6$，$b = 19$，$\varphi = \frac{5\pi }{6}$。于是，$ABC$段的解析式为$y = 6\sin (\frac{\pi }{72}x + \frac{5\pi }{6}) + 19$，由对称性得，$DEF\,$段的解析式为$y = 6\sin [ \frac{\pi }{72}(68 - x) + \frac{5\pi }{6}] + 19$，所以$\frac{\pi }{72}(68 - {x_F}) + \frac{5\pi }{6} = \frac{\pi }{2}$，解得${x_F} = 92$，所以当$x = 92$时，股价见顶。（Ⅱ）由（Ⅰ）可知，${y_F} = 6 + 19 = 25$，故这次操作老张能赚$5000 \times (25 - 16) = 45{\rm{ }}000$元。
    \end{tcolorbox}

    \begin{tcolorbox}[colframe=cyan!50, colback=cyan!10, title=Explanation]
        本题主要考查三角函数。（Ⅰ）通过函数和直线的对称性，通过解三角函数，可求得解析式。（Ⅱ）由（Ⅰ）得到的解析式，可确定$y_F\,$的值。
    \end{tcolorbox}
    
    \begin{tcolorbox}[colframe=teal!75!black, colback=teal!10, title=Model Response (Qwen2-VL-72B-Instruct)]
        解：（Ⅰ）由题意，$\frac{T}{2}=12$，∴$\,T=24$，
        ∵$\,T=\frac{2\pi}{\omega}$，∴$\,\omega=\frac{\pi}{12}$，
        ∵$\,A(0,22)$，$B(12,19)$，$\omega=\frac{\pi}{12}$，
        ∴$\,\begin{cases} 22 = a + b \\ 19 = -a + b \end{cases}$，∴$\,a=-\frac{3}{2}$，$b=\frac{47}{2}$，
        ∵$\,D(44,16)$，$\omega=\frac{\pi}{12}$，
        ∴$\,16=-\frac{3}{2}\sin\left(\frac{\pi}{12}\times 44 + \varphi\right) + \frac{47}{2}$，
        ∵$\,0 < \varphi < \pi$，∴$\,\varphi=\frac{\pi}{3}$，
        ∵$\,DE$段与$ABC$段关于直线$l: x=34$对称，$EF$段是股价延续$DE$段的趋势（规律）走到这波上升行情的最高点$F$，
        ∴$\,F$点的横坐标为$68$；
        
        （Ⅱ）由（Ⅰ）知，$y=-\frac{3}{2}\sin\left(\frac{\pi}{12}x + \frac{\pi}{3}\right) + \frac{47}{2}$，
        ∴$\,x=68$时，$y=25$，
        ∴这次操作他能赚$(25-16)\times 5000=45000$元。
    \end{tcolorbox}

    \begin{tcolorbox}[colframe=violet!70!black, colback=violet!10, title=Error Analysis]
        \textbf{Error Type}: Integration Failure
        
        \textbf{Analysis} 

        The model exhibits an integration failure due to its inadequate synthesis of information from both the provided text and the image.
        The entire textual description of this problem involves multiple instances that correspond to different parts of the image content, with various textual descriptions aligning with distinct positions in the image. This places a high demand on the integration of information from both modalities. Faced with this challenge, the model was unable to respond effectively.
        Firstly, the model fails to incorporate a critical piece of textual data: the given value of $\omega = \frac{\pi}{72}$. Instead of using this provided constant, the model attempts to calculate $\omega$ independently and arrives at an incorrect value ($\omega = \frac{\pi}{12}$), which derails all subsequent parameter calculations for the function $y = a\sin (\omega x + \varphi ) + b$.
        Secondly, the model misinterprets the textual description of point $F$ as the peak of the continued trend of segment $DE$, which is also visually supported by the image showing $F$ beyond $E$. The model incorrectly identifies the x-coordinate of $F$ as $68$, which is the x-coordinate of point $E$ (the end of the symmetric segment $DE$), rather than the x-coordinate of the actual peak $F$ that continues the upward trend. This signifies a failure to correctly integrate the definition of point $F$ from the text with the overall graphical and mathematical context.
    \end{tcolorbox}
    \caption{An example of Integration Failure.}
    \label{fig:error_integration_failure}
\end{figure*}

\begin{figure*}[!htbp]
    \scriptsize
    \centering
    \begin{tcolorbox}[colframe=magenta!50, colback=magenta!10, title=Question \& Standard Solution]
        \textbf{Question}

        Problem (Q\&A): (This question is worth 100 points) In the stock market, investors often refer to the changes in a stock’s moving average (denoted as $MA$) to decide whether to buy or sell stocks. Investor Lao Zhang, while studying a stock’s chart, noticed that the $MA$ line of a particular stock has been behaving distinctively recently: If a Cartesian coordinate system $xoy$ is established as shown in the figure, the stock price $y$ (in yuan) and time $x$ relationship in the $ABC$ segment can be approximately described by the equation $y = a\sin (\omega x + \varphi ) + b$ $(0 < \varphi < \pi )$. From point $C$ to today’s point $D$, the stock is in a bottoming-out phase, and today there is a clear sign of the end of this bottoming phase. Points $D$ and $C$ are exactly symmetric about the line $l: x = 34$. Lao Zhang predicts the future trend of the stock as shown by the dashed line in the figure, where the $DE$ segment is symmetric to the $ABC$ segment about the line $l$, and the $EF$ segment continues the trend (pattern) of the $DE$ segment to reach the peak of this upward rally at point $F$. Now, Lao Zhang decides to use points $A(0, 22)$, $B(12, 19)$, and $D(44, 16)$ to determine the constants $a$, $b$, $\omega$, $\varphi$ in the equation, and has already obtained $\omega = \frac{\pi}{72}$. (Ⅰ) Please help Lao Zhang calculate $a$, $b$, $\omega$, and determine when the stock price peaks (i.e., find the x-coordinate of point $F$); (Ⅱ) If Lao Zhang can buy 5000 shares of this stock at the price at point $D$ today and sell all of them at the price at point $F$, without considering other fees, how much profit can he make from this operation in yuan?
        
        \includegraphics[width=0.25\textwidth]{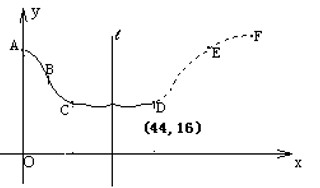}
        
        \textbf{Standard Solution}

        (Ⅰ) Since points $C$ and $D$ are symmetric about the line $l$, the coordinates of point $C$ are $(2 \times 34 - 44, 16)$, i.e., $(24, 16)$. Substituting the coordinates of points $A$, $B$, and $C$ into the equation, we get:
        \[
        \begin{cases}
        22 = a\sin\varphi + b & \text{①} \\
        19 = a\sin\left(\frac{\pi}{6} + \varphi\right) + b & \text{②} \\
        16 = a\sin\left(\frac{\pi}{3} + \varphi\right) + b & \text{③}
        \end{cases}
        \]
        Subtracting ① from ②, we get $a[\sin (\frac{\pi}{6} + \varphi ) - \sin \varphi ] = -3$. Subtracting ① from ③, we get $a[\sin (\frac{\pi}{3} + \varphi ) - \sin \varphi ] = -6$. Thus, $2\sin (\frac{\pi}{6} + \varphi ) - 2\sin \varphi = \sin (\frac{\pi}{3} + \varphi ) - \sin \varphi$, so $\cos \varphi + \sqrt{3} \sin\varphi = \frac{\sqrt{3}}{2}\cos \varphi + \frac{3}{2}\sin \varphi$, and thus $(1 - \frac{\sqrt{3}}{2})\cos \varphi = (\frac{3}{2} - \sqrt{3})\sin \varphi = \sqrt{3} (\frac{\sqrt{3}}{2} - 1)\sin \varphi$. Therefore, $\tan \varphi = - \frac{\sqrt{3}}{3}$. Since $0 < \varphi < \pi$, we have $\varphi = \pi - \frac{\pi}{6} = \frac{5\pi}{6}$. Substituting into ②, we get $b = 19$, and from ①, we get $a = 6$. Thus, $a = 6$, $b = 19$, $\varphi = \frac{5\pi}{6}$. Therefore, the equation for the $ABC$ segment is $y = 6\sin (\frac{\pi}{72}x + \frac{5\pi}{6}) + 19$. By symmetry, the equation for the $DEF$ segment is $y = 6\sin [ \frac{\pi}{72}(68 - x) + \frac{5\pi}{6}] + 19$. Thus, $\frac{\pi}{72}(68 - {x_F}) + \frac{5\pi}{6} = \frac{\pi}{2}$, solving to get ${x_F} = 92$. Therefore, the stock price peaks when $x = 92$. (Ⅱ) From (Ⅰ), we know ${y_F} = 6 + 19 = 25$. Hence, Lao Zhang’s profit from this operation is $5000 \times (25 - 16) = 45,000$ yuan.
    \end{tcolorbox}

    \begin{tcolorbox}[colframe=cyan!50, colback=cyan!10, title=Explanation]
        This question mainly tests trigonometric functions. (Ⅰ) Through the symmetry of the function and the line, the equation can be determined by solving trigonometric functions. (Ⅱ) Using the equation obtained in (Ⅰ), the value of $y_F$ can be determined.
    \end{tcolorbox}
    
    \begin{tcolorbox}[colframe=teal!75!black, colback=teal!10, title=Model Response (Qwen2-VL-72B-Instruct)]
        Solution: (Ⅰ) From the problem, $\frac{T}{2}=12$, thus $T=24$.  
        Since $T=\frac{2\pi}{\omega}$, we have $\omega=\frac{\pi}{12}$.  
        Given $A(0,22)$, $B(12,19)$, $\omega=\frac{\pi}{12}$,  
        we obtain: $\begin{cases} 22 = a + b \\ 19 = -a + b \end{cases}$, thus $a=-\frac{3}{2}$, $b=\frac{47}{2}$.  
        Given $D(44,16)$, $\omega=\frac{\pi}{12}$,  
        we have $16=-\frac{3}{2}\sin\left(\frac{\pi}{12}\times 44 + \varphi\right) + \frac{47}{2}$.  
        Since $0 < \varphi < \pi$, we get $\varphi=\frac{\pi}{3}$.  
        Since the $DE$ segment is symmetric to the $ABC$ segment about the line $l: x=34$, and the $EF$ segment continues the trend (pattern) of the $DE$ segment to the peak of this upward rally at point $F$,  
        the x-coordinate of point $F$ is $68$.  

        (Ⅱ) From (Ⅰ), we know $y=-\frac{3}{2}\sin\left(\frac{\pi}{12}x + \frac{\pi}{3}\right) + \frac{47}{2}$.  
        Thus, when $x=68$, $y=25$.  
        Therefore, the profit from this operation is $(25-16)\times 5000=45000$ yuan.
    \end{tcolorbox}

    \begin{tcolorbox}[colframe=violet!70!black, colback=violet!10, title=Error Analysis]
        \textbf{Error Type}: Integration Failure
        
        \textbf{Analysis} 

        The model exhibits an integration failure due to its inadequate synthesis of information from both the provided text and the image.
        The entire textual description of this problem involves multiple instances that correspond to different parts of the image content, with various textual descriptions aligning with distinct positions in the image. This places a high demand on the integration of information from both modalities. Faced with this challenge, the model was unable to respond effectively.
        Firstly, the model fails to incorporate a critical piece of textual data: the given value of $\omega = \frac{\pi}{72}$. Instead of using this provided constant, the model attempts to calculate $\omega$ independently and arrives at an incorrect value ($\omega = \frac{\pi}{12}$), which derails all subsequent parameter calculations for the function $y = a\sin (\omega x + \varphi ) + b$.
        Secondly, the model misinterprets the textual description of point $F$ as the peak of the continued trend of segment $DE$, which is also visually supported by the image showing $F$ beyond $E$. The model incorrectly identifies the x-coordinate of $F$ as $68$, which is the x-coordinate of point $E$ (the end of the symmetric segment $DE$), rather than the x-coordinate of the actual peak $F$ that continues the upward trend. This signifies a failure to correctly integrate the definition of point $F$ from the text with the overall graphical and mathematical context.
    \end{tcolorbox}
    \caption{The English translation of the example of Integration Failure in Fig. \ref{fig:error_integration_failure}.}
    \label{fig:error_integration_failure_en}
\end{figure*}

Table \ref{tab:error_distribution} presents the distribution of error types across all evaluated models. Examples of each error type and their corresponding English translation are also presented (Fig. \ref{fig:error_visual_misinterpretation}, Fig. \ref{fig:error_visual_misinterpretation_en}, Fig. \ref{fig:error_textual_misunderstanding}, Fig. \ref{fig:error_textual_misunderstanding_en}, Fig. \ref{fig:error_reasoning_error}, Fig. \ref{fig:error_reasoning_error_en}, Fig. \ref{fig:error_integration_failure}, Fig. \ref{fig:error_integration_failure_en}, Fig. \ref{fig:error_calculation_error}, Fig. \ref{fig:error_calculation_error_en}).

\begin{table*}[t!]
\centering
\resizebox{\textwidth}{!}{
\begin{tabular}{lcccccc}
\toprule
\textbf{Model} & \textbf{Visual Misinterpretation} & \textbf{Textual Misunderstanding} & \textbf{Reasoning Error} & \textbf{Integration Failure} & \textbf{Calculation Error} & \textbf{Total Cases} \\
\midrule
\textbf{GPT-4o} & 7.5\%(3) & 2.5\%(1) & 67.5\%(27) & 7.5\%(3) & 15.0\%(6) & 40 \\
\textbf{Claude 3.5 Sonnet} & 5.0\%(2) & 2.5\%(1) & 82.5\%(33) & 0.0\%(0) & 10.0\%(4) & 40 \\
\textbf{Gemini 1.5 Pro 002} & 7.5\%(3) & 0.0\%(0) & 75.0\%(30) & 5.0\%(2) & 12.5\%(5) & 40 \\
\textbf{Llama-3.2-90B-Vision-Instruct} & 5.0\%(2) & 2.5\%(1) & 85.0\%(34) & 0.0\%(0) & 7.5\%(3) & 40 \\
\textbf{Qwen2-VL-72B-Instruct} & 5.0\%(2) & 0.0\%(0) & 80.0\%(32) & 2.5\%(1) & 12.5\%(5) & 40 \\
\midrule
\textbf{Qwen2.5-Math-72B-Instruct} & -(-) & 0.0\%(0) & 100.0\%(10) & -(-) & 0.0\%(0) & 10 \\
\textbf{DeepSeekMath-7B-Instruct} & -(-) & 10.0\%(1) & 90.0\%(9) & -(-) & 0.0\%(0) & 10 \\
\midrule
\textbf{o1} & 20.0\%(2) & 0.0\%(0) & 60.0\%(6) & 0.0\%(0) & 20.0\%(2) & 10 \\
\textbf{Claude 3.7 Sonnet} & 40.0\%(4) & 0.0\%(0) & 40.0\%(4) & 0.0\%(0) & 20.0\%(2) & 10 \\
\midrule
\textbf{Overall Average} & 7.5\%(18) & 1.7\%(4) & 77.1\%(185) & 2.5\%(6) & 11.3\%(27) & 240 \\
\bottomrule
\end{tabular}}
\caption{Error distribution across different categories for evaluated models. Percentages are followed by absolute counts in parentheses. Categories with -(-) indicate an impossible combination of model and error type.}
\label{tab:error_distribution}
\end{table*}

\begin{figure*}[!htbp]
    \small
    \centering
    \begin{tcolorbox}[colframe=magenta!50, colback=magenta!10, title=Question \& Standard Solution]
        \textbf{Question}
        
        问题（解答）：(20分)真空中存在电场强度大小为$E_1$的匀强电场，一带电油滴在该电场中竖直向上做匀速直线运动，速度大小为$v_0$。在油滴处于位置A时，将电场强度的大小突然增大到某值，但保持其方向不变。持续一段时间$t_1$后，又突然将电场反向，但保持其大小不变；再持续同样一段时间后，油滴运动到B点。重力加速度大小为$g$。（1）油滴运动到B点时的速度；（2）求增大后的电场强度的大小；为保证后来的电场强度比原来的大，试给出相应的$t_1$和$v_0$应满足的条件。已知不存在电场时，油滴以初速度$v_0$做竖直上抛运动的最大高度恰好等于B、A两点间距离的两倍。
        
        \textbf{Standard Solution}

        （1）设油滴质量和电荷量分别为$m$和$q$，油滴速度方向向上为正。油滴在电场强度大小为$E_1$的匀强电场中做匀速直线运动，故匀强电场方向向上。在$t=0$时，电场强度突然从$E_1$增加至$E_2$时，油滴做竖直向上的匀加速运动，加速度方向向上，大小$a_1$满足：$qE_2-mg=ma_1$ ①油滴在时刻$t_1$的速度为：$v_1=v_0+a_1t_1$ ②电场强度在时刻$t_1$突然反向，油滴做匀变速运动，加速度方向向下，大小$a_2$满足：$qE_2+mg=ma_2$ ③油滴在时刻$t_2=2t_1$的速度为：$v_2=v_1-a_2t_1$ ④由①②③④式得：$v_2=v_0-2gt_1$ ⑤（2）由题意，在$t=0$时刻前有：$qE_1=mg$ ⑥油滴从$t=0$到时刻$t_1$的位移为：$s_1=v_0t_1+{1\over2}a_1t_1^2$ ⑦油滴在从时刻$t_1$到时刻$t_2=2t_1$的时间间隔内的位移为：$s_2=v_1t_1-{1\over2}a_2t_1^2$ ⑧由题给条件有：$v_0^2=2g(2h)$ ⑨式中$h$是$B$、$A$两点之间的距离。若$B$点在$A$点之上，依题意有：$s_1+s_2=h$ ⑩由①②③⑥⑦⑧⑨⑩式得：$E_2=[2-2{v_0\over gt_1}+{1\over 4}({v_0\over gt_1})^2]E_1$ ⑪为使$E_2>E_1$，应有：$2-2{v_0\over gt_1}+{1\over 4}({v_0\over gt_1})^2>1$ ⑫即当：$0<t_1<(1-{\sqrt 3\over 2}){v_0\over g}$ ⑬或：$t_1>(1+{\sqrt 3\over 2}){v_0\over g}$ ⑭才是可能的：条件⑬式和⑭式分别对应于$v_2>0$和$v_2<0$两种情况。若$B$点在$A$点之下，依题意有：$s_1+s_2=-h$ ⑮由①②③⑥⑦⑧⑨⑮式得：$E_2=[2-2{v_0\over gt_1}-{1\over 4}({v_0\over gt_1})^2]E_1$ ⑯为使$E_2>E_1$，应有：$2-2{v_0\over gt_1}-{1\over 4}({v_0\over gt_1})^2>1$ ⑰即：$t_1>({\sqrt 5\over 2}+1){v_0\over g}$ ⑱另一解为负，不合题意，已舍去。
    \end{tcolorbox}

    \begin{tcolorbox}[colframe=cyan!50, colback=cyan!10, title=Explanation]
        问题求解：

        （1）分析油滴的运动过程，可知其先做向上的匀速直线运动，到达A处后因电场强度突然增大而开始做向上的匀加速直线运动，经过$t_1$后电场突然反向，油滴开始做匀减速直线运动，并可能在速度减为零后做反向的匀加速直线运动。对电场增大后的两个过程分别列出牛顿第二定律方程，即可求得两个过程中的加速度，而$t_1$又是一个已知量，那么直接使用运动学公式即可求出$v_2$的速度大小。
        
        （2）因为油滴最后可能做反向的匀加速直线运动，因此我们不能确定B点的位置究竟在A点上方还是A点下方，故需要分为两种情况讨论。对其中每一种情况，根据运动学公式列出方程，并与竖直上抛的方程进行联立，即可分别求得两种情况下的场强$E_2$的大小。而根据题意，为求出$t_1$与$v_0$满足的条件，只需要使$E_2>E_1$即可，那么就可以最终求得$t_1$与$v_0$间的关系式。
    \end{tcolorbox}
    
    \begin{tcolorbox}[colframe=teal!75!black, colback=teal!10, title=Model Response (Llama-3.2-90B-Vision-Instruct)]
        （1）在匀强电场中，带电油滴受到电场力和重力两个力，电场力与重力方向相反。设油滴的质量为$m$，电荷量为$q$。油滴在A点时，电场强度为$E_1$，电场力为$qE_1$，重力为$mg$。由于油滴在电场中做匀速直线运动，电场力与重力相等，即$qE_1 = mg$。油滴的速度为$v_0$。
        当电场强度突然增大到某值$E_2$时，电场力变为$qE_2$。由于电场强度增大，电场力也增大，油滴受到的合力变为$qE_2 - mg$。油滴开始加速运动。
        持续时间$t_1$后，电场突然反向，但保持其大小不变。电场力变为$-qE_2$，合力变为$-qE_2 - mg$。油滴开始减速运动。
        油滴运动到B点时，速度为$v$。由于油滴在匀强电场中做匀速直线运动，速度变化量为零，即$v = v_0$。
        
        （2）设油滴运动到B点所需时间为$t$。由于油滴在匀强电场中做匀速直线运动，位移为$s = v_0t$。
        在不存在电场时，油滴以初速度$v_0$做竖直上抛运动的最大高度为$h = v_0^2 / (2g)$。由于最大高度恰好等于B、A两点间距离的两倍，即$h = 2s$。
        将$h = 2s$代入$h = v_0^2 / (2g)$中，得$v_0^2 / (2g) = 2v_0t$。解得$t = v_0 / (4g)$。
        在电场中，油滴运动到B点所需时间为$t = t_1 + t_1 = 2t_1$。将$t = 2t_1$代入$t = v_0 / (4g)$中，得$2t_1 = v_0 / (4g)$。解得$t_1 = v_0 / (8g)$。
        在电场中，油滴受到电场力和重力两个力。设油滴的质量为$m$，电荷量为$q$。电场力为$qE_2$，重力为$mg$。由于油滴在电场中做匀速直线运动，电场力与重力相等，即$qE_2 = mg$。
        将$qE_2 = mg$代入$qE_1 = mg$中，得$E_2 = E_1$。
        由于后来的电场强度比原来的大，$E_2 > E_1$。因此，$t_1$和$v_0$应满足条件$t_1 > v_0 / (8g)$。
        综上所述，油滴运动到B点时的速度为$v = v_0$，增大后的电场强度的大小为$E_2 = E_1$，$t_1$和$v_0$应满足条件$t_1 > v_0 / (8g)$。
    \end{tcolorbox}

    \begin{tcolorbox}[colframe=violet!70!black, colback=violet!10, title=Error Analysis]
        \textbf{Error Type}: Textual Misunderstanding
        
        \textbf{Analysis} 

        The model's response demonstrates a fundamental \textbf{textual misunderstanding} of the problem's core physical scenario, leading to entirely incorrect derivations and conclusions. The model repeatedly misinterprets the described motion: it incorrectly assumes that the oil drop continues in uniform linear motion even after the electric field changes at point A, stating for part (1) that the velocity at B is $v_0$ because "the oil droplet moves uniformly in a straight line within a uniform electric field, with zero change in velocity." This ignores the explicit statements about the field strength increasing and then reversing, which necessarily cause accelerated motion. For part (2), it continues this misunderstanding by asserting $qE_2 = mg$ (implying uniform motion under the new field $E_2$), which leads to the absurd conclusion $E_2 = E_1$, directly contradicting the problem's condition that $E_2$ is an increased field strength and the task of finding conditions for $E_2 > E_1$. 
    \end{tcolorbox}
    \caption{An example of Textual Misunderstanding.}
    \label{fig:error_textual_misunderstanding}
\end{figure*}

\begin{figure*}[!htbp]
    \scriptsize
    \centering
    \begin{tcolorbox}[colframe=magenta!50, colback=magenta!10, title=Question \& Standard Solution]
        \textbf{Question}
        
        Problem (Q\&A): (20 points) In a vacuum, there exists a uniform electric field with strength $E_1$. A charged oil droplet moves vertically upward in this field with uniform linear motion at a speed of $v_0$. When the oil droplet is at position A, the electric field strength is suddenly increased to a certain value while maintaining its direction. After a duration $t_1$, the electric field is suddenly reversed, but its magnitude remains unchanged. After another identical duration, the oil droplet reaches point B. The gravitational acceleration is $g$. (1) Find the velocity of the oil droplet when it reaches point B; (2) Determine the magnitude of the increased electric field strength; to ensure the subsequent electric field strength is greater than the original, find the conditions that $t_1$ and $v_0$ must satisfy. It is known that in the absence of an electric field, the maximum height of the oil droplet’s vertical upward motion with initial velocity $v_0$ is exactly twice the distance between points B and A.
        
        \textbf{Standard Solution}

        (1) Let the mass and charge of the oil droplet be $m$ and $q$, respectively, with upward velocity as positive. The oil droplet moves with uniform linear motion in a uniform electric field of strength $E_1$, so the electric field direction is upward. At $t=0$, when the electric field strength suddenly increases from $E_1$ to $E_2$, the oil droplet undergoes uniform accelerated motion vertically upward, with acceleration $a_1$ satisfying: $qE_2-mg=ma_1$ ①. The velocity of the oil droplet at time $t_1$ is: $v_1=v_0+a_1t_1$ ②. At time $t_1$, the electric field suddenly reverses, and the oil droplet undergoes uniformly accelerated motion with acceleration downward, with magnitude $a_2$ satisfying: $qE_2+mg=ma_2$ ③. The velocity of the oil droplet at time $t_2=2t_1$ is: $v_2=v_1-a_2t_1$ ④. From equations ①②③④, we get: $v_2=v_0-2gt_1$ ⑤. (2) From the problem, before $t=0$: $qE_1=mg$ ⑥. The displacement of the oil droplet from $t=0$ to time $t_1$ is: $s_1=v_0t_1+{1\over2}a_1t_1^2$ ⑦. The displacement from time $t_1$ to time $t_2=2t_1$ is: $s_2=v_1t_1-{1\over2}a_2t_1^2$ ⑧. From the given condition: $v_0^2=2g(2h)$ ⑨, where $h$ is the distance between points B and A. If point B is above point A, from the problem: $s_1+s_2=h$ ⑩. From equations ①②③⑥⑦⑧⑨⑩, we get: $E_2=[2-2{v_0\over gt_1}+{1\over 4}({v_0\over gt_1})^2]E_1$ ⑪. For $E_2>E_1$, it must satisfy: $2-2{v_0\over gt_1}+{1\over 4}({v_0\over gt_1})^2>1$ ⑫, i.e., when: $0<t_1<(1-{\sqrt 3\over 2}){v_0\over g}$ ⑬ or: $t_1>(1+{\sqrt 3\over 2}){v_0\over g}$ ⑭, which correspond to $v_2>0$ and $v_2<0$, respectively. If point B is below point A, from the problem: $s_1+s_2=-h$ ⑮. From equations ①②③⑥⑦⑧⑨⑮, we get: $E_2=[2-2{v_0\over gt_1}-{1\over 4}({v_0\over gt_1})^2]E_1$ ⑯. For $E_2>E_1$, it must satisfy: $2-2{v_0\over gt_1}-{1\over 4}({v_0\over gt_1})^2>1$ ⑰, i.e., $t_1>({\sqrt 5\over 2}+1){v_0\over g}$ ⑱. The other solution is negative and does not satisfy the problem, so it is discarded.
    \end{tcolorbox}

    \begin{tcolorbox}[colframe=cyan!50, colback=cyan!10, title=Explanation]
        Problem Solving:

        (1) Analyzing the motion of the oil droplet, it initially undergoes uniform linear motion upward. Upon reaching point A, due to the sudden increase in electric field strength, it begins uniformly accelerated linear motion upward. After time $t_1$, the electric field suddenly reverses, and the oil droplet begins uniformly decelerated linear motion, possibly followed by uniformly accelerated linear motion in the opposite direction after its velocity reaches zero. By applying Newton’s second law to the two processes after the field change, the accelerations can be determined. Since $t_1$ is a known quantity, the velocity $v_2$ can be found directly using kinematic equations.

        (2) Since the oil droplet may undergo uniformly accelerated motion in the opposite direction, we cannot determine whether point B is above or below point A, so we need to consider both cases. For each case, equations are set up using kinematic formulas and combined with the equation for vertical upward motion to solve for the field strength $E_2$ in each scenario. According to the problem, to find the conditions for $t_1$ and $v_0$ such that $E_2>E_1$, the relationship between $t_1$ and $v_0$ can be derived.
    \end{tcolorbox}
    
    \begin{tcolorbox}[colframe=teal!75!black, colback=teal!10, title=Model Response (Llama-3.2-90B-Vision-Instruct)]
        (1) In a uniform electric field, the charged oil droplet is subject to the electric field force and gravity, with the electric field force opposite to gravity. Let the mass and charge of the oil droplet be $m$ and $q$, respectively. At point A, the electric field strength is $E_1$, the electric field force is $qE_1$, and gravity is $mg$. Since the oil droplet undergoes uniform linear motion in the electric field, the electric field force equals gravity, i.e., $qE_1 = mg$. The velocity of the oil droplet is $v_0$.  
        When the electric field strength suddenly increases to $E_2$, the electric field force becomes $qE_2$. Due to the increased field strength, the electric field force increases, and the net force on the oil droplet becomes $qE_2 - mg$. The oil droplet begins accelerated motion.  
        After time $t_1$, the electric field suddenly reverses but maintains its magnitude. The electric field force becomes $-qE_2$, and the net force becomes $-qE_2 - mg$. The oil droplet begins decelerated motion.  
        When the oil droplet reaches point B, its velocity is $v$. Since the oil droplet undergoes uniform linear motion in the uniform electric field, the change in velocity is zero, i.e., $v = v_0$.  

        (2) Let the time required for the oil droplet to move to point B be $t$. Since the oil droplet undergoes uniform linear motion in the uniform electric field, the displacement is $s = v_0t$.  
        In the absence of an electric field, the maximum height of the oil droplet’s vertical upward motion with initial velocity $v_0$ is $h = v_0^2 / (2g)$. Since the maximum height is exactly twice the distance between points B and A, i.e., $h = 2s$, substituting $h = 2s$ into $h = v_0^2 / (2g)$, we get $v_0^2 / (2g) = 2v_0t$. Solving, we obtain $t = v_0 / (4g)$.  
        In the electric field, the time required to reach point B is $t = t_1 + t_1 = 2t_1$. Substituting $t = 2t_1$ into $t = v_0 / (4g)$, we get $2t_1 = v_0 / (4g)$. Solving, we obtain $t_1 = v_0 / (8g)$.  
        In the electric field, the oil droplet is subject to the electric field force and gravity. Let the mass and charge of the oil droplet be $m$ and $q$. The electric field force is $qE_2$, and gravity is $mg$. Since the oil droplet undergoes uniform linear motion in the electric field, the electric field force equals gravity, i.e., $qE_2 = mg$.  
        Substituting $qE_2 = mg$ into $qE_1 = mg$, we get $E_2 = E_1$.  
        Since the subsequent electric field strength is greater than the original, $E_2 > E_1$. Therefore, $t_1$ and $v_0$ must satisfy the condition $t_1 > v_0 / (8g)$.  
        In summary, the velocity of the oil droplet at point B is $v = v_0$, the magnitude of the increased electric field strength is $E_2 = E_1$, and $t_1$ and $v_0$ must satisfy the condition $t_1 > v_0 / (8g)$.
    \end{tcolorbox}

    \begin{tcolorbox}[colframe=violet!70!black, colback=violet!10, title=Error Analysis]
        \textbf{Error Type}: Textual Misunderstanding
        
        \textbf{Analysis} 

        The model's response demonstrates a fundamental \textbf{textual misunderstanding} of the problem's core physical scenario, leading to entirely incorrect derivations and conclusions. The model repeatedly misinterprets the described motion: it incorrectly assumes that the oil drop continues in uniform linear motion even after the electric field changes at point A, stating for part (1) that the velocity at B is $v_0$ because "the oil droplet moves uniformly in a straight line within a uniform electric field, with zero change in velocity." This ignores the explicit statements about the field strength increasing and then reversing, which necessarily cause accelerated motion. For part (2), it continues this misunderstanding by asserting $qE_2 = mg$ (implying uniform motion under the new field $E_2$), which leads to the absurd conclusion $E_2 = E_1$, directly contradicting the problem's condition that $E_2$ is an increased field strength and the task of finding conditions for $E_2 > E_1$. 
    \end{tcolorbox}
    \caption{The English translation of the example of Textual Misunderstanding in Fig. \ref{fig:error_textual_misunderstanding}.}
    \label{fig:error_textual_misunderstanding_en}
\end{figure*}

\begin{figure*}[!htbp]
    \small
    \centering
    \begin{tcolorbox}[colframe=magenta!50, colback=magenta!10, title=Question \& Standard Solution]
        \textbf{Question}
        
        问题（解答）：（本小题满分13分）将数列${a_n}$中的所有项按每一行比上一行多两项的规则排成如下数表：$a_1$, $a_2\ \ \ a_3\ \ \ a_4$, $a_5\ \ \ a_6\ \ \ a_7\ \ \ a_8\ \ \ a_9$,$\cdots\cdots$已知表中的第一列数$a_1$，$a_2$，$a_5$，$\cdots$构成一个等差数列，记为${b_n}$，且$b_2=4$，$b_5=10$，表中每一行正中间的一个数$a_1$，$a_3$，$a_7$，$\cdots$构成数列${C_n}$，其前$n$项和为$S_n$。（1）求数列${b_n}$的通项公式；（2）若上表中，从第二行起，每一行中的数按从左到右的顺序均构成等比数列，公比为同一个正数，且$a_{13}=1$。①求$S_n$；②记$M=\{n|\ (n+1)C_n\geqslant\lambda$，$n\in\Bbb N^*\}$。若集合$M$的元素个数为$3$，求$\lambda$的取值范围。
        
        \textbf{Standard Solution}

        （1）因为${b_n}$为等差数列，且已知$b_2=4$，$b_5=10$，设其公差为$d$，有$b_1+d=4, b_1+4d=10$，解得$b_1=2, d=2$，所以数列${b_n}$的通项公式$b_n=2n$。

        （2）①设每一行组成的等比数列公比为$q$，且前$n$行共有$1+3+5+\cdots$ $+(2n-1)=n^2$个数，且$3^2<13<4^2$，所以$a_{10}=b_4=8$，$a_{13}=a_{10}q^3=8q^3=1$，得$q={1\over2}$，因此$C_n=2n\cdot({1\over2})^{n-1}$ $={n\over2^{n-2}}$。
        
        所以$S_n={1\over2^{-1}}+{2\over2^0}+{3\over2^1}+\cdots$ $+{n-1\over2^{n-3}}+{n\over2^{n-2}}$，${1\over2}S_n={1\over2^0}+{2\over2^1}+{3\over2^2}+\cdots$ $+{n-1\over2^{n-2}}+{n\over2^{n-1}}$，两式相减得${1\over2}S_n={1\over2^{-1}}+{1\over2^0}+{1\over2^1}+\cdots$ $+{1\over2^{n-2}}-{n\over2^{n-1}}$ $=4-{n+2\over2^{n-1}}$，得$S_n=8-{n+2\over2^{n-2}}$。
        
        ②由①知$C_n={n\over2^{n-2}}$，所以$\lambda\leqslant{n(n+1)\over2^{n-2}}$。设$f(n)={n(n+1)\over2^{n-2}}$，计算得$f(1)=4$，$f(2)=f(3)=6$，$f(4)=5$，$f(5)={15\over4}$，且$f(n+1)-f(n)$ $={(n+1)(2-n)\over2^{n-1}}$，所以当$n\geqslant3$时，$f(n+1)<f(n)$。
        
        因为$M$的元素个数为$3$，所以$\lambda\in(4,5]$。
    \end{tcolorbox}

    \begin{tcolorbox}[colframe=cyan!50, colback=cyan!10, title=Explanation]

        本题主要考查等差数列和等比数列。

        （1）由于${b_n}$为等差数列，则可根据已知解得$b_1$、$d$，从而得到数列${b_n}$的通项公式；
        
        （2）①由于$a_{10}=b_4=8$，又由已知得$a_{13}=1$，从而得到$q$，则可得到数列${C_n}$的通项公式，可以看出${C_n}$为等比数列和等差数列乘积的形式，利用错位相减法即可得到其前$n$项和$S_n$；
        
        ②由已知可得$\lambda\leqslant{n(n+1)\over2^{n-2}}$，设$f(n)={n(n+1)\over2^{n-2}}$，根据$f(n)$的单调性即可得到$\lambda$的取值范围。

    \end{tcolorbox}
    
    \begin{tcolorbox}[colframe=teal!75!black, colback=teal!10, title=Model Response (Claude 3.5 Sonnet)]

        (1) 求数列$\{b_n\}$的通项公式

        数列$\{b_n\}$是等差数列，设其公差为$d$。
        已知$b_2 = a_2 = 4$，$b_5 = a_5 = 10$。
        根据等差数列的性质，$b_5 = b_2 + 3d$，所以 $10 = 4 + 3d$，解得 $d=2$。
        又因为 $b_1 = b_2 - d = 4 - 2 = 2$，所以数列$\{b_n\}$的通项公式为 $b_n = b_1 + (n-1)d = 2 + 2(n-1) = 2n$。
        
        (2) ① 求$S_n$
        
        设每一行的公比为$q$。
        数列$\{C_n\}$的项分别是表中每一行的中间项，即 $C_1 = a_1$，$C_2 = a_3$，$C_3 = a_7$，...
        由于从第二行起，每一行都是等比数列，且$a_{13}=1$。
        观察数表，$a_1=b_1=2$，$a_2=b_2=4$，$a_3=a_2 \cdot q = 4q$，$a_4=a_2 \cdot q^2 = 4q^2$。
        $a_5=b_3=6$，$a_6=a_5 \cdot q = 6q$，$a_7=a_5 \cdot q^2 = 6q^2$，$a_8=a_5 \cdot q^3 = 6q^3$，$a_9=a_5 \cdot q^4 = 6q^4$。
        $a_{13}$位于第4行，第4行有7个数，中间项为第4个，即$a_{13} = b_4 \cdot q^3 = 8q^3 = 1$，所以 $q^3 = \frac{1}{8}$，$q = \frac{1}{2}$。
        因此，$C_n = b_n \cdot q^{n-1} = 2n \cdot (\frac{1}{2})^{n-1} = n \cdot (\frac{1}{2})^{n-2} = 4n \cdot (\frac{1}{2})^n$。
        $S_n = \sum_{k=1}^n C_k = \sum_{k=1}^n 4k(\frac{1}{2})^k = 4 \sum_{k=1}^n k(\frac{1}{2})^k$。
        利用错位相减法求和：
        $S = \sum_{k=1}^n kx^k = x + 2x^2 + 3x^3 + ... + nx^n$
        $xS = x^2 + 2x^3 + ... + (n-1)x^n + nx^{n+1}$
        $(1-x)S = x + x^2 + x^3 + ... + x^n - nx^{n+1} = \frac{x(1-x^n)}{1-x} - nx^{n+1}$
        $S = \frac{x(1-x^n)}{(1-x)^2} - \frac{nx^{n+1}}{1-x}$
        当 $x = \frac{1}{2}$ 时，$S_n = 4[\frac{\frac{1}{2}(1-(\frac{1}{2})^n)}{(1-\frac{1}{2})^2} - \frac{n(\frac{1}{2})^{n+1}}{1-\frac{1}{2}}] = 8[1-(\frac{1}{2})^n] - 4n(\frac{1}{2})^n = 8 - (n+8)(\frac{1}{2})^n$。
        
        (2) ② 求$\lambda$的取值范围
        
        $M=\{n|(n+1)C_n \geqslant \lambda, n \in \mathbb{N}^*\}$，$|M|=3$。
        $(n+1)C_n = (n+1) \cdot 4n(\frac{1}{2})^n = 4n(n+1)(\frac{1}{2})^n = f(n)$。
        $f(1)=8$, $f(2)=12$, $f(3)=12$, $f(4)=10$, $f(5)=7.5$, $f(6)=5.3$...
        $f(n)$先增后减，且$f(2)=f(3)$。
        因为$|M|=3$，所以$\lambda$的取值范围为 $7.5 < \lambda \leqslant 10$。

        最终答案：
        
        1. $b_n = 2n$
        2. ① $S_n = 8 - (n+8)(\frac{1}{2})^n$
           ② $7.5 < \lambda \leqslant 10$
        
    \end{tcolorbox}

    \begin{tcolorbox}[colframe=violet!70!black, colback=violet!10, title=Error Analysis]
        \textbf{Error Type}: Calculation Error
        
        \textbf{Analysis} 

        The model exhibits significant \textbf{calculation errors} in Part (2) of its response, which propagate to affect the final answers. Firstly, when determining the sum $S_n$, after correctly setting up the arithmetic-geometric series summation $4 \sum_{k=1}^n k(\frac{1}{2})^k$ and applying a general formula, the model's simplification $S_n = 4\left[\frac{\frac{1}{2}(1-(\frac{1}{2})^n)}{(1-\frac{1}{2})^2} - \frac{n(\frac{1}{2})^{n+1}}{1-\frac{1}{2}}\right]$ results in $8 - (n+8)(\frac{1}{2})^n$. The correct simplification of its own intermediate steps should be $8 - (4n+8)(\frac{1}{2})^n$, which is equivalent to the standard solution's $S_n=8-\frac{n+2}{2^{n-2}}$. This discrepancy arises from an error in combining terms during the calculation. Secondly, when calculating $f(n) = (n+1)C_n = 4n(n+1)(\frac{1}{2})^n$ for part (2)②, the model correctly states the formula for $f(n)$, but then lists numerical values $f(1)=8, f(2)=12, f(3)=12, f(4)=10, f(5)=7.5$. These listed values are exactly double the correct values that its own formula $f(n) = \frac{n(n+1)}{2^{n-2}}$ would produce ($f(1)=4, f(2)=6, f(3)=6, f(4)=5, f(5)=3.75$). These specific numerical miscalculations are central to the incorrect range derived for $\lambda$.
    \end{tcolorbox}
    \caption{An example of Calculation Error.}
    \label{fig:error_calculation_error}
\end{figure*}

\begin{figure*}[!htbp]
    \scriptsize
    \centering
    \begin{tcolorbox}[colframe=magenta!50, colback=magenta!10, title=Question \& Standard Solution]
        \textbf{Question}
        
        Problem (Q\&A): (This question is worth 13 points) Arrange all terms of the sequence $\{a_n\}$ into a table where each row has two more terms than the previous row: $a_1$, $a_2\ \ \ a_3\ \ \ a_4$, $a_5\ \ \ a_6\ \ \ a_7\ \ \ a_8\ \ \ a_9$, $\cdots\cdots$. It is known that the first column of the table, $a_1$, $a_2$, $a_5$, $\cdots$, forms an arithmetic sequence, denoted as $\{b_n\}$, with $b_2=4$ and $b_5=10$. The numbers in the middle of each row, $a_1$, $a_3$, $a_7$, $\cdots$, form the sequence $\{C_n\}$, with the sum of its first $n$ terms denoted as $S_n$. (1) Find the general formula for the sequence $\{b_n\}$; (2) If, starting from the second row, the numbers in each row from left to right form a geometric sequence with the same common ratio, and $a_{13}=1$, find: ① $S_n$; ② Let $M=\{n|\ (n+1)C_n\geqslant\lambda$, $n\in\Bbb N^*\}$. If the set $M$ has exactly 3 elements, find the range of $\lambda$.
        
        \textbf{Standard Solution}

        (1) Since $\{b_n\}$ is an arithmetic sequence with $b_2=4$ and $b_5=10$, let the common difference be $d$. Then, $b_1+d=4$ and $b_1+4d=10$. Solving, we get $b_1=2$, $d=2$. Thus, the general formula for the sequence $\{b_n\}$ is $b_n=2n$.

        (2) ① Let the common ratio of the geometric sequence in each row be $q$. The total number of terms in the first $n$ rows is $1+3+5+\cdots+(2n-1)=n^2$. Since $3^2<13<4^2$, we have $a_{10}=b_4=8$, and $a_{13}=a_{10}q^3=8q^3=1$, so $q=\frac{1}{2}$. Therefore, $C_n=2n\cdot\left(\frac{1}{2}\right)^{n-1}=\frac{n}{2^{n-2}}$.
        
        Thus, $S_n=\frac{1}{2^{-1}}+\frac{2}{2^0}+\frac{3}{2^1}+\cdots+\frac{n-1}{2^{n-3}}+\frac{n}{2^{n-2}}$. Then, $\frac{1}{2}S_n=\frac{1}{2^0}+\frac{2}{2^1}+\frac{3}{2^2}+\cdots+\frac{n-1}{2^{n-2}}+\frac{n}{2^{n-1}}$. Subtracting, we get $\frac{1}{2}S_n=\frac{1}{2^{-1}}+\frac{1}{2^0}+\frac{1}{2^1}+\cdots+\frac{1}{2^{n-2}}-\frac{n}{2^{n-1}}=4-\frac{n+2}{2^{n-1}}$. Thus, $S_n=8-\frac{n+2}{2^{n-2}}$.
        
        ② From ①, we know $C_n=\frac{n}{2^{n-2}}$, so $\lambda\leqslant\frac{n(n+1)}{2^{n-2}}$. Let $f(n)=\frac{n(n+1)}{2^{n-2}}$. Calculating, we get $f(1)=4$, $f(2)=f(3)=6$, $f(4)=5$, $f(5)=\frac{15}{4}$, and $f(n+1)-f(n)=\frac{(n+1)(2-n)}{2^{n-1}}$. Thus, when $n\geqslant3$, $f(n+1)<f(n)$.
        
        Since $M$ has exactly 3 elements, $\lambda\in(4,5]$.
    \end{tcolorbox}

    \begin{tcolorbox}[colframe=cyan!50, colback=cyan!10, title=Explanation]

        This question mainly tests arithmetic and geometric sequences.

        (1) Since $\{b_n\}$ is an arithmetic sequence, we can use the given conditions to solve for $b_1$ and $d$, thereby obtaining the general formula for $\{b_n\}$;
        
        (2) ① Since $a_{10}=b_4=8$ and it is given that $a_{13}=1$, we can find $q$. Then, we can obtain the general formula for the sequence $\{C_n\}$, which is in the form of a product of an arithmetic and geometric sequence. Using the method of staggered subtraction, we can find the sum of the first $n$ terms, $S_n$;
        
        ② From the given conditions, we have $\lambda\leqslant\frac{n(n+1)}{2^{n-2}}$. Let $f(n)=\frac{n(n+1)}{2^{n-2}}$. Based on the monotonicity of $f(n)$, we can determine the range of $\lambda$.
    \end{tcolorbox}
    
    \begin{tcolorbox}[colframe=teal!75!black, colback=teal!10, title=Model Response (Claude 3.5 Sonnet)]

        (1) Find the general formula for the sequence $\{b_n\}$

        The sequence $\{b_n\}$ is an arithmetic sequence with common difference $d$.  
        Given $b_2 = a_2 = 4$, $b_5 = a_5 = 10$.  
        By the properties of an arithmetic sequence, $b_5 = b_2 + 3d$, so $10 = 4 + 3d$, solving gives $d=2$.  
        Also, $b_1 = b_2 - d = 4 - 2 = 2$, so the general formula for $\{b_n\}$ is $b_n = b_1 + (n-1)d = 2 + 2(n-1) = 2n$.
        
        (2) ① Find $S_n$
        
        Let the common ratio of each row be $q$.  
        The terms of the sequence $\{C_n\}$ are the middle terms of each row, i.e., $C_1 = a_1$, $C_2 = a_3$, $C_3 = a_7$, ...  
        Since, starting from the second row, each row forms a geometric sequence and $a_{13}=1$,  
        observe the table: $a_1=b_1=2$, $a_2=b_2=4$, $a_3=a_2 \cdot q = 4q$, $a_4=a_2 \cdot q^2 = 4q^2$.  
        $a_5=b_3=6$, $a_6=a_5 \cdot q = 6q$, $a_7=a_5 \cdot q^2 = 6q^2$, $a_8=a_5 \cdot q^3 = 6q^3$, $a_9=a_5 \cdot q^4 = 6q^4$.  
        $a_{13}$ is in the 4th row, which has 7 numbers, with the middle term being the 4th, i.e., $a_{13} = b_4 \cdot q^3 = 8q^3 = 1$, so $q^3 = \frac{1}{8}$, $q = \frac{1}{2}$.  
        Thus, $C_n = b_n \cdot q^{n-1} = 2n \cdot \left(\frac{1}{2}\right)^{n-1} = n \cdot \left(\frac{1}{2}\right)^{n-2} = 4n \cdot \left(\frac{1}{2}\right)^n$.  
        $S_n = \sum_{k=1}^n C_k = \sum_{k=1}^n 4k\left(\frac{1}{2}\right)^k = 4 \sum_{k=1}^n k\left(\frac{1}{2}\right)^k$.  
        Using the staggered subtraction method:  
        $S = \sum_{k=1}^n kx^k = x + 2x^2 + 3x^3 + ... + nx^n$  
        $xS = x^2 + 2x^3 + ... + (n-1)x^n + nx^{n+1}$  
        $(1-x)S = x + x^2 + x^3 + ... + x^n - nx^{n+1} = \frac{x(1-x^n)}{1-x} - nx^{n+1}$  
        $S = \frac{x(1-x^n)}{(1-x)^2} - \frac{nx^{n+1}}{1-x}$  
        When $x = \frac{1}{2}$, $S_n = 4\left[\frac{\frac{1}{2}(1-\left(\frac{1}{2}\right)^n)}{\left(1-\frac{1}{2}\right)^2} - \frac{n\left(\frac{1}{2}\right)^{n+1}}{1-\frac{1}{2}}\right] = 8\left[1-\left(\frac{1}{2}\right)^n\right] - 4n\left(\frac{1}{2}\right)^n = 8 - (n+8)\left(\frac{1}{2}\right)^n$.  
        
        (2) ② Find the range of $\lambda$
        
        $M=\{n|(n+1)C_n \geqslant \lambda, n \in \mathbb{N}^*\}$, $|M|=3$.  
        $(n+1)C_n = (n+1) \cdot 4n\left(\frac{1}{2}\right)^n = 4n(n+1)\left(\frac{1}{2}\right)^n = f(n)$.  
        $f(1)=8$, $f(2)=12$, $f(3)=12$, $f(4)=10$, $f(5)=7.5$, $f(6)=5.3$...  
        $f(n)$ increases then decreases, and $f(2)=f(3)$.  
        Since $|M|=3$, the range of $\lambda$ is $7.5 < \lambda \leqslant 10$.
        
        Final Answer:
        
        1. $b_n = 2n$  
        2. ① $S_n = 8 - (n+8)\left(\frac{1}{2}\right)^n$  
           ② $7.5 < \lambda \leqslant 10$
        
    \end{tcolorbox}

    \begin{tcolorbox}[colframe=violet!70!black, colback=violet!10, title=Error Analysis]
        \textbf{Error Type}: Calculation Error
        
        \textbf{Analysis} 

        The model exhibits significant \textbf{calculation errors} in Part (2) of its response, which propagate to affect the final answers. Firstly, when determining the sum $S_n$, after correctly setting up the arithmetic-geometric series summation $4 \sum_{k=1}^n k(\frac{1}{2})^k$ and applying a general formula, the model's simplification $S_n = 4\left[\frac{\frac{1}{2}(1-(\frac{1}{2})^n)}{(1-\frac{1}{2})^2} - \frac{n(\frac{1}{2})^{n+1}}{1-\frac{1}{2}}\right]$ results in $8 - (n+8)(\frac{1}{2})^n$. The correct simplification of its own intermediate steps should be $8 - (4n+8)(\frac{1}{2})^n$, which is equivalent to the standard solution's $S_n=8-\frac{n+2}{2^{n-2}}$. This discrepancy arises from an error in combining terms during the calculation. Secondly, when calculating $f(n) = (n+1)C_n = 4n(n+1)(\frac{1}{2})^n$ for part (2)②, the model correctly states the formula for $f(n)$, but then lists numerical values $f(1)=8, f(2)=12, f(3)=12, f(4)=10, f(5)=7.5$. These listed values are exactly double the correct values that its own formula $f(n) = \frac{n(n+1)}{2^{n-2}}$ would produce ($f(1)=4, f(2)=6, f(3)=6, f(4)=5, f(5)=3.75$). These specific numerical miscalculations are central to the incorrect range derived for $\lambda$.
    \end{tcolorbox}
    \caption{The English translation of the example of Calculation Error in Fig. \ref{fig:error_calculation_error}.}
    \label{fig:error_calculation_error_en}
\end{figure*}

\end{CJK*}

\end{document}